\documentclass[letterpaper, 10 pt, journal, twoside]{ieeetran}
\IEEEoverridecommandlockouts

\usepackage{pdflscape}
\graphicspath{{./images/}}
\usepackage{amsfonts}
\usepackage{booktabs}
\usepackage{multirow}
\usepackage{amsmath}
\usepackage{bm}
\usepackage{hyperref}
\hypersetup{
    colorlinks=true,
    linkcolor=blue,
    filecolor=magenta,      
    urlcolor=cyan,
    pdftitle={Overleaf Example},
    pdfpagemode=FullScreen,
    }
\usepackage[export]{adjustbox}
\usepackage[normalem]{ulem}
\usepackage{tabularx,ragged2e,caption}
\usepackage{array}
\usepackage[style=ieee,sorting=none,backend=bibtex]{biblatex}
\newcolumntype{A}{>{\arraybackslash}m{3em}}
\newcolumntype{B}{>{\centering\arraybackslash}m{1em}}
\newcolumntype{C}{>{\centering\arraybackslash}m{2em}}
\newcolumntype{D}{>{\centering\arraybackslash}m{5.5em}}
\newcolumntype{E}{>{\arraybackslash}m{21em}}
\newcolumntype{F}{>{\arraybackslash}m{0.95\columnwidth}}
\newcommand{\rp}[1]{\textcolor{black}{#1}}
\newcommand{\rps}[1]{\textcolor{red}{#1}}

\usepackage{caption}
\usepackage{subcaption}
\title{\LARGE \bf
DDBot: Differentiable Physics-based Digging Robot for Unknown Granular Materials
}

\author{Xintong Yang$^{1}$, Minglun Wei$^{1}$, Yu-Kun Lai$^{2}$, Ze Ji$^{1}$% <-this % stops a space
\thanks{}% <-this % stops a space
\thanks{$^{1}$Xintong Yang, Minglun Wei and Ze Ji are with the School of Engineering, Cardiff University, Cardiff, United Kingdom. {\tt\small \{yangx66, weim9, jiz1\}@cardif.ac.uk}}%
\thanks{$^{2}$Yu-Kun Lai is with the School of Computer Science and Informatics, Cardiff University, Cardiff, United Kingdom. {\tt\small laiy4@cardif.ac.uk}}%
}

\begin{document}

\maketitle
\thispagestyle{empty}
\pagestyle{empty}

%%%%%%%%%%%%%%%%%%%%%%%%%%%%%%%%%%%%%%%%%%%%%%%%%%%%%%%%%%%%%%%%%%%%%%%%%%%%%%%%
\begin{abstract}
Automating the manipulation of granular materials poses significant challenges due to complex contact dynamics, unpredictable material properties, and intricate system states. Existing approaches often fail to achieve efficiency and accuracy in such tasks. To fill the research gap, this paper studies the small-scale and high-precision granular material digging task with unknown physical properties. A new framework, named differentiable digging robot (DDBot), is proposed to manipulate granular materials, including sand and soil.

Specifically, we equip DDBot with a differentiable physics-based simulator, tailored for granular material manipulation, powered by GPU-accelerated parallel computing and automatic differentiation. DDBot can perform efficient differentiable system identification and high-precision digging skill optimisation for unknown granular materials, which is enabled by a differentiable skill-to-action mapping, a task-oriented demonstration method, gradient clipping and line search-based gradient descent.

Experimental results show that DDBot can efficiently (converge within $5$ to $20$ minutes) identify unknown granular material dynamics and optimise digging skills, with high-precision results in zero-shot real-world deployments, highlighting its practicality. Benchmark results against state-of-the-art baselines also confirm the robustness and efficiency of DDBot in such digging tasks.
\end{abstract}

%%%%%%%%%%%%%%%%%%%%%%%%%%%%%%%%%%%%%%%%%%%%%%
\section{INTRODUCTION} \label{sec:intro}
Granular material is everywhere in our daily life, ranging from sand, soil, rocks, rice, sweetcorn, beans, etc. Automating the manipulation of such materials is of vital importance. For instance, it is highly valuable to the development of a robotic farm or greenhouse, where machines need to dig soil for planting. It is also of high value to robotic excavation tasks, which play an important role in disaster rescue missions or object retrieval in out-of-reach places. Moreover, it is also applicable to soil sampling tasks that need to be carried out in dangerous caves, mountains, or even other planets that humans cannot reach. Unfortunately, state-of-the-art robotic manipulation methods still struggle to handle such materials with high precision, flexibility and efficiency.

In this work, we aim to develop a high-precision and efficient framework for granular material manipulation, designed for real-world tasks with unknown material properties, extended task horizons, and diverse manipulation targets. Our approach focuses on small-scale but high-precision digging tasks with varying target sizes and locations. Such tasks typically require a robot to precisely control the depth and location to insert a shovel and the angle and distance to push the material.

The first challenge in developing a practical granular material manipulation system lies in the accurate and efficient prediction of contact dynamics, which is crucial for optimising robotic manipulation solutions~\cite{kroemer2021review}. Advances in simulation techniques have significantly progressed rigid object manipulation in the last few decades. In recent years, numerical modelling techniques, such as the material point method (MPM) and its variants, have greatly improved the fidelity and efficiency of simulations of non-rigid materials~\cite{lin2022diffskill,xian2023fluidlab}. 
MPM models non-rigid materials as particle collections, calculating dynamics based on physical laws on the particles and a background Eulerian grid, achieving superior accuracy compared to data-driven approaches~\cite{hu2018moving}. However, it remains unintuitive to tune the values of granular material properties, such as Young's modulus, Poisson's ratio, density, and friction angle, to match the real-world dynamics, hindering the development of high-precision manipulation solutions. While differentiable physics-based system identification (DPSI) has demonstrated efficiency for deformable objects, like plasticine~\cite{xy2024sysid}, its feasibility on granular materials remains unexplored.

\rp{Secondly, there is a lack of investigation of how state-of-the-art planning and control methods perform in such tasks.}
The most popular learning-based method for robotic manipulation in recent years is arguably deep reinforcement learning (DRL), which has proved successful in many robotic tasks~\cite{kroemer2021review}. DRL methods seek to learn a policy that outputs actions to maximise a certain reward function via (mostly) randomised exploration~\cite{sutton2018reinforcement}. However, \rp{DRL is notoriously data-hungry, and can be easily stuck in local minima due to limited positive feedback. DRL also struggles to generalise to unseen scenarios without fine-tuning or retraining. On the other hand, model predictive control (MPC) is another set of common methods for robotic planning and control~\cite{yin2021modeling}. MPC typically applies stochastic sampling or evolutionary methods to optimise a distribution of the manipulation motion with a specified cost function (or fitness function). Despite its successes in various domains, MPC methods tend to evolve solutions based on a cost function that evaluates the system's final state, lacking consideration of the physical dynamics. It ignores the connection between the manipulation skills and the resultant manipulation outcomes, leading to poor local minima and poor interpretability of the found solutions. In addition, it remains unclear how it may perform in such granular manipulation tasks.}

Recently, an emerging set of robotic manipulation methods that rely on differentiable physics-based simulators has shown promising results in generating manipulation solutions for non-rigid materials~\cite{xu2021accelerated,chen2022diffsrl}. These methods combine the realism and accuracy of the laws of physics (e.g., MPM), and the automated differentiation mechanism and parallel computing technique of the simulator (e.g., Torch, TaiChi)~\cite{hu2019difftaichi,newbury2024review} to calculate physically meaningful gradients of the manipulation trajectory with respect to a loss function. Though researchers have demonstrated the feasibility of direct gradient-based trajectory optimisation with non-rigid materials that are modelled by the fixed corotated elastic energy and the von Mises plastic criterion~\cite{huang2021plasticinelab,xian2023fluidlab}, \rp{no work has been done to study the feasibility of gradient-based methods on granular materials}, which are more accurately modelled by the St. Venant-Kirchoff (SVK) elastic energy and the Drucker-Prager (DP) plastic yield criterion~\cite{klar2016drucker}. In addition, first-order gradient descent methods, despite their efficiency, are sensitive to numerical instability, particularly in large systems with long trajectories and rugged loss landscapes~\cite{wei2024automachef}. Thus, there is a need to study the feasibility of gradient-based optimisation for granular material manipulation.

\rp{To fill the research gap, this paper makes the following contributions to the robotisation of real-world, small-scale and high-precision granular material digging tasks with unknown material properties.}

First, to the best of our knowledge, this is the first attempt to develop an efficient and high-fidelity differentiable simulator specifically designed for granular material manipulation. It supports efficient forward dynamics computation to train data-driven methods, such as DRL, and evaluate non-learning methods, such as MPC. It also enables efficient automatic differentiation to support gradient-based optimisation methods, such as RMSProp~\cite{tieleman2012rmsprop}.

\begin{figure*}
    \centering
    \includegraphics[width=\textwidth]{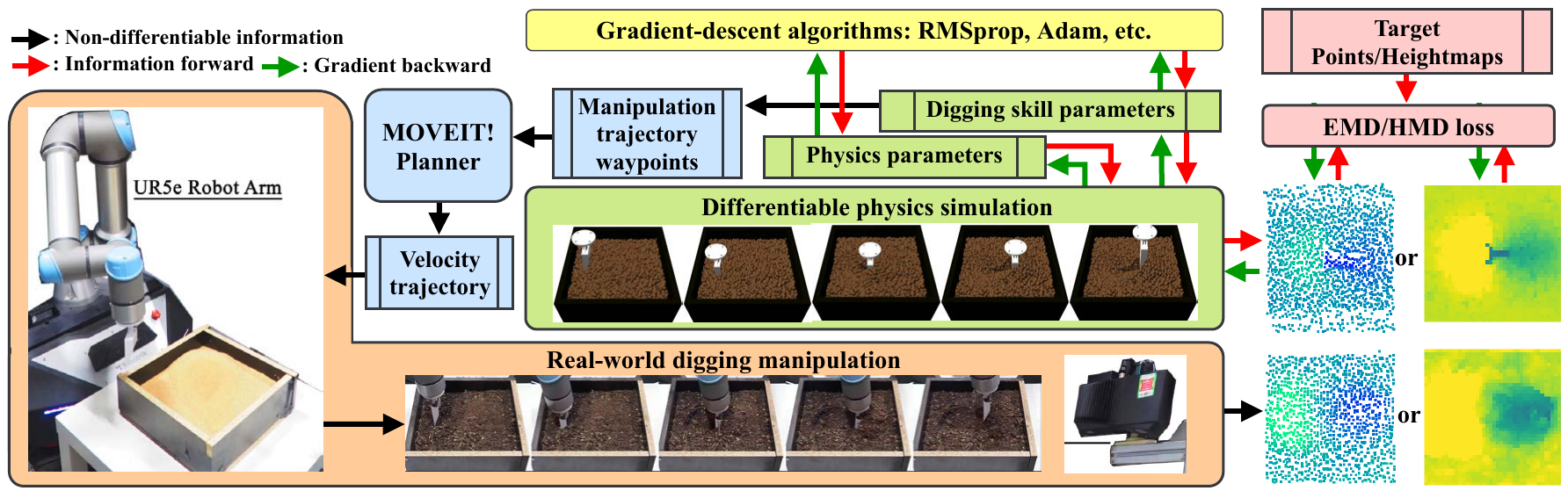}
    \caption{The DDBot (\underline{D}ifferentiable \underline{D}igging Ro\underline{bot}) system overview. DDBot has five modules. 1) A robot platform (orange) that performs real-world manipulation and collects real-world surface point clouds of the granular materials. \rp{The five small images visualise a real-world digging trajectory that matches the trajectory optimised in simulation shown in the green box.} 2) A differentiable physics simulator (green) that simulates the digging skill trajectory given the physics and skill parameters. 3) A differentiable loss function (pink) that compares the geometrical differences between the simulated and real manipulation results. 4) A gradient-based optimisation algorithm (yellow) that updates the physics or skill parameters. 5) The Robot Operating System (ROS) communication protocol and the MoveIt! motion planning library \cite{Gorner2019Moveit} that generates a velocity trajectory from the skill parameters to control the real robot.}
    \label{fig:framework}
\end{figure*}

Secondly, based on the proposed simulator, we present a novel framework, named differentiable digging robot (DDBot) (Figure~\ref{fig:framework}), which allows first-order gradient-based optimisation in system identification tasks and small-scale high-precision digging tasks for real-world granular materials with unknown properties. We experimentally show that gradient-clipping is effective in stopping gradient explosion in SVK and DP-based granular dynamics. \rps{Video illustration is available on \href{https://www.youtube.com/watch?v=eoNx5V688H0}{YouTube}. Codes are publicly available on \href{https://github.com/IanYangChina/DDBot}{Github}.}

\rp{DDBot uses a differentiable digging skill to significantly reduce the solution space and improve manipulation quality, initialises manipulation solution before optimisation via target-oriented demonstrations to accelerate convergence, and performs line search-based gradient descent with adaptive step sizes to stabilise convergence.}

\rp{Last, we experimentally confirm the feasibility of differentiable system identification with soil and sand materials. For digging skill optimisation, we evaluate and benchmark, for the first time, the performances on a series of soil and sand digging tasks with state-of-the-art manipulation optimisation methods, confirming the robustness and efficiency of our DDBot system. Evaluation on a real-world UR5 robot arm is performed to confirm the practicality of the optimised digging motions.}

The rest of this paper is structured as follows. Section~\ref{sec:rw} discusses existing literature related to this work. Section~\ref{sec:method} presents our method in detail. Section~\ref{sec:expdesign} presents the experiment design and summarises the key questions for investigation. Section~\ref{sec:result-grad} addresses the unstable gradients and rugged loss landscapes. Section~\ref{sec:result-sysid} discusses the results of system identification, followed by the results of digging skill optimisation in Section~\ref{sec:result-skill}. Section~\ref{sec:ablation} presents the performance benchmark and ablation study, and Section~\ref{sec:conclude} concludes the paper.

\section{RELATED WORK}\label{sec:rw}
\subsection{Robotic soil manipulation tasks}
A vast majority of robotic digging research focuses on soil excavation tasks in the construction context, where a large amount of soil, sand, and rocks are to be removed~\cite{stentz1999robotic,jud2017planning,egli2022soil,aoshima2024examining}. An early work presented an autonomous truck-loading system that searches for a starting location of the bucket of a given digging motion to fill the bucket with soils~\cite{stentz1999robotic}. More recently, methods have been proposed to tackle such excavation tasks with uncertain soil properties and unpredictable soil-bucket interaction forces via hierarchical optimisation~\cite{jud2017planning} and reinforcement learning~\cite{egli2022soil}. Another work also studied the sim-to-real gap for soil loading tasks and demonstrated the possibility of reaching low sim-to-real differences with the discrete element method (DEM) for soils with known and uniform physical properties~\cite{aoshima2024examining}.

In agriculture, heavy machines play a key role in large-scale ploughing, seed-sowing, and harvesting tasks but are typically equipped with tailored soil-digging tools and carried out in a large farming field by a human operator~\cite{rondelli2022review}, while agricultural robots for planting or seed-sowing are mostly designed as large vehicles operating in large outdoor fields~\cite{aravind2017task,agriengineering6030166}. In terms of indoor and small-scale soil-related manipulations, such as greenhouses and household gardening tasks, not much effort has been spent on automated soil manipulation with precision, flexibility, and adaptability.

Beyond agriculture, soil manipulation is also required to be automated for soil sampling tasks, either for space exploration~\cite{wang2024advances} or determining farming field soil properties~\cite{joe2024soilsampling}. Though these operations are done in a much more constrained space, they are typically carried out by specialised tools, such as drills, that do not consider changing soil properties nor require flexible and high-precision control.

\rp{The disadvantages of these approaches are primarily twofold. First, they are restricted to a single manipulation motion, inapplicable to ever-changing task requirements and unknown soil properties. Second, they were developed for large-scale operations without the need to satisfy high-precision small-scale control, which is heavily challenged by the difficulty of precisely modelling the physical processes of the materials.
Against this background, we develop a robotic granular material manipulation method that deals with \textit{high-precision} small-scale operations in unstructured environments with \textit{changing task requirements} (varying hole shapes) and \textit{unknown material properties}}.

\subsection{Robotic granular material manipulation methods}
After several decades, robotic manipulation methods have evolved from hand-crafted motions and simplified model-based motion planning~\cite{dong2023review} to data-driven (learning-based) approaches~\cite{kroemer2021review} and differentiable physics-based trajectory optimisation~\cite{yin2021modeling,newbury2024review}. Our focus is on granular material manipulation, whose numbers of dimensions cause classic motion planning methods to struggle, and only a few works in recent years have dealt with such materials.

For granular materials, effort has been devoted to simplifying the representation and dynamics.
\citeauthor*{zhang2020learning}~\cite{zhang2020learning} trained a reinforcement learning (RL) policy to shape granular materials on a table represented by at most $200$ particles in a simulator powered by position-based dynamics (PBD). Due to the costs of high-fidelity simulations, \citeauthor*{mateo2020manipulation}~\cite{mateo2020manipulation} studied a granular material removal task by simplifying the dynamics into a 2D shape control problem represented by a Gaussian mixture model and optimising the control velocity in the task space to achieve a target shape. \citeauthor*{tuomainen2022manipulation}~\cite{tuomainen2022manipulation} employed a graph neural network to estimate the particle states of granular material being poured out from a cup and plan trajectories for different desired pouring outcomes. Their training data was, however, collected using Taichi and MPM-based simulation with about $1400$ particles. \citeauthor*{xue2023neural}~\cite{xue2023neural} proposed to avoid the costly computation of the physics-based interactions by using a fully convolutional neural network to achieve differentiable rendering and thus gradient-based trajectory optimisation for tabletop granular material gathering tasks. \citeauthor*{zhu2023fewshot}~\cite{zhu2023fewshot} proposed a meta-learning approach to predict the scooped volume of different granular materials of a scoop action from RGB-D image patches and used Bayesian optimisation to find the parameters of a simplified scoop motion that maximises the scooped volume. Another recent work also presented a framework to train RL policies for granular material transportation tasks using demonstrations collected by gradient-based trajectory optimisation in a differentiable fluid simulation environment to bypass gradient explosion~\cite{wei2024automachef}.

In summary, \rp{there is no existing approach that targets \textit{high-precision} digging tasks, which are challenged by the difficulty of simulating the high-fidelity dynamics of the target granular material with \textit{unknown properties}. This paper fills this gap by studying such tasks. Method-wise, existing approaches are inefficient with high-fidelity physics dynamics and have to trade off manipulation precision to increase computation efficiency. On the contrary, our novel gradient-based digging skill optimisation method can maintain both high manipulation precision and computational efficiency through a GPU-accelerated parallel computing-supported simulator.}

\section{METHOD}\label{sec:method}
This section presents details of our method for robotic soil/sand digging. An overview of the proposed DDBot system is given in Figure~\ref{fig:framework}. \rp{The primary novelties of the DDBot system are threefold: 1) the use of GPU-accelerated, parallel computing-supported, and MPM-based physics simulation for granular material dynamics that is both time-efficient and physically precise, 2) the complete differentiability of the physics simulator that allows efficient gradient descent algorithms to optimise physics parameters and manipulation skills/trajectories, and 3) a task-oriented digging demonstration method that provides initial solutions to accelerate skill optimisation. With the use of gradient-clipping and line search-based gradient update, DDBot can accomplish highly precise granular digging tasks in $\sim5$ minutes. The rest of this section starts by mathematically formalising the digging task and the proposed digging skill, introducing the skill demonstration, discussing the differentiability of the system components, and explaining how to stop gradient explosion and converge over rugged losses.}

\subsection{Problem definition}\label{subsec:problem}
We adopt the partially observable goal-conditioned Markov decision processes (POGCMDPs) to mathematically describe the soil/sand digging problem, in which a robot takes actions to interact with the environment to accomplish a task goal. Formally, the POGCMDPs comprise a set of observations $o\in\mathcal{O}$, a set of system states $s\in\mathcal{S}$, a set of goals $g\in\mathcal{G}$, a set of actions $a\in\mathcal{A}$, an initial state distribution $p(s_0)$, a (deterministic) system dynamics function $s'=f(s, a): \mathcal{S}\times\mathcal{A}\rightarrow\mathcal{S}$, a goal-conditioned reward function, i.e., a loss function $r(s,a,\hat{g})$, where $\hat{g}\in\hat{\mathcal{G}}\subset\mathcal{G}$ is a desired goal and $\hat{}$ denotes \textit{desired} or \textit{target}, and a discount factor $\gamma$. 

\textbf{States and observations}: In our problem, we focus on a real-world digging environment, where a UR5e robot arm equipped with a 3D-printed shovel is tasked to dig holes in a container full of soil/sand. While we assume a system state $s\in\mathcal{S}$ that comprises all the soil/sand particles' positions, velocities, and accelerations in the world frame (in both reality and simulation), realistically the robot can only observe the surface point clouds (Cartesian points) of the soil/sand obtained from a 3D optical camera. In other words, $\mathcal{O}=\{\pmb{x}|\pmb{x}\in\mathcal{X}_{surface}\}$ where $\mathcal{X}_{surface}\subset\mathbb{R}^3$ is the set of observed surface points. For clearer reading, we overload $o$ to denote both states and observations in the following text and use $\pmb{x}$ to denote Cartesian coordinates.

In practice, the surface points in our experiments are generated by transforming the original point clouds from the camera frame to the world frame, cropped within the area of the container and downsampled with a $40\times40$ grid of size $0.24$, centred at the centre of the soil/sand container. In the simulation, this observation is generated by selecting the highest particles in a $40\times40$ square grid of size $0.24$ meters, centred at the centre of the soil/sand container. Both observations contain $1600$ particles.

The initial state distribution of our problem is defined as the soil/sand being flattened in the container at a certain height ($0.07$ meters). In the simulation, the particle body is created by filling a $0.28\times0.28\times0.07$ box with a fixed density ($5\times10^6$ particles per m$^3$), which results in $27440$ simulated particles. The radius of the particle is determined by dividing the box by the number of particles created ($r_p=2\times10^{-7}$m$^2$). Manually flattening the soil/sand is required to reset the real-world system to its initial state. The initial state of the robot is defined as the end-effector being positioned on top of the centre of the soil/sand container, with the tip of the shovel touching the surface of the soil/sand and orientated to face its digging side towards the positive $x$ direction.

\textbf{Goal and reward/loss}: For digging tasks, a meaningful dense reward function is difficult to define. Instead, desired goals are much easier to specify. We use the 3D-printed shovel to manually make the soil/sand into different shapes for different tasks (Figure~\ref{fig:task-pcds}). The surface point clouds of the results of these manual operations are then taken as desired goals, i.e., $\hat{\mathcal{G}}=\{\hat{\pmb{x}}|\hat{\pmb{x}}\in\hat{\mathcal{X}}_{surface}\}$. The reward (loss) function is defined as the unnormalised Earth Mover's distance (EMD) between the next observation and a desired goal, irrelevant to the action:

\begin{equation}\label{eq:emd}
\begin{split}
r_{EMD}(o',\hat{g}) & = d_{EMD}(\mathcal{X}, \hat{\mathcal{X}}) \\
& = \sum_{x\in\mathcal{X}}\min_{\phi^{EMD}:\mathcal{X}\rightarrow\hat{\mathcal{X}}}||x-\phi(x)||_2
\end{split}
\end{equation}

\noindent where $\phi^{EMD}$ denotes a one-to-one injective mapping that only exists when $|\mathcal{X}|\leq|\hat{\mathcal{X}}|$. In practice, a linear assignment algorithm is used to calculate $\phi^{EMD}$. 

In addition to the EMD distance, we also define a height-map-based distance (HMD) function as an alternative reward/loss. In other words, the desired goal can also be represented by a target height map $\hat{\mathcal{I}}$. Given the resultant height map, $\mathcal{I}$, the height-map-based reward is defined as the summed pixel difference:

\begin{equation}\label{eq:hm}
r_{HMD}(o',\hat{g}) = d_{HMD}(\mathcal{I}, \hat{\mathcal{I}}) = \sum_i\sum_j ||\mathcal{I}_{ij}-\hat{\mathcal{I}}_{ij}||_2
\end{equation}

\noindent where $i$, $j$ are pixel indexes. A height map is generated by traversing all points/particles, computing their referenced pixel indices from their $x$ and $y$ coordinates with an offset of the particle radius $r_p$. This links each particle to $5$ pixel indices corresponding to $(x,y);(x+r_p, y);(x-r_p, y);(x,y+r_p);(x,y-r_p)$. The height map is initialised to zero, and the pixel values are replaced with $z$ values of the particles when they are higher than the existing values. This avoids empty pixels in height map generation due to point sparsity.

\textbf{Actions:} An action of the robot manipulator at each step here is a 6D displacement vector in the Cartesian space (i.e., end-effector tip frame) $a=[\Delta x, \Delta y, \Delta z, \Delta a, \Delta b, \Delta c]\in\mathcal{A}\subset\mathbb{R}^6$, \rp{i.e., it controls the amount of translational changes in the $x$, $y$, $z$ directions and the rotational changes about these three axes}. In the real world, motion planning (MOVEIT! and RRT connect~\cite{kuffner2000rrt,sucan2012ompl,Gorner2019Moveit}) are used to compute a velocity trajectory from a series of Cartesian space waypoints (a trajectory of actions) to control the actual robot arm. In the simulation, the Cartesian displacement for each global step is divided evenly by the number of simulation substeps and applied to the pose of the end-effector before collision detection at each substep.

\textbf{Parameterised skill}: For long-horizon manipulation tasks, controlling the 6D displacements in the Cartesian space is difficult due to the continuous high-dimensional solution space compounded over time. Specifically, a trajectory of $T$ timesteps consists of $6T$ continuous variables, where $T$ usually is greater than $300$ in robot manipulation tasks with a simulation step size of $0.01$s. 
Therefore, we employ temporal abstraction over the trajectory space by introducing a parameterised skill, denoted as $\bar{a}(\pmb{\theta})=\{a_0, a_1, ..., a_T\}$, where $\pmb{\theta}\in[-1,1]^{|\pmb{\theta}|}$ is a set of continuous parameters for calculating the action trajectory of the skill and the length $T$ is determined by a skill-to-action mapping, $\bar{a}(\pmb{\theta})$, which is task-specific (detailed in subsection~\ref{subsec:skill}). 
Although multiple skills can be defined, one skill is sufficient in this work. 
This leads to a one-step POGCMDP whose dynamics function can be written as $o' = f(o, \bar{a})$.

\textbf{System dynamics}: The dynamics function (or transition probability distribution in stochastic cases) determines the next state of the system given the current state and a given action. Obtaining a perfect function to model such real-world dynamics accurately is nearly impossible. However, it can be approximated using numerical modelling methods. \rp{In our work, we adopt} the moving least squares material point methods (MLS-MPM) to simulate granular materials~\cite{hu2018moving}, coupled with the St. Venant-Kirchhoff (SVK) constitutive model for hyperelasticity and the Drucker-Prager (DP) yield criterion for sand/soil plasticity~\cite{drucker1952soil,klar2016drucker}.

MPM represents the soil/sand body as a collection of particles with positions, velocities, deformation gradients, and their material properties, including Young's modulus, Poisson's ratio, density, and sand friction angle. It efficiently integrates the system's velocity over time and handles collision and fracture on a background Eulerian grid~\cite{klar2016drucker}. MLS-MPM adopts a novel weak form discretisation of the governing equations and replaces the shape functions in the force computation with MLS approximators, achieving faster and more realistic simulation and two-way coupling with rigid objects that classic MPM fails~\cite{hu2018moving}. 

\rp{For the sake of clarity, we present a simplified procedure of MLS-MPM, the key equations of the constitutive models (hyperelasticity and plasticity) and a brief description of the collision contact process below. Please refer to~\cite{klar2016drucker,hu2018moving,xy2024sysid} for detailed illustrations of the dynamic model.} 

\rp{As an explicit simulation algorithm, MLS-MPM advances simulation time in a global/local time-stepping process~\cite{hu2018moving}. In other words, for the $t$-th simulation step ($0 < t\leq T$) with a simulation stepsize $\Delta t$, there are $N$ substeps with a sub-stepsize $\Delta t_{sup}=\Delta t / N$. Denote $\pmb{x}^{n+1}_p = f_{sub}^{t,n}(\pmb{x}^n_p, a_t, \pmb{\Theta}, \pmb{g}, \Delta t_{sup})$ as the function of the $n$-th substep of the $t$-th step, it can be summarised as the list of consecutive functions shown in Algorithm 1.}

\begin{table}[b]
\small
\centering
\begin{tabular}{F}
\toprule
\textbf{Algorithm 1}: A simplified MLS-MPM substep procedure ($f^{t,n}_{sub}$)\\
\midrule
1.\ \ For each particle $p$:\\
2.\ \ \ $|$\ \ $F^{tmp}_p = (I + \Delta t_{sup} C_p) F_p$\\
3.\ \ \ $|$\ \ $U_p, S_p, V_p = f_{svd}(F^{tmp}_p)$\\
4.\ \ \ $|$\ \ $F'_p, \sigma_p = f_{constitutive}(U_p, S_p, V_p, \pmb{\Theta})$\\
5.\ \ \ $|$\ \ $\Delta \pmb{v}_{grid} = f_{p2g}(\pmb{x}_p, \pmb{v}_p, \sigma_p, C_p, \Delta t_{sup})$\\
6.\ \ \ $|$\ \ $\pmb{v}_{grid} +=\Delta \pmb{v}_{grid}$\\
7.\ \ $\pmb{x}'_{agent}, \pmb{q}'_{agent} = f_{move\_agent}(a_t, \Delta t_{sup})$\\
8.\ \ $\pmb{v}'_{grid} = \pmb{v}_{grid} + \Delta t_{sup}\cdot\pmb{g}$\\
9.\ \ $\pmb{v}'_{grid} = f_{collision}(\pmb{x}'_{agent}, \pmb{q}'_{agent}, \pmb{x}'_{container}, \pmb{v}'_{grid}, \Delta t_{sup})$\\
10. For each particle $p:$\\
11. \ $|$\ \ $\pmb{v}'_p, C'_p = f_{g2p}(\pmb{v}'_{grid}, \Delta t_{sup})$\\
12. \ $|$\ \ $\pmb{v}'_p = f_{collision}(\pmb{x}'_{agent}, \pmb{q}'_{agent}, \pmb{x}'_{container}, \pmb{x}_p, \pmb{v}'_p, \Delta t_{sup})$\\
13. \ $|$\ \ $\pmb{x}'_p = \pmb{x}_p + \Delta t_{sup}\cdot\pmb{v}'_p$\\
\bottomrule
\footnotesize $F_p$: deformation gradient; $C_p$: affine matrix; $U,S,V$: results of singular value decomposition; $\pmb{\Theta}$: Material parameters; $\sigma_p$: Cauchy stress; $'$ denotes next substep; $a$: agent action; $\pmb{g}$: gravity vector; $\pmb{x}, \pmb{q}, \pmb{v}$: position, rotation and velocity vectors.
\end{tabular}
\end{table}

\rp{In Algorithm 1, line 4, the constitutive model $f_{constitutive}$ first alters the deformation gradient $F^{tmp}_p$ based on the plasticity model for any particle that undergoes plastic deformation. Then it computes the stress of that particle using the altered deformation gradient $F_p'$ and the hyperelasticity energy model. With the singular value decomposition (line 3 of Algorithm 1) of the deformation gradient for particle $p$ after the last iteration update, the SVK elastic energy density function can be written in terms of the diagonal matrix $S_p$:}

\rp{\begin{equation}\label{eq:SVK}
\psi_p(F) = \mu \mathrm{tr}((\ln S_p)^2) + \frac{1}{2}\lambda ( \mathrm{tr}(\ln S_p))^2
\end{equation}}

\noindent \rp{whose derivative is the first Piola–Kirchhoff stress tensor:}

\rp{\begin{equation}\label{eq:piola}
\bm{P}_p = \frac{\partial\psi_p}{\partial F} = U_p\ (2\mu S^{-1}_p\ln S_p + \lambda \mathrm{tr}(\ln S_p)S^{-1}_p)\ V^T_p
\end{equation}}

\noindent \rp{where $\mu$ and $\lambda$ are the Lam\'e constants and can be derived from the Young's modulus $E$ and Poisson's ratio $\nu$:
\begin{equation}
\mu = \frac{E}{2(1+\nu)} \quad \lambda = \frac{E \nu}{(1+\nu)(1-2\nu)}
\end{equation}}

\rp{Plasticity is most conveniently controlled by altering $S_p$. The DP yield criterion states the following. Let $\pmb{\epsilon}_p = \ln S'_p$, and
\begin{equation}
\hat{\pmb{\epsilon}_p} = \pmb{\epsilon}_p - \frac{\mathrm{tr}(\pmb{\epsilon}_p)}{dim}I_{dim} \quad \delta_{\gamma_{p}} = \left\Vert \hat{\pmb{\epsilon}_p} \right\Vert + \frac{\lambda dim+2\mu}{2\mu}\mathrm{tr}(\pmb{\epsilon}_p) \alpha_f
\end{equation}}

\noindent \rp{where $\alpha_f = \sqrt{\frac{2}{3}}\frac{2\sin{\phi_{f}}}{3-\sin{\phi_{f}}}$, $dim$ is the spatial dimension of the problem ($3$ in our case), $I_{dim}$ denotes the identity matrix of dimension $dim$, $\delta_{\gamma_{i}}$ is the amount of plastic deformation, and $\phi_{f}$ is the friction angle of sand/soil. The DP model alters $S_p$ under different conditions as follows:
\begin{equation}
    \hat{S}_p=
    \begin{cases}
    I_{dim} \hfill & \mathrm{tr}(\pmb{\epsilon}_p) > 0,\\
    S_p \hfill & \mathrm{tr}(\pmb{\epsilon}_p) \leq 0 \And \delta \gamma_{p} \leq 0, \\
    \exp(\pmb{\epsilon}_p-\delta \gamma_{p}\frac{\hat{\pmb{\epsilon}_p}}{\|\hat{\pmb{\epsilon}_p}\|}) \hfill & \mathrm{tr}(\pmb{\epsilon}_p) \leq 0 \And \delta \gamma_{p} > 0.
    \end{cases}\label{eq:DP}
\end{equation}}

\rp{Please refer to~\cite{klar2016drucker} for details of the three conditions. With $\hat{S}_p$, the deformation gradient after plasticity is $F_p' = U_p \hat{S}_p V^{T}_p$ and the Cauchy stress $\sigma_p = \frac{1}{det(F'_p)}\bm{P}_p'{F'_p}^T$, where $\bm{P}_p'$ is computed by substituting $\hat{S}_p$ into Eq.~\ref{eq:piola}, and $\det(F'_p) = 1$ is imposed to enforce incompressibility.}

\rp{Given an action (Cartesian space displacement) at time $t$, the 6D Cartesian pose of the manipulation (agent) is updated by evenly distributing the action over the $N$ substeps and adding it to the manipulator pose at each substep (line 7 in Algorithm 1). The collision behaviours between the particles/grid and the manipulator are modelled by a signed distance field-based collision detection mechanism and the Columb frictional contact model, following the practices of previous works~\cite{hu2018moving,xy2024sysid}.}

\subsection{Digging skill design and its differentiability}\label{subsec:skill}
This subsection presents the details of a parameterised digging skill. It produces a variable-length trajectory: $\bar{a}(\pmb{\theta})=\{a_0, a_1,...,a_T\}$, 
where $\pmb{\theta} = \{\theta_{displace}, \theta_{rotate}, \theta_{insert\_dist}, \theta_{push\_angle}, \theta_{push\_dist}\}$. 

\begin{figure}[t]
\centering
\includegraphics[width=\columnwidth]{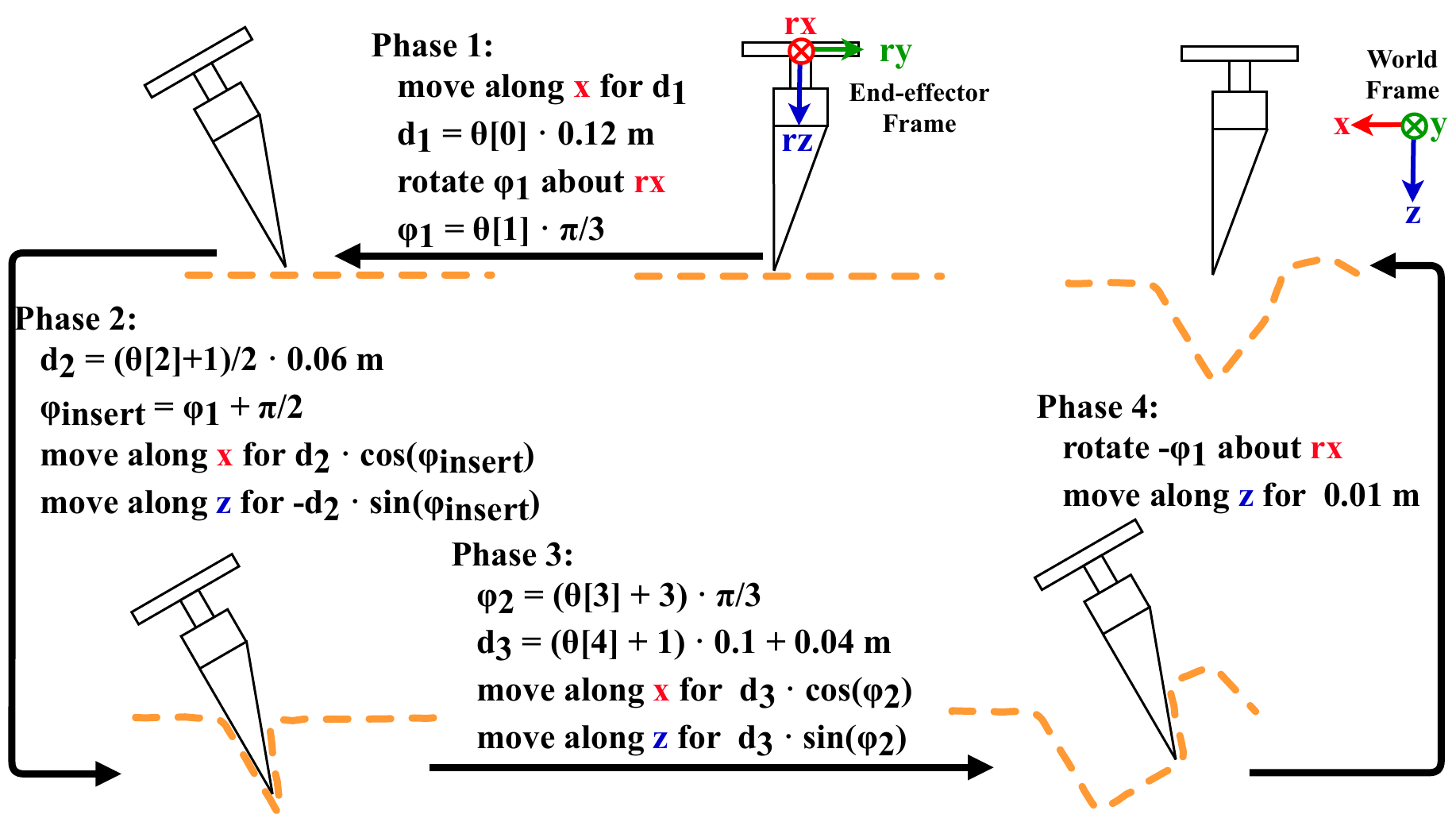}
\caption{Visualising the digging skill modelled by Eqs.~\ref{eq:d1} to~\ref{eq:delta_d4}. Orange dashed line: soil or sand.}\label{fig:skill-vis}
\end{figure}

\rp{The skill comprises four phases graphically illustrated by Figure~\ref{fig:skill-vis} and mathematically defined by Eqs.~\ref{eq:d1} to ~\ref{eq:delta_d4}, with a list of notations comprised in Table~\ref{tab:skill-notation}. Algorithm 2 presents the assembly of the calculated end-effector displacements into the complete trajectory $\bar{a}(\pmb{\theta})$.}

\rp{\textbf{Phase 1:} moving the shovel in the world's $x$ axis by a distance $d_1$ and rotating about the shovel's $x$ axis by an angle $\varphi_1$. Eqs.~\ref{eq:d1} to ~\ref{eq:delta_d1} define the calculation of the actions for phase 1 from $\theta_{displace}$ and $\theta_{rotate}$.}

\begingroup\makeatletter\def\f@size{8}\check@mathfonts\rp{\vspace{-0.3cm}
\begin{flalign}
    &d_1 = \theta_{displace} \cdot 0.12, &&\varphi_1 = \theta_{rotate} \cdot \pi / 3\label{eq:d1}\\
    &T^{float}_{d_1} = |d_1 / (v_l\cdot\Delta t)|, &&T^{float}_{\varphi_1} = |\varphi_1/(v_w\cdot\Delta t)|\label{eq:Tf1}\\
    &T^{float}_1=max(T^{float}_{d_1}, T^{float}_{\varphi_1}), &&T_1 = round(T^{float}_1)\label{eq:Ti1}\\
    &\Delta x_1 = \begin{cases}
        \frac{d_1}{T_1},\ T_1>0\\
        0,\  T_1=0
        \end{cases}\hspace{-0.4cm},
    &&\Delta rx_1 = \begin{cases}
        \frac{\varphi_1}{T_1},\ T_1>0\\
        0,\  T_1=0
        \end{cases}\label{eq:delta_d1}
\end{flalign}}
\endgroup

\begin{table}[h]
\small
\centering
\begin{tabular}{lE}
\toprule
\textbf{Symbol} & \textbf{Meaning} \\
\midrule
$\theta_{\rm displace}$       & parameter: translational displacement in the world's $x$ direction \\
$\theta_{\rm rotate}$         & parameter: rotational displacement about $rx$ (end-effector's $x$ direction) \\
$\theta_{\rm insert\_dist}$   & parameter: insertion distance \\
$\theta_{\rm push\_dist}$     & parameter: pushing distance \\
$\theta_{\rm push\_ang}$      & parameter: pushing angle \\
$d_i$                         & translational displacement in phase $i$ (m) \\
$\varphi_i$                   & rotational displacement in phase $i$ (rad) \\
$v_l$                         & end‐effector linear speed (m/s) \\
$v_w$                         & end‐effector angular speed (rad/s) \\
$\Delta t$                    & simulation time stepsize\\
$T^{float}_i$                 & real‐valued timesteps required in phase $i$\\
$T_{i}$                       & integer timesteps used in phase $i$ \\
$max()$                       & returning the maximal value among the inputs\\
$round()$                     & returning the rounded value of an input\\
$exp()$                       & returning the exponential of the inputs\\
$det()$                       & returning the determinant of the inputs\\
$\Delta x_i,\Delta z_i$       & per‐step translations along $x,z$ axes for phase $i$ (m) \\
$\Delta rx_i$                 & per‐step rotation about $rx$ in phase $i$ (rad) \\
$\bar{a}[i:j]$                & the $i$-th (inclusive) to the $j$-th (exclusive) actions of the skill\\
$a_t[i]$                      & the $i$-th element of the $t$-th action of the skill\\
\bottomrule
\end{tabular}
\caption{Notations used in the digging skill definition and its derivatives (Eqs.~\ref{eq:d1} to ~\ref{eq:da-dtheta2-t2-unrounded}).}\label{tab:skill-notation}
\end{table}

\rp{\textbf{Phase 2:} inserting the shovel into the soil/sand by a distance $d_2$ along its pointing direction. Eqs.~\ref{eq:d2} to ~\ref{eq:delta_d2} define the calculation of the actions for phase 2 from $\theta_{rotate}$ and $\theta_{insert\_dist}$.}

\begingroup\makeatletter\def\f@size{8}\check@mathfonts\rp{\vspace{-0.3cm}
\begin{flalign}
    &d_2 = (\theta_{insert\_dist}+1)\cdot0.03,\ \varphi_2 = \varphi_1 + \pi /2 &&\label{eq:d2}\\
    &T^{float}_2 = d_2 / (v_l\cdot\Delta t)\label{eq:Tf2}\\
    &T_2 = round(T^{float}_2)\label{eq:Ti2}\\
    &\Delta x_2 = \begin{cases}
        \frac{d_2 \cdot \cos(\varphi_{2})}{T_2},\ T_2>0\\
        0,\  T_2=0
        \end{cases},\ 
    \Delta z_2 = \begin{cases}
        \frac{-d_2 \cdot \sin(\varphi_{2})}{T_2},\ T_2>0\\
        0,\  T_2=0
        \end{cases}\hspace{-1cm}&&\label{eq:delta_d2}
\end{flalign}}
\endgroup

\rp{\textbf{Phase 3:} pushing towards an angle $\varphi_3$ by a distance $d_3$. Eqs.~\ref{eq:d3} to~\ref{eq:delta_d3} define the calculation of the actions for phase 3 from $\theta_{push\_dist}$ and $\theta_{push\_angle}$}.
\begingroup\makeatletter\def\f@size{8}\check@mathfonts\rp{\vspace{-0.3cm}
\begin{flalign}
    &d_3 = (\theta_{push\_dist} + 1) \cdot 0.1 + 0.04,\ \varphi_3 = (\theta_{push\_angle} + 3) \cdot \pi /3\hspace{-0.4cm}&&\label{eq:d3}\\
    &T^{float}_3 = |d_3 / (v_l\cdot\Delta t)|&&\label{eq:Tf3}\\
    &T_3 = round(T^{float}_3)&&\label{eq:Ti3}\\
    &\Delta x_3 = \begin{cases}
        \frac{d_3 \cdot \cos(\varphi_3)}{T_3},\ T_3>0\\
        0,\  T_3=0
        \end{cases},\
    \Delta z_3 = \begin{cases}
        \frac{d_3 \cdot \sin(\varphi_3)}{T_3},\ T_3>0\\
        0,\  T_3=0
        \end{cases}\hspace{-1cm}&&\label{eq:delta_d3}
\end{flalign}}
\endgroup

\rp{\textbf{Phase 4:} rotating back to straight and lifting for a constant distance $d_4=0.01$ m. Eqs.~\ref{eq:Tf4} to~\ref{eq:delta_d4} define the calculation of actions for phase 4.}
\begingroup\makeatletter\def\f@size{8}\check@mathfonts\rp{\vspace{-0.3cm}
\begin{flalign}
    &T^{float}_{d_4} = d_4 / (v_l\cdot\Delta t) &&&\label{eq:Tf4}\\
    &T^{float}_4 = max(T^{float}_{\varphi_1}, T^{float}_{lift}), &&T_4 = round(T^{float}_4)&&&\label{eq:Ti4}\\
    &\Delta rx_4 = \frac{- \varphi_1}{T_4},&&\Delta z_4 = \frac{d_4}{T_4}&&&\label{eq:delta_d4}
\end{flalign}}
\endgroup

\begin{table}[t]
\small
\centering
\begin{tabular}{F}
\toprule
\textbf{Algorithm 2}: Skill-to-action mapping $\bar{a}(\pmb{\theta})=\{a_0,..,a_T\}$\\
\midrule
Inputs: $\theta_{ displace},\theta_{ rotate},\theta_{ insert\_dist},\theta_{ push\_dist},\theta_{ push\_angle}$\\
$d_{ lift}, v_l,v_w,\Delta t$. \\
Output: $\bar{a}=\{a_0,a_1,...,a_T\}$\\
\midrule
With Eqs.~\ref{eq:d1} to~\ref{eq:delta_d4}, Compute: \\
$T_1$, $T_2$, $T_3$, $T_4$, $\Delta x_1$, $\Delta rx_1$, $\Delta x_2$, $\Delta z_2$, $\Delta x_3$, $\Delta z_3$, $\Delta rx_4$, $\Delta z_4$\\
Let $T =T_1+T_2+T_3+T_4$\\
For $t$ from $0$ to $T$: \hspace{0.73cm} $a_t = [0.0, 0.0, 0.0, 0.0, 0.0, 0.0]$\\
For $t$ from $0$ to $T_1-1$: \hspace{0.07cm} $a_t[0] = \Delta x_1$, $a_t[3]=\Delta rx_1$\\
For $t$ from $T_1$ to $T_2-1$: $a_t[0] = \Delta x_2$, $a_t[2]=\Delta z_2$\\
For $t$ from $T_2$ to $T_3-1$: $a_t[0] = \Delta x_3$, $a_t[2]=\Delta z_3$\\
For $t$ from $T_3$ to $T_4-1$: $a_t[3] = \Delta rx_4$, $a_t[2]=\Delta z_4$\\
\bottomrule
\end{tabular}
\end{table}

\rp{For each phase, the number of timesteps is calculated by dividing the total displacements over a given linear or angular velocity times $\Delta t$ and rounding it into an integer. As the rounding operation (Eqs.~\ref{eq:Ti1},~\ref{eq:Ti2},~\ref{eq:Ti3} and~\ref{eq:Ti4}) is discontinuous and thus not differentiable, using it to calculate the actions will lead to incomplete gradients. In other words, the changes in actions can only backpropagate through the computation of the 6D displacements for each action, but not through the computation of the number of timesteps (e.g., in phase 1, $\frac{\partial\Delta x_1}{d_1}\neq0$ while $\frac{\partial\Delta x_1}{T_1}=0$).}

\rp{An alternative is to assign the actions by dividing the total displacement over the pre-rounded number of simulation steps (replacing $T_i$ with $T^{float}_i$). However, in phase 2, $\Delta x_2$ and $\Delta z_2$ become only related to $\varphi_2$ with this alternative (i.e., $\Delta x_2=\frac{d_2 \cdot \cos(\varphi_{2})}{T^{float}_2}=v_l\cdot\Delta t\cdot\cos(\varphi_2)$), which renders the gradient of $\theta_{insert\_dist}$ effectively zero. In phase 3, this also leads to zero gradients for $\theta_{push\_dist}$. To avoid this, $T_2$ and $T_3$ (the integer versions) are retained for this alternative version of the skill-to-action mapping.}

\rp{We present a complete derivation of the gradient chains for $\frac{\partial\bar{a}}{\partial\pmb{\theta}}$ in Appendix A. Ablation experiments are conducted to examine the effect of using an unrounded number of timesteps for per-action displacement calculation in phases 1 and 4. The derived gradients of $\pmb{\theta}$ with rounded numbers of timesteps are as follows.}

\begingroup\makeatletter\def\f@size{8}\check@mathfonts\rp{\vspace{-0.3cm}
\begin{flalign}
    &\frac{\partial \bar{a}}{\partial \theta_{displace}}\Bigg|_{rounded\_T_1}=\frac{\partial \bar{a}[0:T_1]}{\partial \theta_{displace}}\Bigg|_{rounded\_T_1}=0.12&&\label{eq:da-dtheta1-rounded}\\
    &\frac{\partial \bar{a}}{\partial \theta_{rotate}}\Bigg|_{rounded\_T_1T_2T_4}&&\notag\\
    &\ \ \ =\frac{\partial \bar{a}[0:T_1]}{\partial \theta_{rotate}}\Bigg|_{rounded\_T_1}+\frac{\partial \bar{a}[T_1:T_1+T_2]}{\partial \theta_{rotate}}\Bigg|_{rounded\_T_2}&&\notag\\
    &\ \ \ \ \ \ \ \ +\frac{\partial \bar{a}[T_1+T_2+T_3:T_1+T_2+T_3+T_4]}{\partial \theta_{rotate}}\Bigg|_{rounded\_T_4}&&\notag\\
    &\ \ \ =\frac{\pi}{3} -d_2\cdot\frac{\pi}{3}\cdot\left(\sin(\varphi_1+\frac{\pi}{2})+\cos(\varphi_1+\frac{\pi}{2})\right)-\frac{\pi}{3}&&\label{eq:da-dtheta2-rounded}\\
    &\frac{\partial \bar{a}}{\partial \theta_{insert\_dist}}\Bigg|_{rounded\_T_2}=\frac{\partial \bar{a}[T_1:T_1+T_2]}{\partial \theta_{insert\_dist}}\Bigg|_{rounded\_T_2}&&\notag\\
    &\ \ \ =0.03\cdot(\cos\varphi_2-\sin\varphi_2)&&\label{eq:da-dtheta3-rounded}
\end{flalign}}
\endgroup

\begingroup\makeatletter\def\f@size{8}\check@mathfonts\rp{\vspace{-0.3cm}
\begin{flalign}
    &\frac{\partial \bar{a}}{\partial \theta_{push\_dist}}\Bigg|_{rounded\_T_3}=\frac{\partial \bar{a}[T_1+T_2:T_1+T_2+T_3]}{\partial \theta_{push\_dist}}\Bigg|_{rounded\_T_3}&&\notag\\
    &\ \ \ =0.1\cdot\left[\cos(\varphi_3)+\sin(\varphi_3)\right]&&\label{eq:da-dtheta4-rounded}\\
    &\frac{\partial \bar{a}}{\partial \theta_{push\_angle}}\Bigg|_{rounded\_T_3}=\frac{\partial \bar{a}[T_1+T_2:T_1+T_2+T_3]}{\partial \theta_{push\_angle}}\Bigg|_{rounded\_T_3}&&\\
    &\ \ \ =\frac{\pi\cdot d_3}{3}[\cos(\varphi_3)-\sin(\varphi_3)]&&\label{eq:da-dtheta5-rounded}
\end{flalign}}
\endgroup

\rp{If unrounded numbers of timesteps are used (i.e., replacing $T_1$ and $T_4$), the gradients of $\theta_{displace}$ and $\theta_{rotate}$ become:}

\begingroup\makeatletter\def\f@size{8}\check@mathfonts\rp{\vspace{-0.3cm}
\begin{flalign}
    &\frac{\partial \bar{a}}{\partial \theta_{displace}}\Bigg|_{unrounded\_T_1}=\frac{\partial \bar{a}[0:T_1]}{\partial \theta_{displace}}\Bigg|_{unrounded\_T_1}&& \notag\\
    &\ \ \ =\begin{cases}
    T_1\cdot\frac{-\pi\cdot\theta_{rotate}\cdot v_l\cdot\Delta t}{0.36\cdot{\theta_{displace}}^2},\ \ T^{float}_{d_1}\geq T^{float}_{\varphi_1}>0,\ \theta_{displace}>0\\
    T_1\cdot\left(\frac{2\cdot v_l\cdot\Delta t}{\theta_{displace}}+\frac{\pi\cdot\theta_{rotate}\cdot v_l\cdot\Delta t}{0.36\cdot{\theta_{displace}}^2}\right),\\
    \hspace{3cm} T^{float}_{d_1}\geq T^{float}_{\varphi_1}>0,\ \theta_{displace}<0\\
    T_1\cdot\frac{0.12\cdot v_w\cdot\Delta t}{|\theta_{rotate}|\cdot\pi/3},\hspace{0.7cm} 0<T^{float}_{d_1}<T^{float}_{\varphi_1}
    \end{cases}\hspace{-1cm}\label{eq:da-dtheta1-unrounded}&&
\end{flalign}}
\endgroup\vspace{-0.3cm}

\begingroup\makeatletter\def\f@size{8}\check@mathfonts\rp{\vspace{-0.3cm}
\begin{flalign}
    &\frac{\partial \bar{a}}{\partial \theta_{rotate}}\Bigg|_{unrounded\_T_1T_2T_4}&&\notag\\
    &\ \ \ =\frac{\partial \bar{a}[0:T_1]}{\partial \theta_{rotate}}\Bigg|_{unrounded\_T_1}+\frac{\partial \bar{a}[T_1:T_1+T_2]}{\partial \theta_{rotate}}\Bigg|_{rounded\_T_2}&&\notag\\
    &\ \ \ \ \ \ \ \ +\frac{\partial \bar{a}[T_1+T_2+T_3:T_1+T_2+T_3+T_4]}{\partial \theta_{rotate}}\Bigg|_{unrounded\_T_4}&&\label{eq:da-dtheta2-unrounded}
\end{flalign}}
\endgroup\vspace{-0.3cm}

\noindent\rp{where,}

\begingroup\makeatletter\def\f@size{8}\check@mathfonts\rp{\vspace{-0.3cm}
\begin{flalign}
    &\frac{\partial \bar{a}[T_1+T_2+T_3:T_1+T_2+T_3+T_4]}{\partial \theta_{rotate}}\Bigg|_{unrounded\_T_4}&&\notag\\
    &\ \ \ =\begin{cases}
        T_4\cdot\frac{-3\cdot d_4\cdot v_w\cdot\Delta t}{{\theta_{rotate}}^2\cdot\pi}, \ \ T^{float}_{\varphi_1}\geq T^{float}_{lift}>0, \ \theta_{rotate}>0\\
        T_4\cdot\left(\frac{-2\cdot v_w\Delta t}{\theta_{rotate}}+\frac{3\cdot d_4\cdot v_w\cdot\Delta t}{{\theta_{rotate}}^2\cdot\pi}\right),\\
        \hspace{2.45cm} T^{float}_{\varphi_1}\geq T^{float}_{lift}>0, \ \theta_{rotate}<0\\
        T_4\cdot\frac{-\pi\cdot v_l\cdot\Delta t}{3\cdot d_4}, \hspace{0.6cm} 0<T^{float}_{\varphi_1}<T^{float}_{lift}
    \end{cases}\hspace{-3cm}&&
\end{flalign}}
\endgroup\vspace{-0.3cm}

\begingroup\makeatletter\def\f@size{8}\check@mathfonts\rp{\vspace{-0.3cm}
\begin{flalign}
    &\frac{\partial \bar{a}[0:T_1]}{\partial \theta_{rotate}}\Bigg|_{unrounded\_T_1}&&\notag\\
    &\ \ =\begin{cases}
    T_1\cdot\frac{-0.36\cdot\theta_{displace}\cdot v_w\cdot\Delta t}{{\theta_{rotate}}^2\cdot\pi},\\
    \hspace{2.9cm} T^{float}_{\varphi_1}>T^{float}_{d_1}>0,\ \theta_{rotate}>0\\
    T_1\cdot\left(\frac{2\cdot v_w\Delta t}{\theta_{rotate}}+\frac{0.36\cdot\theta_{displace}\cdot v_w\cdot\Delta t}{{\theta_{rotate}}^2\cdot\pi}\right),\\
    \hspace{2.9cm} T^{float}_{\varphi_1}>T^{float}_{d_1}>0,\ \theta_{rotate}<0\\
    T_1\cdot\frac{\frac{\pi}{3}\cdot v_l\cdot\Delta t}{|\theta_{displace}|\cdot0.12},\hspace{0.4cm} 0<T^{float}_{\varphi_1}\leq T^{float}_{d_1}
    \end{cases}\hspace{-3cm}&&\\
    &\frac{\partial \bar{a}[T_1:T_1+T_2]}{\partial \theta_{rotate}}\Bigg|_{rounded\_T_2}&&\notag\\
    &\ \ \ =-d_2\cdot\frac{\pi}{3}\cdot\left(\sin(\varphi_1+\frac{\pi}{2})+\cos(\varphi_1+\frac{\pi}{2})\right)&&\label{eq:da-dtheta2-t2-unrounded}
\end{flalign}}
\endgroup\vspace{-0.3cm}

\subsection{Demonstration and skill prior}
To improve the convergence speed, we rely on human priors to compose demonstrations \rp{to serve as initial guesses for skill optimisation or to guide data collection for learning-based methods}. We take inspiration from the fact that human children learn to manipulate by extrapolating a known skill over different manipulation scenarios and targets. In order words, given a set of skill parameters $\pmb{\theta^-}$, we set $\pmb{\theta^-}$ as the initial skill parameters before running optimisations. The first parameter of the demonstration skill, $\theta_{displace}$, is calculated based on the location of the hole presented in different task target point clouds: 

$$\theta_{displace} = clip(\frac{(\pmb{\hat{\mathcal{X}}}[arg\min_z\pmb{\hat{\mathcal{X}}}][0] - 0.02)}{0.12}, -1.0, 1.0)$$

\noindent where $\pmb{\hat{\mathcal{X}}}$ is the target surface point cloud. The rest of the demonstration skill parameters are constant: $\theta_{rotate}=0.2$, $\theta_{insert\_distance}=0.8$, $\theta_{push\_angle}=0.0$, $\theta_{push\_distance}=-0.5$, which gives a digging trajectory that digs a small hole near the hole presented in the task target.

\subsection{Gradient-based skill optimisation}
In classic trajectory optimisation, the objective is to find a trajectory of $T$ actions: $\tau=\{a_0, a_1, ..., a_T\}$, such that the return is maximised: 
$$\tau^* = arg\max_{\tau}r(o',\hat{g})|_{o'\sim f(o, \tau)}$$

\noindent \rp{where, $r$ is the reward/cost function, $o'$ is the observation of the post-manipulation granular material, and $f(o, a)$ is the system dynamics.} Instead of explicit optimisation for every action in a fixed-length trajectory, we seek to optimise the proposed digging skill that produces a variable-length trajectory with five parameters, resulting in a much smaller solution space for optimisation. The skill optimisation problem with parameters $\pmb{\theta}$ can be formulated as: 

\begin{equation*}
\begin{split}
\max_{\pmb{\theta}} \ & J(\pmb{\theta}) = r(o',\hat{g})\\
s.t. \ & o' = f(o, \tau) \\
& \tau = \bar{a}(\pmb{\theta})\\
& \pmb{\theta} \in [-1, 1]^{|\pmb{\theta}|}
\end{split}
\end{equation*}

\noindent \rp{where $J(\pmb{\theta})$ is the objective function, $\bar{a}(\pmb{\theta})$ is the skill-to-action mapping and $\pmb{\theta}$ is the set of skill parameters.} 

Many methods have been developed for such constrained optimisation problems, yet gradient-descent methods are the most attractive due to faster convergence under the condition of the whole system being differentiable (smooth)~\cite{bonnans2006numerical}. We are seeking to compute $\partial r/\partial\pmb{\theta}$, and apply first-order gradient-descent algorithms to iteratively update $\pmb{\theta}$ towards local minima.
Three key components need to be differentiable (smooth) in computing the gradients of the skill parameters via the chain rule, i.e., the loss (reward) function $r$, the skill-to-action mapping $\bar{a}(\pmb{\theta})$ and the dynamics function $f(o,\bar{a})$:

\begin{equation}
\frac{\partial r}{\partial \pmb{\theta}} = \frac{\partial r(o', \hat{g})}{\partial o'|_{o'=f(o,\bar{a})}}\frac{\partial f(o,\bar{a})}{\partial \bar{a}}\frac{\partial \bar{a}}{\partial \pmb{\theta}}\label{eq:grad-chain}
\end{equation}
 
In this work, the loss function is either the earth mover's distance (Eq.~\ref{eq:emd}) or the height-map-based distance (Eq.~\ref{eq:hm}), which is either the sum of Euclidean distances of pairs of 3D points or the sum of the differences of their $z$ values, whose first-order derivatives \rp{(the first term of the right-hand side of Eq~\ref{eq:grad-chain})} are well-defined. The differentiability of the skill-to-action mapping \rp{(the third term of the right-hand side of Eq~\ref{eq:grad-chain})} was introduced in subsection~\ref{subsec:skill} with a full derivation given in Appendix A.

\textbf{The differentiability of the dynamics:} \rp{Here, we discuss the second term of the right-hand side of Eq~\ref{eq:grad-chain}.} As introduced in subsection~\ref{subsec:problem}, the dynamics function for soil/sand-rigid-body interactions is simulated by the MLS-MPM method~\cite{hu2018moving}. 
The gradient chain of the $t$-th simulation step comprising $N$ substeps from the resultant observation $o_{t+1}$ to the action $a_t$ can be written as follows:

\begin{equation}\label{eq:po/pa}
\begin{split}
\frac{\partial o_{t+1}}{\partial a_t} 
&=\sum_{\pmb{x}_p\in o_{t+1}}\sum_{n=0}^N\frac{df_{sub}^{t,n}(\pmb{x}^{n-1}_p, a_t, \pmb{\Theta}, \pmb{g}, \Delta t_{sup})}{da_t}\\
&= \sum_{\pmb{x}_p\in o_{t+1}}\sum_{n=0}^N\Bigl[
\frac{\partial f_{sub}^{t,n}}{\partial \pmb{x}_p^{n-1}}\cdot
\frac{d f_{sub}^{t,n-1}}{d a_t} 
+ \frac{\partial f_{sub}^{t,n}}{\partial a_t}
\Bigr]    
\end{split}
\end{equation}

Algorithm 1 indicates that, as long as the collision function $f_{collision}$, the grid-to-particle function $f_{g2p}$, the agent position updating function $f_{move\_agent}$, the particle-to-grid function $f_{p2g}$ are differentiable over the particle positions and the agent's actions, one can obtain $\partial o_{t+1}/\partial a_t$ with Eq.~\ref{eq:po/pa} using the chain rule. The derivatives of these functions have been provided in \cite{hu2019chainqueen}. Notice that the constitutive model, $f_{constitutive}$, needs not to be differentiable if only the gradients of actions are of interest. However, our work also seeks to perform system identification that optimises the physics parameters for simulating soil/sand with gradient-based optimisation (see subsection~\ref{subsec:sysid}). Therefore, we further provide proof of the differentiability of the constitutive model in Appendix B.

\rp{Combined with the discussion on differentiable skill in Section~\ref{subsec:skill}, we establish the theoretical foundation for gradient-based digging skill optimisation. The next two subsections will discuss the issues and solutions for exploding gradients and rugged loss landscapes that stop gradient descent from converging.}

\subsection{Stopping gradients from exploding}\label{subsec:grad-explode}
A critical issue that destabilises or even disables gradient-based optimisation with differentiable physics is the numerical instability of gradient computation caused by long task horizons and the large particle number. As mentioned in existing works, gradients tend to explode or vanish as the length of the computation chain (number of actions/simulation steps) or the number of particles increases~\cite{xu2021accelerated,lin2022diffskill}. This effect can be predicted based on Eq.~\ref{eq:po/pa}, where $\partial o_{t+1}/\partial a_t$ depends heavily on the number of simulation steps $T$, substeps $N$, and particles $|o_{t+1}|$. It increases drastically (explodes) when there are too many sub-gradients that are greater than $1$ ($\partial f_{sub}/\partial \pmb{x}_p>1$), and vice versa (vanishes). This problem is exacerbated when the physical model becomes more complex, preventing previous works from using efficient first-order optimisation methods on tasks with long horizons or large numbers of particles~\cite{xu2021accelerated,chen2022diffsrl,lin2022diffskill,xian2023fluidlab}. 

We \rp{experiment on} three techniques to regularise the scale of the gradient for key variables in the MLS-MPM algorithm, namely \textit{clipping, dynamic-scaling, and normalisation}. 
%\rps{A more in-depth theoretical analysis of the gradient explosion issue and the effectiveness of these techniques are provided in Appendix C}

Specifically, denote a variable vector of interest as $\pmb{x}$. The \textit{clipping} operation bounds its gradient in $\nabla \pmb{x}\in[-\nabla_{clip}, \nabla_{clip}]$ where $\nabla_{clip}$ is a user-defined threshold value. 

The \textit{dynamic-scaling} operation reduces the order of magnitude of the whole gradient vector by an integer such that the largest element of the gradient vector is kept below a certain order of magnitude: 
$\nabla \pmb{x} \leftarrow \nabla \pmb{x} / (10^{\Delta oom}+\delta)$
where $\Delta oom=round(\log(max(\nabla \pmb{x})))/\log 10 - oom^*$ and $oom^*$ is the maximal order of magnitude for the gradient defined by the user, and $\delta$ is a small value that prevents dividing by zero. 

The \textit{normalisation} operation is straightforward, which divides a gradient vector by its magnitude (norm), $\nabla \pmb{x}\leftarrow \nabla\pmb{x}/(||\nabla\pmb{x}|| +\delta)$, where $\delta$ is a small value. In our experiments, $\nabla_{clip} = 1e4$, $oom^*=4$ and $\delta=10^{-6}$. 

Except for the material parameters in the SVK and DP models (Young's modulus $E$, Poisson's ratio $\nu$, material density $\rho$ and sand friction angle $\phi_f$) and the actions (skill parameters), the key variables in the MLS-MPM algorithm that play important roles in how the material behaves in response to external forces (actions) include the deformation gradient ($F_p$), position ($\pmb{x}_p$) and velocity ($\pmb{v}_p$) for each particle $p$ and the grid mass $m_{grid}$ and the velocities before and after collision detection $\pmb{v}_{grid}$, $\pmb{v}_{grid}'$. In section~\ref{sec:result-grad}, we examine the vanilla gradient values of these variables and how the proposed techniques confine their scales.
 
\subsection{Handling rugged loss and fluctuating gradients}\label{subsec:grad-fluctuate}

Another cause that hinders efficient gradient descent is a rugged loss landscape, which causes gradients to fluctuate. Such fluctuating gradients lead to the divergence of the optimisation process because they cause the so-called ``\textit{overshooting}'' issue, a phenomenon in gradient-based optimisation processes where the solution is updated with a stepsize that is too large such that it updates the solution too far in the gradient direction and passes over the local minimum, which happens very often with rugged loss landscapes. Unfortunately, because of the use of surface point clouds as the geometric representation of the materials, both the EMD and HMD losses present different levels of ruggedness at different sampling resolutions over the spaces of physics parameters and skill parameters. Typically, without any treatment, we observed that optimisation with a fixed stepsize may encounter a good local minimum but cannot converge to that solution. For example, the red and pink curves in Figure~\ref{fig:sysid} show that vanilla optimisation may reach a validation loss similar to optimisations with a line search technique (see below), but is unstable and exhibits high variance. A detailed examination of the loss landscapes and their ruggedness is given in subsection~\ref{sec:result-grad}.

To address this challenge, we propose to use a simple yet effective technique called \textit{line search}. Specifically, before each gradient update, we evaluate solutions that are updated by $5$ different stepsizes along the gradient direction (by multiplying the stepsizes given in Section~\ref{sec:expdesign} with $\alpha_{ls}\in\{0.1, 0.5, 1.0, 1.5, 2.0\}$). The one that minimises the HMD loss is selected to update the solution, after which the optimisation proceeds to the next iteration. The HMD loss is selected for line search evaluation because of its smoothness. 

\section{EXPERIMENT DESIGN}\label{sec:expdesign}

\subsection{Real-world platform}
\begin{figure}[b]
\centering
\includegraphics[width=\columnwidth]{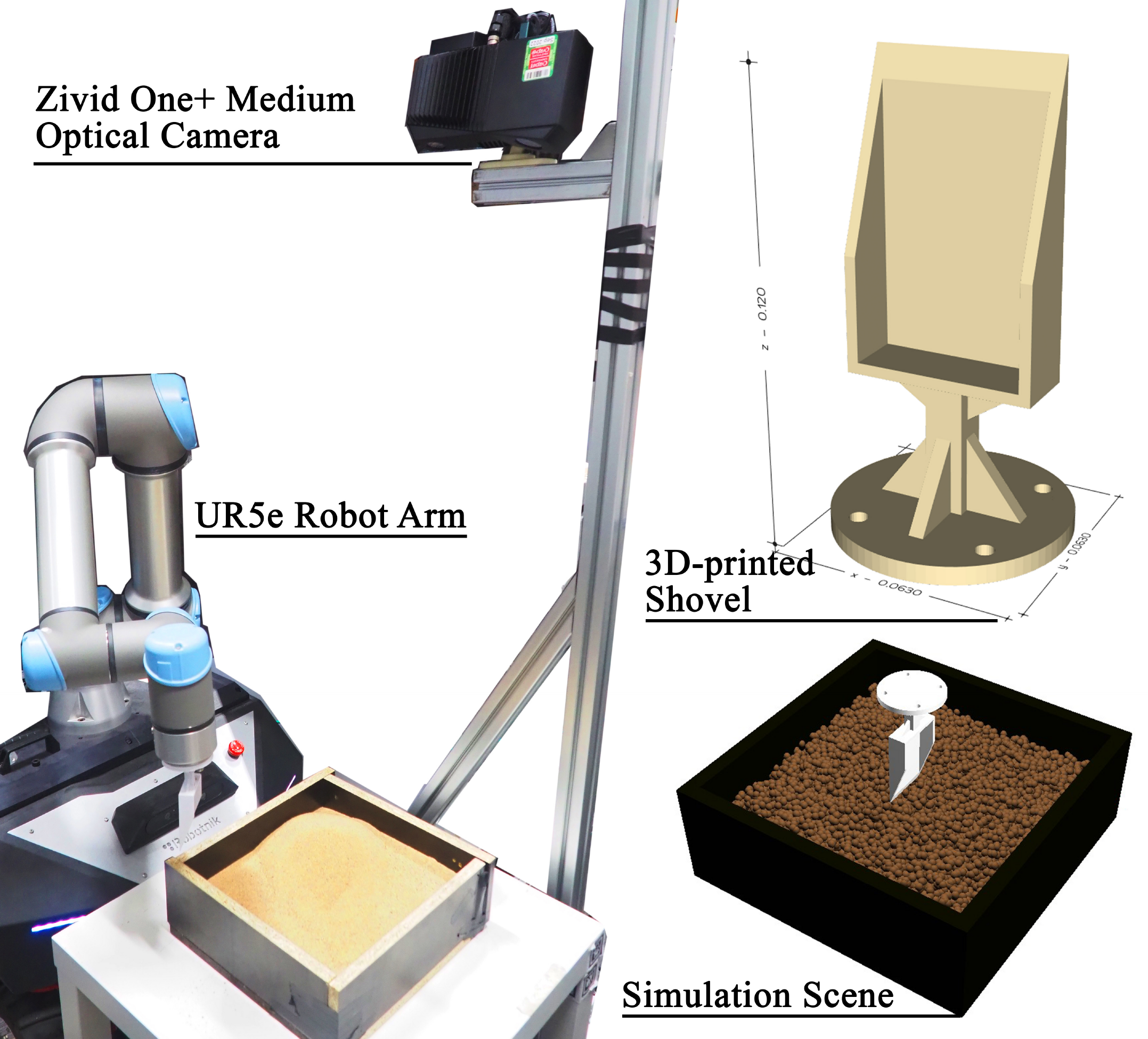}
\caption{Real manipulation platform}\label{fig:rplat}
\end{figure}

We set up a UR5e robotic arm to perform digging manipulations in the real world. As shown in Figure~\ref{fig:rplat}, the robot is equipped with a 3D-printed shovel as the end-effector. We place a $0.28\times0.28$ m$^2$ wooden box with $0.11$ metre depth to contain sand or soil for manipulation next to the robot. In the simulation, we use a box of the same size and the 3D model of the shovel as the simulated end-effector. We fill the container with the material, which could be sand or solid, to a height of $0.07$ metres. A Zivid Medium One+ high-resolution camera~\cite{zividone} is used to collect real-world surface point clouds.

\subsection{Soil/sand system identification}\label{subsec:sysid}
Before running skill learning and parameter optimisation with the simulation dynamics, an important task is to ensure that the simulated soil/sand dynamics match the real ones as closely as possible. To achieve this, we follow the Differentiable Physics-based System Identification (DPSI) framework introduced in work~\cite{xy2024sysid}, which can identify physics parameters with only one datapoint. In simple terms, the DPSI framework adjusts the key parameters of a physics model using gradients backpropagated from a loss function that compares the simulated and real point clouds of the same manipulation motion. 

In this work, we seek to identify Young's modulus, Poisson's ratio, material density, and the friction angle of sand/soil for a more realistic simulation. MOVEIT! is used in this work to compute two manipulation trajectories - one for optimisation and one for validation - for the system identification tasks. 

The optimisation trajectory moves the shovel from its initial configuration by $+0.09$ m in the $x$ direction, $-0.05$ m in the $z$ direction that inserts the shovel, $-0.12$ m in the $x$ direction that pushes the material and another $+0.12$ m in the $z$ direction that lifts the shovel. The validation trajectory moves the shovel to the corner of the container facing towards the diagonal corner, inserts it into the material for $-0.05$ m in the $z$ direction, pushes the material diagonally ($+0.12$ m in both $x$ and $y$ directions), and lifts the shovel for $+0.12$ m in the $z$ direction. For both soil and sand, the camera records the surface point clouds after the two trajectories as real-world ground truths. The velocity-based trajectories generated by MOVEIT! for system identification are transformed into a series of position-based waypoints with even timesteps for recreating the same end-effector motion in simulation.

The ranges of the physics parameters are defined according to an open-source structural engineering dataset~\cite{StructX}. They are summarised together with the gradient descent step sizes in Table~\ref{tab:lr}. The RMSprop algorithm with $\beta=0.9$~\cite{tieleman2012rmsprop} is used for gradient-based optimisation in this work.

\begin{table}[h]
\footnotesize
\centering
\begin{tabular}{l|cccc}
\toprule
Parameter     & $E$ (kPa) & $\nu$ & $\rho$ (kg/m$^3$) & $\phi_f$ (deg)\\
\midrule
Step size     & $10000$   & $0.01$ & $50$ & $1$ \\
Min value     & $50000$  & $0.1$ & $1200$ & $10$\\
Max value     & $200000$ & $0.4$  & $2200$ &$40$\\
\bottomrule
\end{tabular}
\caption{Stepsizes and the ranges of values for the physics parameters: Young's modulus $E$, Poisson's ratio $\nu$, material density $\rho$, and sand friction angle $\phi_f$, selected based on an open-source structural engineering dataset~\cite{StructX}.}
\label{tab:lr}
\end{table}

The feasibility of DPSI for physics parameter identification has been demonstrated with playdough-like plastic objects, which are governed by the fixed-corotated elastic energy model and the von Mises plasticity model~\cite{xy2024sysid}. In this paper, we cope with granular materials that are modelled by the St. Venant-Kirchoff (SVK) elastic energy and the Drucker-Prager (DP) plastic yield criterion~\cite{klar2016drucker}. 

\subsection{Digging tasks} 
\begin{figure}[t]
\centering
\includegraphics[width=\columnwidth]{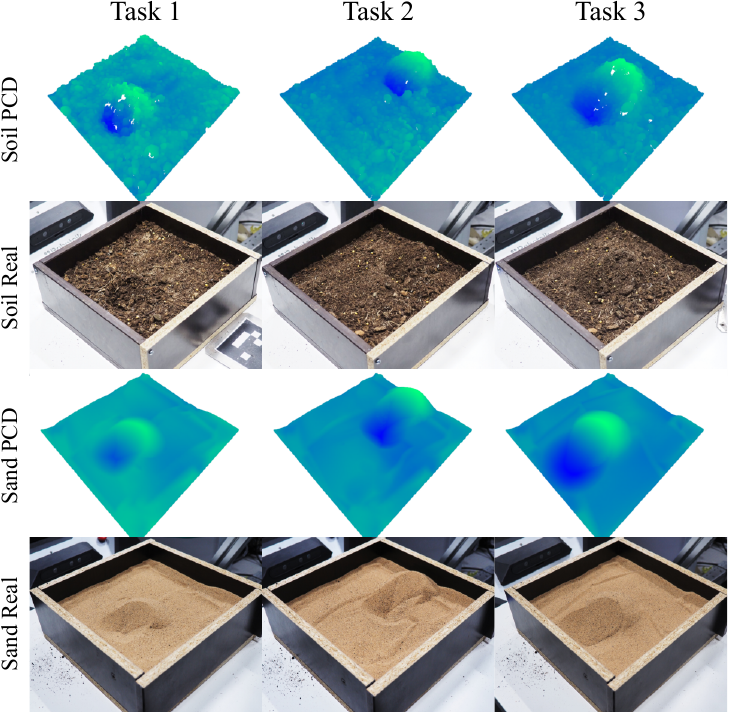}
\caption{Digging task goal visualisation. PCD: point cloud.}\label{fig:task-pcds}
\end{figure}

After the simulation is aligned with the real materials, we examine the feasibility of gradient-based skill optimisation. We manually use the shovel to create $3$ digging target shapes with both soil and sand. The optical camera then captures the resultant surface point clouds of the materials as task goals (see Figure~\ref{fig:task-pcds}). We compare our skill optimisation method to direct trajectory optimisation (controlling 6D Cartesian displacement at every global step), Covariance Matrix Adaptation MAP-Annealing (CMA-MAE)~\cite{fontaine2023cmamae, tjanaka2023pyribs} and goal-conditioned SAC (see Appendix C for implementation details). 

For both gradient-based skill and trajectory optimisation, we perform $20$ gradient updates. RMSprop is used for gradient-based optimisation with a stepsize of $0.03$ for skill parameters and $0.004$ for trajectory optimisation, respectively. 

\rp{CMA-MAE is a state-of-the-art evolutionary algorithm that does not use gradients to optimise a solution but optimises the problem by searching for a batch of high-quality and diverse solutions~\cite{fontaine2023cmamae, tjanaka2023pyribs}. For each task, it is run with two optimisation directions (i.e., two emitters) with a solution batch size of $10$ for $20$ iterations with a stepsize of $0.03$.}

The SAC agent is trained for $200$ episodes, which results in $200$ datapoints. Exploration is performed for $50$ episodes by sampling the parameters from a Gaussian exploration strategy with the demonstration skill parameters as the mean. The critic and policy are updated $10$ times with a batch size of $10$ at every episode with a learning rate of $0.001$. At every $5$ training episodes, the policy is evaluated for $5$ episodes to record its performance over training. 

\subsection{Experiment questions}
We present our experiment in four parts: 1) gradient and loss examination, 2) differentiable system identification, 3) differentiable skill optimisation and 4) benchmarking and ablation.

In section~\ref{sec:result-grad}, we take a closer look at the behaviours of the gradients and the ruggedness of the loss landscapes. In section~\ref{sec:result-sysid}, we examine the performance of system identification tasks on soil and sand with the proposed gradient regularisation and step size treatments. In section~\ref{sec:result-skill}, we examine the performance of the gradient-based skill optimisation methods \rp{with and without skill demonstrations} using the most performant gradient treatment found in the system identification experiments. \rp{In section~\ref{sec:ablation}, we benchmark the performances of the baselines and examine the gradient-based optimisation performance with and without rounding the number of timesteps in the differentiable digging skill.}

\section{RESULTS - VISUALISING GRADIENTS AND LOSSES}\label{sec:result-grad}

\textbf{Exploding gradients.} \rp{In Appendix D, we visualise how the gradients of the key MLS-MPM variables, physics parameters, and actions evolve over the simulation's backwards direction. It clearly shows the effectiveness of the three techniques in constraining the gradient scales of these variables, thus ensuring stable gradients for optimising the physics parameters and actions. To further examine how the processed gradients will affect the actual optimisation process, we experiment to compare their performances in the system identification task in section~\ref{sec:result-sysid}.} 

\begin{figure*}
    \centering
    \begin{subfigure}{\linewidth}
    \centering
    \includegraphics[width=0.9\linewidth]{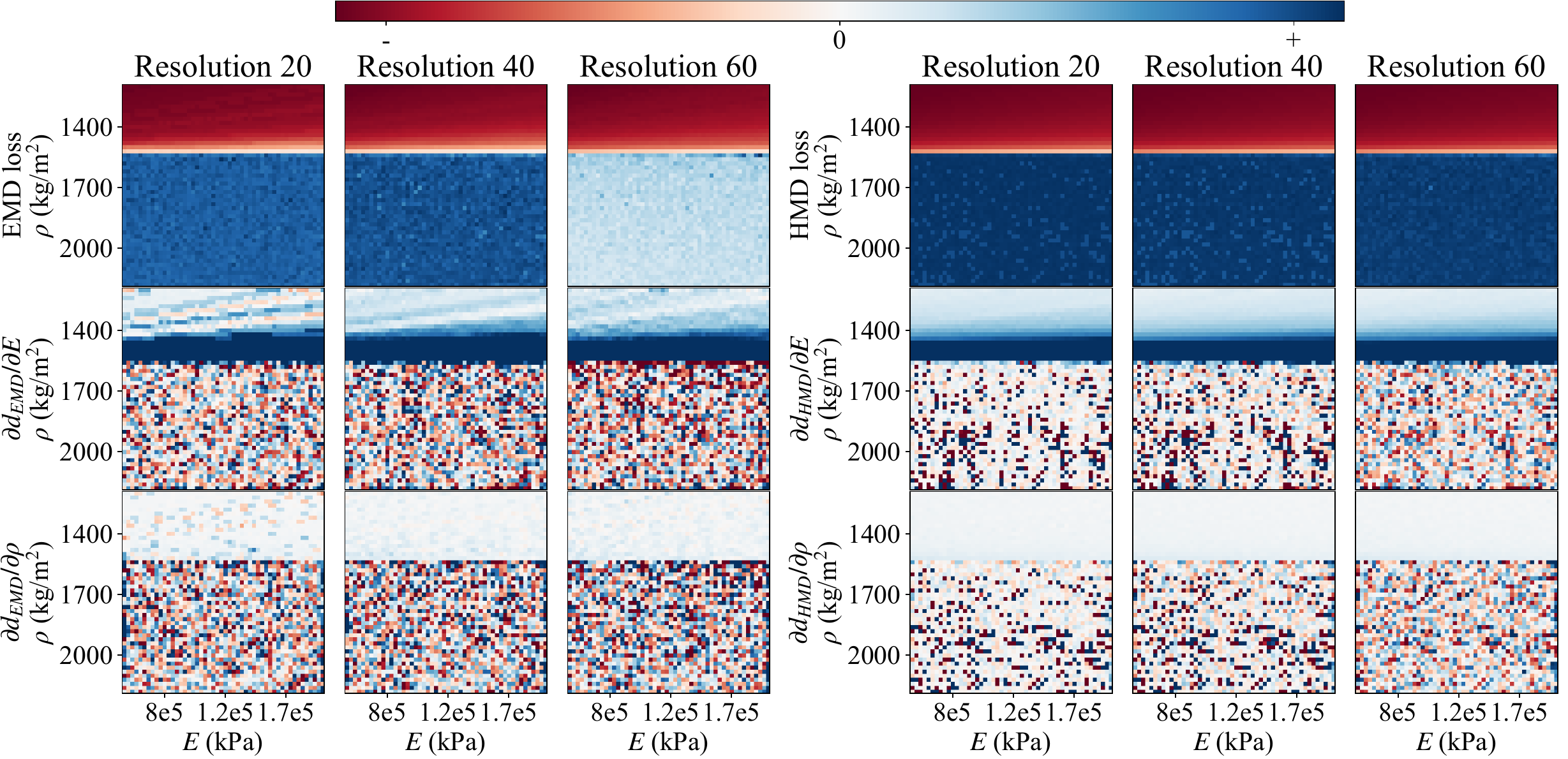}
    \caption{System identification task loss landscapes. On Young's Modulus and Material density.}\label{subfig:ER-loss}
    \end{subfigure}\\
    \begin{subfigure}{\linewidth}
    \centering
    \includegraphics[width=0.89\linewidth]{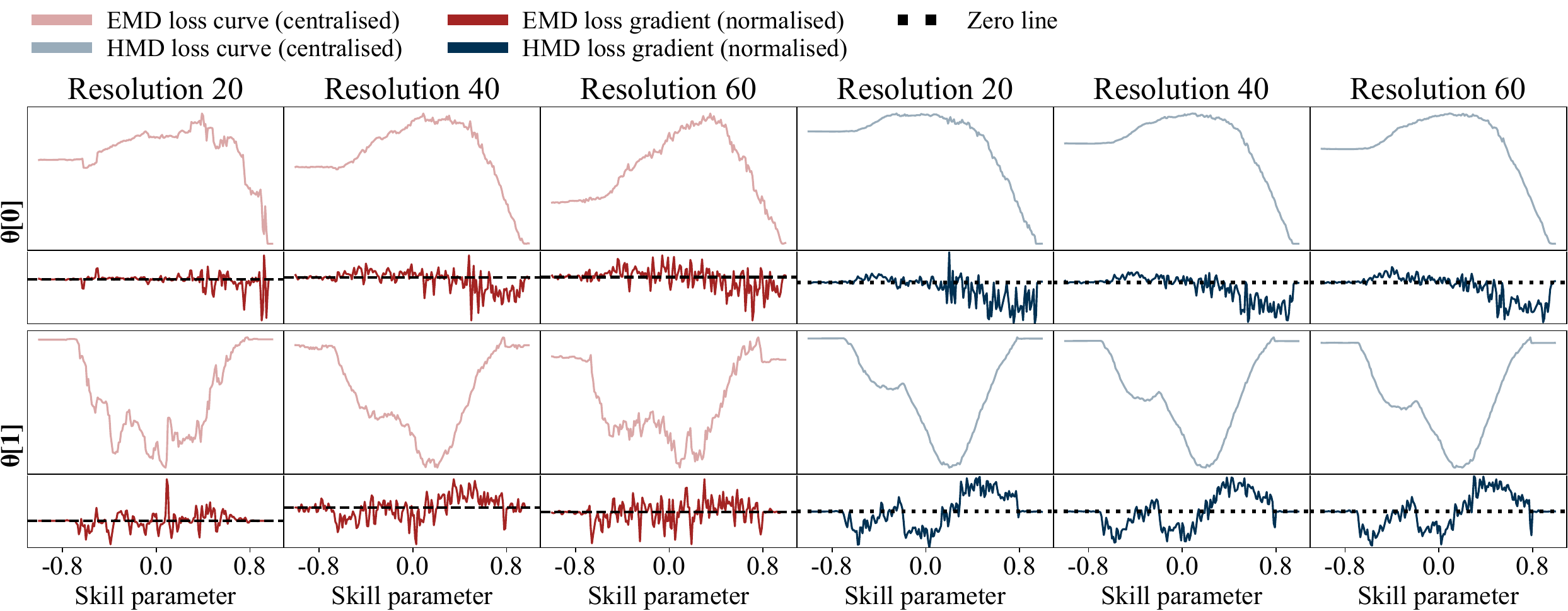}
    \caption{Skill optimisation task loss landscapes. On skill parameter $0$ and $1$.}\label{subfig:skill-loss}
    \end{subfigure}
    \caption{Loss landscapes and their gradient distributions.}
    \label{fig:loss-grad}
\end{figure*}

\textbf{Rugged loss landscapes/fluctuating gradients.} To understand the distributions of the EMD and HMD losses, we visualise the loss landscapes of both losses over the space of the physics parameters and the skill parameters with a grid search method. For visualisation, the values of two of the physics parameters are fixed to calculate the losses while varying the values of the other two parameters over their ranges with $50$ even steps. The system identification losses are calculated using the optimisation motion and data for soil material. For the skill parameters, the values of four of the parameters are fixed, based on the demonstration skill, and the losses are calculated, while the value of the singled-out parameter is varied over its range with $100$ even intervals. The task losses are calculated using the first soil-digging task. For all loss landscapes, their approximate gradients are also calculated using the second-order central differences at the inner points and the one-sided (forward or backward) difference at the boundaries. Note that, as we are interested in the ruggedness of the losses and the directions of their gradients, these losses are centralised, and the gradients are normalised to facilitate more intuitive analyses.

Figure~\ref{fig:loss-grad} visualises these losses and gradients over Young's Modulus and material density and the first two skill parameters ($\theta_{displace}$ and $\theta_{rotate}$). They are computed with the point sets (simulated particles or real-world surface point clouds) being sampled with resolutions $20$, $40$ and $60$, which correspond to $400$, $1600$ and $3600$ points. Due to the page limit, the losses and gradients over Poisson's ratio and sand friction angle and the last three skill parameters ($ \theta_{insert\_distance}$, $\theta_{push\_angle}$, and $\theta_{push\_distance}$) can be found in Appendix D. 

\begin{figure*}
    \centering
    \begin{subfigure}{\linewidth}
    \includegraphics[width=0.8545\linewidth]{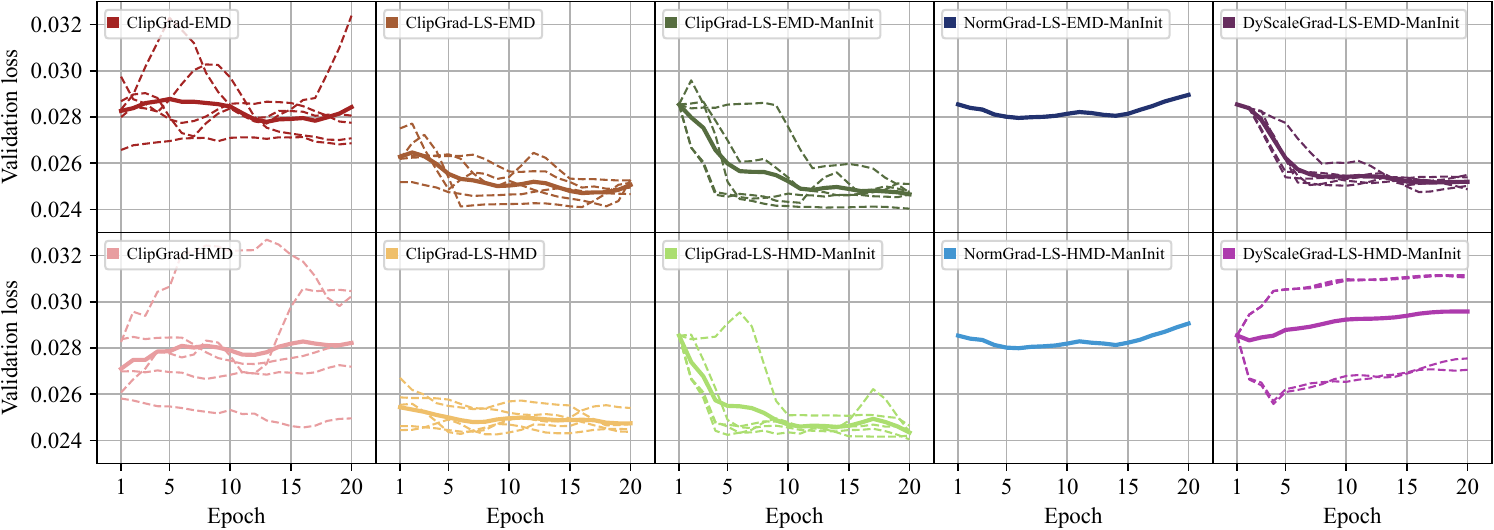}
    \includegraphics[width=0.139\linewidth]{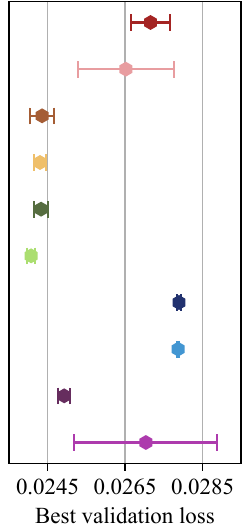}
    \caption{Soil}\label{subfig:sysid-soil}
    \end{subfigure}\\
    \begin{subfigure}{\linewidth}
    \includegraphics[width=0.8545\linewidth]{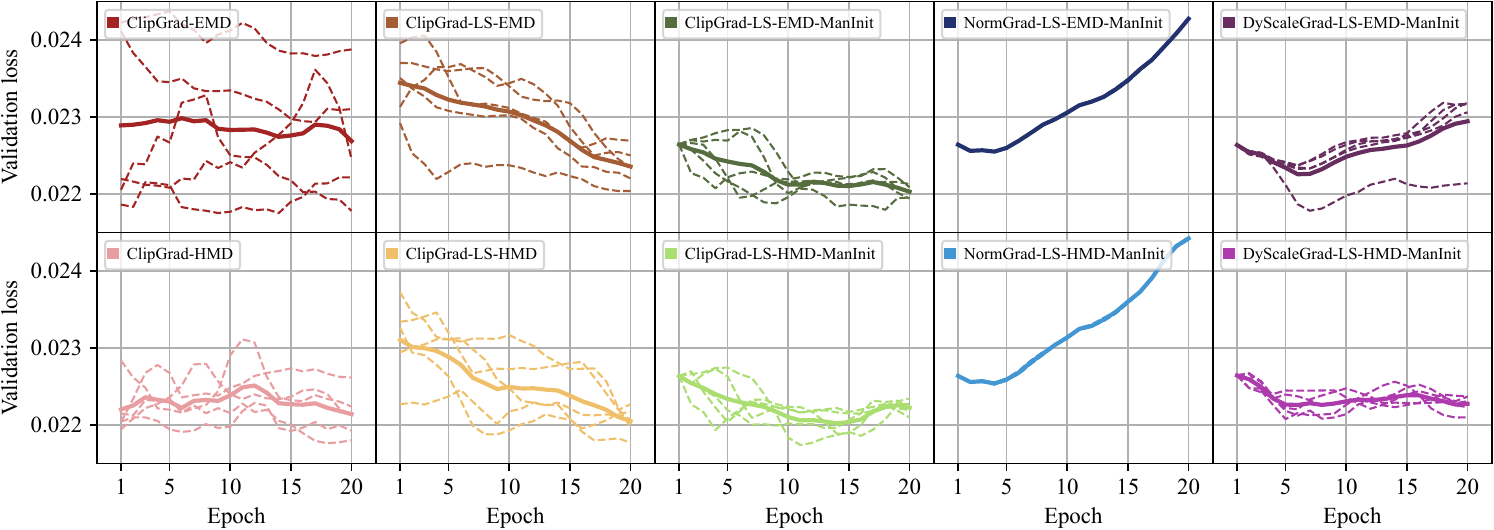}
    \includegraphics[width=0.139\linewidth]{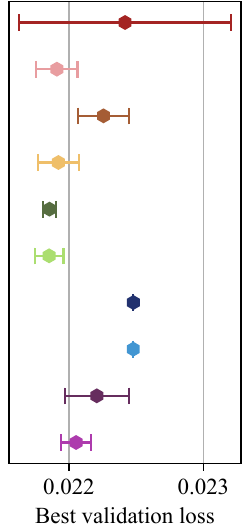}
    \caption{Sand}\label{subfig:sysid-sand}
    \end{subfigure}
    \caption{System identification performances with soil/sand. LS: line search. Validation loss: the sum of the EMD and HMD loss averaged over the sampling resolution ($40\times40$). EMD: earth mover's distance. HMD: height map distance. ManInit: manually selected initial solution. NormGrad: gradient normalisation. DyScaleGrad: gradient dynamics-scaling. The validation loss values are computed by summing the EMD and HMD losses averaged by their sampling resolution ($1600$ points/pixels). For each material, we present the loss curves over gradient updates and the mean/std of the best validation loss values found across five random seeds.}
    \label{fig:sysid}
\end{figure*}

\begin{figure*}
\centering
\includegraphics[width=0.9\linewidth]{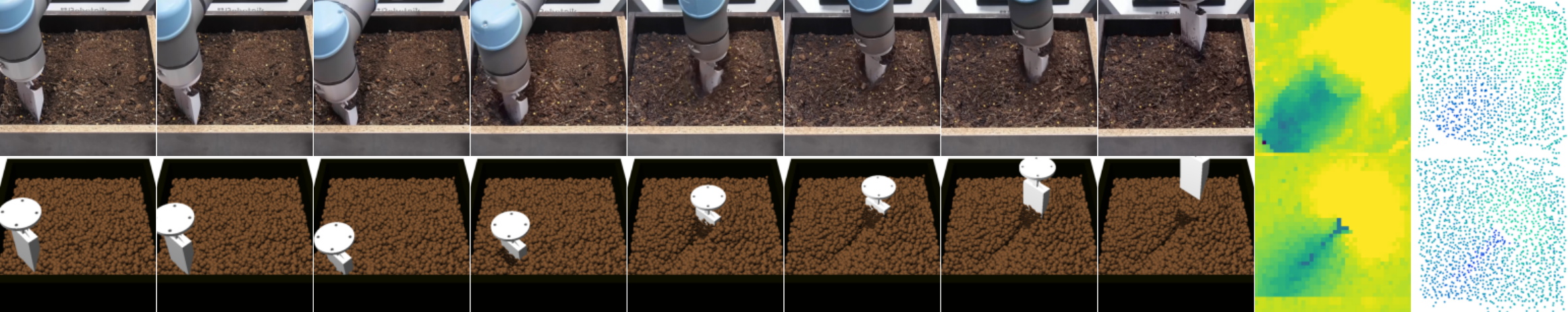}\\
\includegraphics[width=0.9\linewidth]{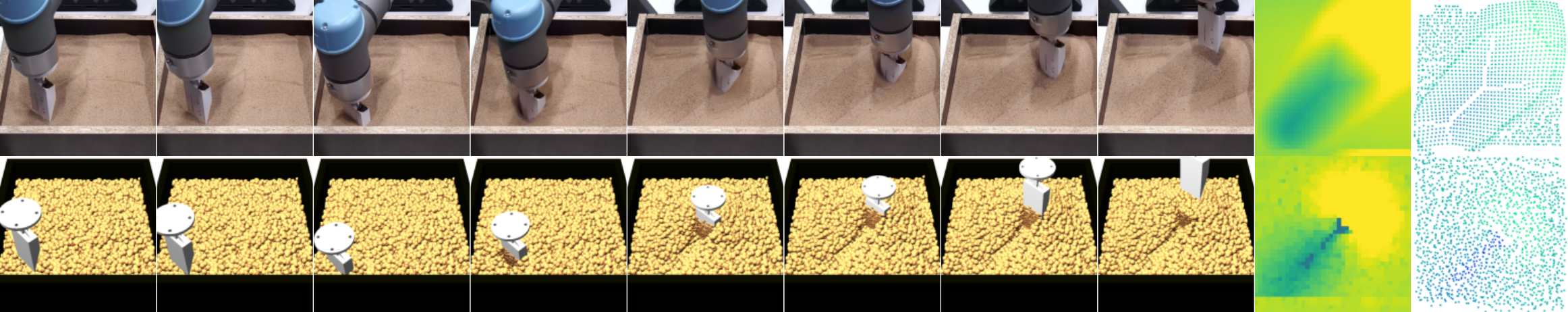}\\
\includegraphics[width=0.9\linewidth]{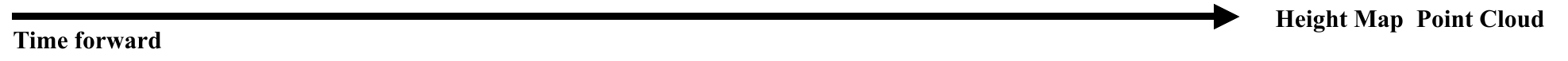}
\caption{Visualising the real and simulated system identification validation trajectories with the best parameters found (``ClipGrad-LS-HM-ManInit'' in Table~\ref{tab:sys-id}).}\label{fig:sisid-tr}
\end{figure*}

First of all, by comparing the loss distributions between EMD and HMD losses, one can see that, although they exhibit similar distributional tendencies, the HMD losses tend to be smoother. This is shown in both tasks. In Figure~\ref{subfig:ER-loss}, within either the negative (red) or positive (blue) parts of the losses (first row in Figure~\ref{subfig:ER-loss}), the HMD loss produces more gradients that are closer to zeros (the second and third rows of the right column in Figure~\ref{subfig:ER-loss}). In Figure~\ref{subfig:skill-loss}, it is obvious that the light blue curves are smoother than the light red curves. This suggests that when approaching local minima, the HMD loss is more likely to settle down with its gradients approaching zeros. Secondly, the sampling resolution seems to have a mild influence on the level of gradient noisiness. Specifically, Figure~\ref{subfig:ER-loss} (the second and third rows on the right) shows that the HMD loss seems to be more rugged with increased resolution in system identification tasks, and Figure~\ref{subfig:skill-loss} (red curves) shows that the EMD loss seems to be more rugged with resolution $20$ and $60$. 

In general, with the gradients fluctuating frequently around zero when approaching local minima, gradient-descent algorithms will have difficulty converging to a solution. In addition, as the different parameters will have different gradient magnitudes, a one-size-fits-all stepsize (or stepsize schedule) is unlikely to be found. These observations suggest the need for a more adaptive stepsize selection method, such as the line search method described in subsection~\ref{subsec:grad-fluctuate}. 

The next section compares the performances of different gradient operations and the line search method in the system identification task for both soil and sand materials. As the sampling resolution has a mild effect on the level of loss ruggedness, we compute the losses in the rest of the experiments with a resolution of $40$ ($1600$ points or $40\times40$ height maps).

\section{RESULTS - SYSTEM IDENTIFICATION}\label{sec:result-sysid}

This section presents the system identification task performance with differentiable physics and the ablation study that helps to determine the effectiveness of the EMD and HMD losses, the gradient operations (clipping, dynamic-scaling and normalisation), and the line search technique (five random seeds). To recall, we follow ~\cite{xy2024sysid} to conduct the DPSI task by optimising four physics parameters (Young's modulus, Poisson's ratio, material density, and sand friction angle) with gradients from either the EMD or HMD loss between the simulated and real manipulated material deformed by the same end-effector motion. In this work, we perform DPSI with soil and sand. As performing optimisation with a good initial solution accelerates the convergence of DPSI suggested by~\cite{xy2024sysid}, we also present the performance of starting from initial parameter values based on~\cite{StructX} (denoted by ``-ManInit'').

The results are presented in Figure~\ref{fig:sysid}. First of all, it can be seen that the optimisation diverges with both materials and loss functions (the red and pink lines) while the use of line search greatly stabilises gradient descent and converges to solutions with much reduced variances. This evidence demonstrates that our hypothesis of the optimisation being affected by a rugged loss landscape is accurate and the line search method is effective in addressing this problem. Secondly, with a good initial solution (dark and light green curves), DPSI can converge faster to better solutions with lower variances in both materials, aligned with~\cite{xy2024sysid}. Thirdly, the normalisation (dark and light blue curves) and dynamic-scaling (dark and light purple curves) operations perform worse than simply clipping the gradients (dark and light green curves). Lastly, the HMD loss performs slightly better than the EMD loss as all the light-coloured curves/scatter-points show slightly improved values than the dark-coloured ones (specifically, lower validation losses and variances).

In short, from the system identification results, we can be confident that the combination of HMD loss, gradient clipping, line search, and a good initial guess can more stably and efficiently lead to good optimisation results. 

\begin{table}[h]
\centering
\begin{tabular}{lccccc}
\toprule
Material             & E (kPa) & $\nu$ & $\rho$ (kg/m$^3$) & $\phi$ ($^\circ$)\\
\midrule
Soil                 & 182683 & 0.242 & 1566 & 18.882\\
\midrule
Sand                 & 121378 & 0.198 & 1974 & 19.019\\
\bottomrule
\end{tabular}
\caption{Physics parameters associated with the best validation loss optimised using gradient clipping (ClipGrad), line search (LS), Height map distance (HMD) loss and manually initialised solution (ManInit)\label{tab:sys-id}}
\end{table}

For both soil and sand, Table~\ref{tab:sys-id} presents the physics parameter values associated with the best validation loss across five random seeds (``ClipGrad-LS-HMD-ManInit'', light green lines in Figure~\ref{fig:sysid}). The validation trajectories for soil and sand simulated with these parameters are presented in Figure~\ref{fig:sisid-tr}. These parameters are used in the following skill or trajectory optimisations and the RL training environment.

\section{RESULTS - DIGGING TASKS}\label{sec:result-skill}
\begin{figure}[t]
\centering
    \begin{subfigure}{\columnwidth}
        \includegraphics[width=\columnwidth]{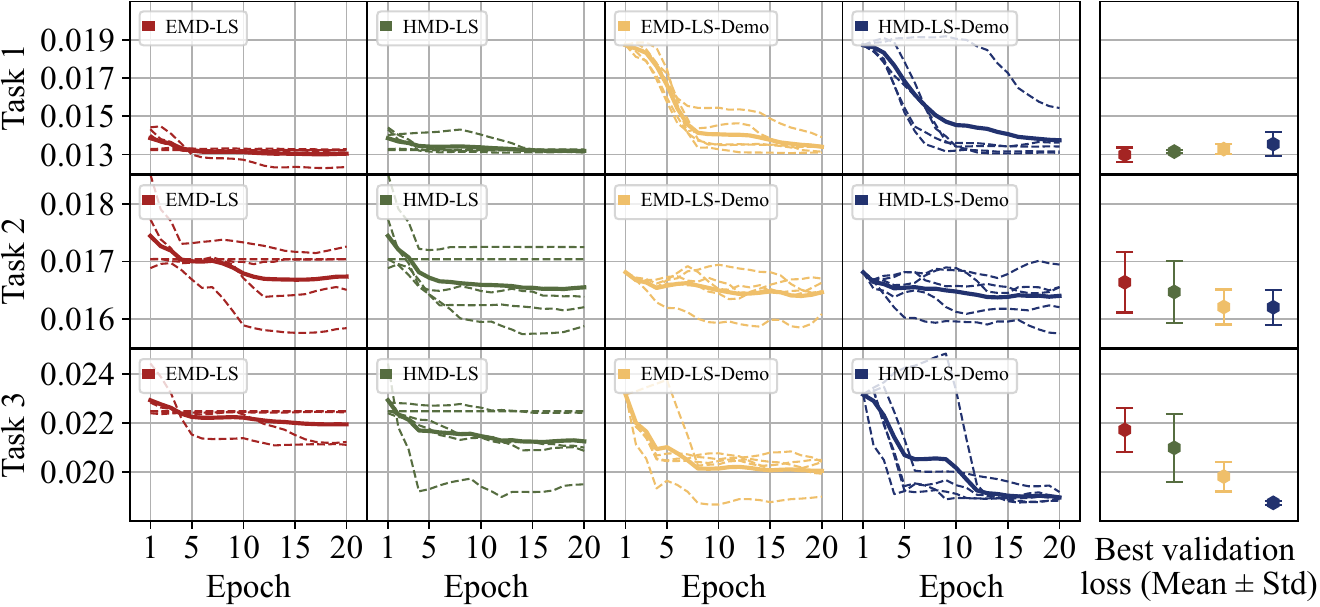}
    \caption{Soil}\label{subfig:task-soil}
    \end{subfigure}\\
    \begin{subfigure}{\columnwidth}
        \includegraphics[width=\columnwidth]{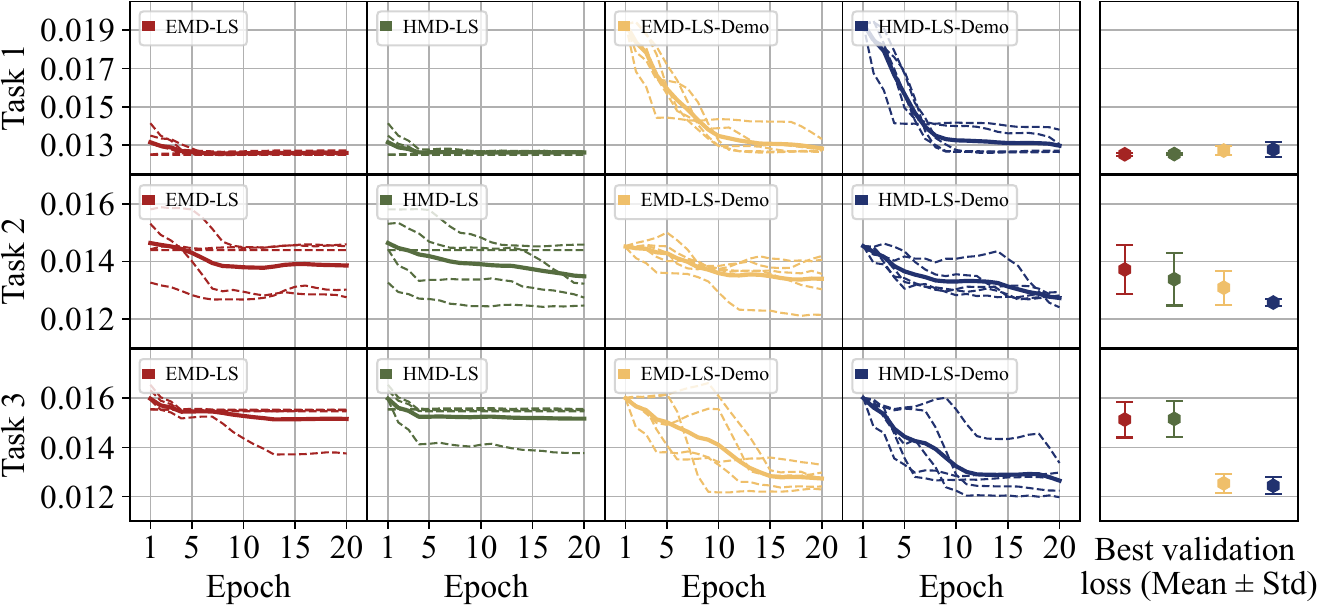}
    \caption{Sand}\label{subfig:task-sand}
    \end{subfigure}\\
\caption{Validation losses of the soil and sand digging tasks for the proposed gradient-based skill optimisation method. Validation loss: the sum of the EMD and HMD loss averaged over the sampling resolution ($40\times40$). SO: gradient-based skill optimisation. EMD: Earth mover's distance. HMD: height map distance. LS: line search. Demo: demonstration-based skill prior.}\label{fig:so-task-curve}
% TR: gradient-based trajectory optimisation. SAC: soft actor-critic. HER: hindsight experience replay.
\end{figure}

\begin{figure*}
\centering
\includegraphics[width=\linewidth]{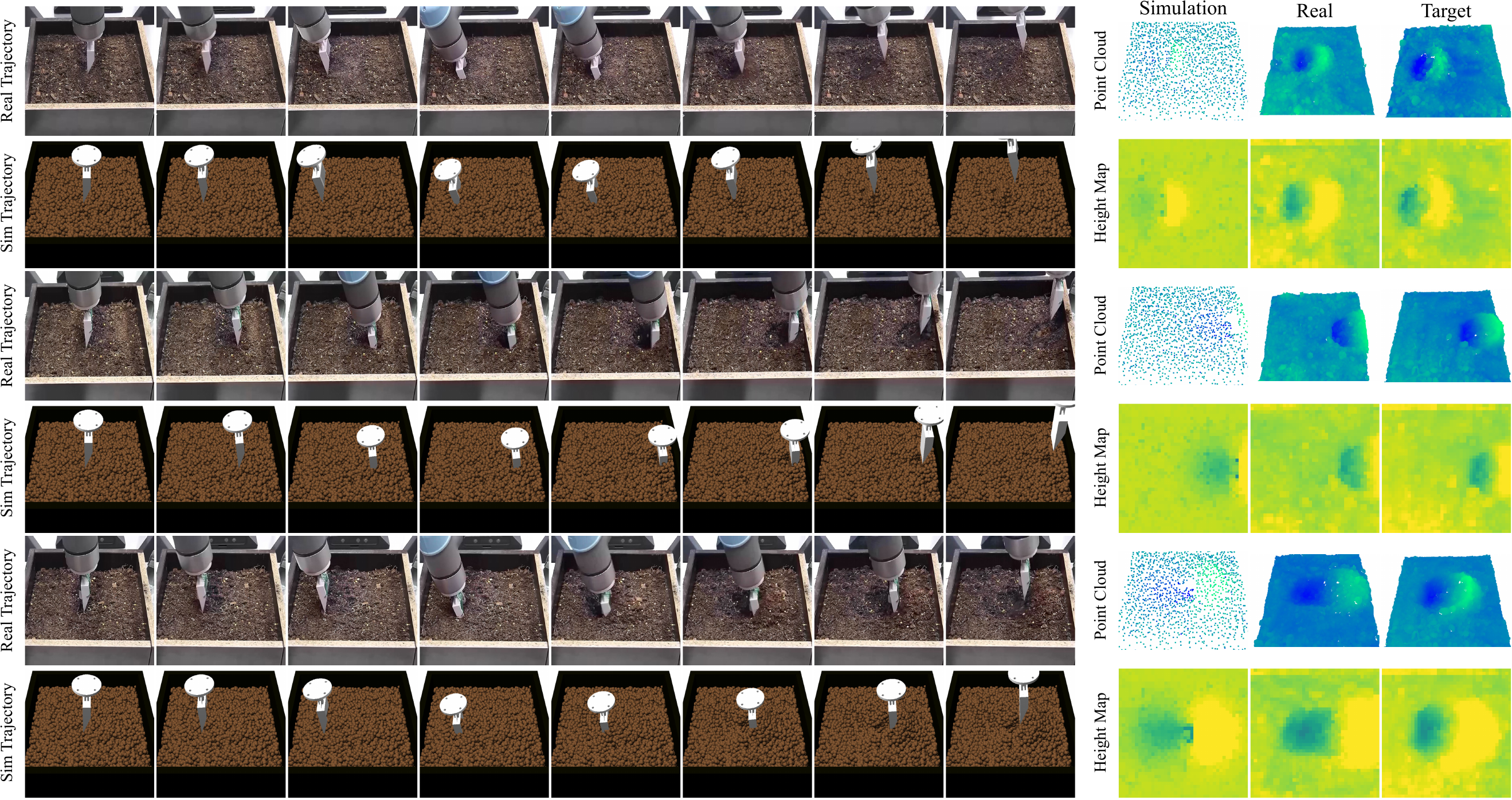}\\ \vspace{0.5cm}
\includegraphics[width=\linewidth]{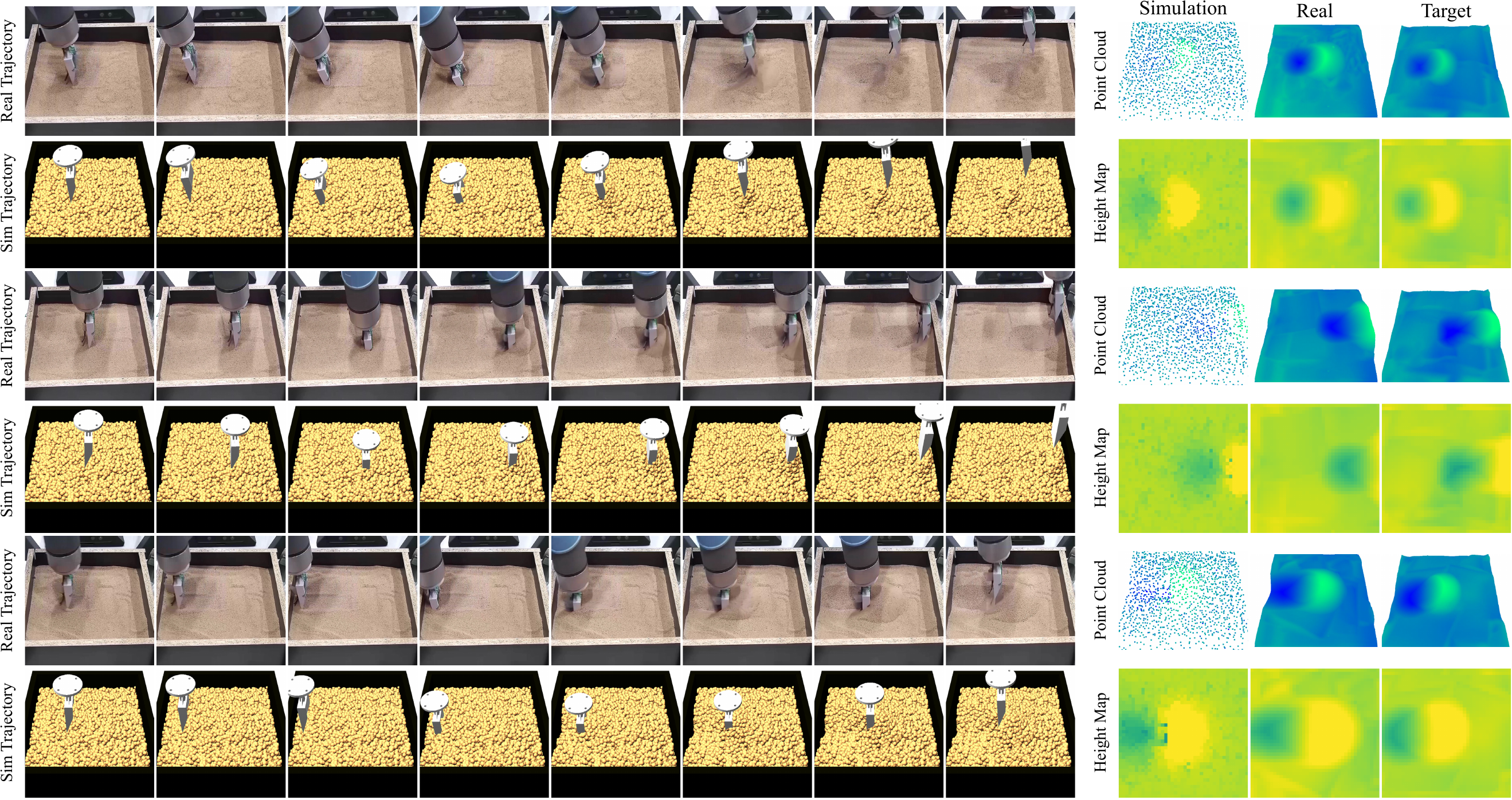}
\caption{Visualisation of the simulated and real trajectories associated with the best validation losses for soil and sand digging tasks found by the proposed gradient-based skill optimisation method.}\label{fig:task-tr}
\end{figure*}

This section examines the performance of the $6$ digging tasks (Figure~\ref{fig:task-pcds}) achieved by the proposed gradient-based skill optimisation method. In Figure~\ref{fig:so-task-curve}, we plot their validation losses. The best-performing skills found by the gradient-based skill optimisation method are deployed on the real robot. Their simulation and real trajectories, resultant point clouds and height maps, are visualised in Figure~\ref{fig:task-tr}.

\begin{table}[t]
\centering
\begin{tabular}{ABCDDD}
\toprule
\textbf{Features} (cm)          &                       &         & \textbf{Task 1} & \textbf{Task 2} & \textbf{Task 3} \\
\midrule
                                & \multirow{2}{*}{Soil} & Targets & (6.79, -0.08) & (-6.23, 0.03) & (3.12, -0.97)\\
\textbf{Center}                 &                       & Results & (5.61, -0.52) & (-6.73, -0.65)& (3.27, -0.64)\\
\cline{2-6}
\textbf{(x, y)}                 & \multirow{2}{*}{Sand} & Targets & (6.60, -0.15) & (-2.58, 0.17) & (8.57, 0.18)\\
                                &                       & Results & (5.92, -0.63) & (-4.11, -0.51) & (8.85, -0.86)\\
\midrule
\multirow{4}{*}{\textbf{Depth}} & \multirow{2}{*}{Soil} & Targets & 1.74   & 1.48   & 2.43 \\
                                &                       & Results & 1.69   & 1.48   & 2.78 \\ 
\cline{2-6}
                                & \multirow{2}{*}{Sand} & Targets & 0.74   & 1.75   & 1.48\\
                                &                       & Results & 1.18   & 1.38   & 1.76\\
\midrule
\multirow{4}{*}{\textbf{Area}}  & \multirow{2}{*}{Soil} & Targets & 24.84  & 27.72  & 46.80 \\
                                &                       & Results & 24.12  & 32.04  & 78.48\\ 
\cline{2-6}
                                & \multirow{2}{*}{Sand} & Targets & 17.28  & 52.20  & 50.40 \\
                                &                       & Results & 30.96  & 52.20  & 46.08\\
\bottomrule
\end{tabular}
\caption{\rp{Statistics of the real-world dug holes (cm). The averaged differences of hole centre, depth and area across the six tasks are (0.60, 0.59) cm, 0.25 cm and 10.8 cm$^2$.}}
\label{tab:holes}
\end{table}

From Figure~\ref{fig:so-task-curve}, it can be seen that the \rp{proposed gradient-based skill optimisation method can successfully reduce the loss values and converge to local minima. However, the} use of demonstration (skill prior) significantly accelerates optimisation and leads to better performances, especially in tasks $2$ and $3$ for both soil and sand materials (see the yellow and blue curves in Figures~\ref{subfig:task-soil} and~\ref{subfig:task-sand}). For the first soil and sand digging tasks, the optimisation without demonstrations performs better. This indicates that the demonstrated skill is close to a good solution, but not in every case (in the first rows of Figures~\ref{subfig:task-soil} and~\ref{subfig:task-sand}, the yellow and blue validation losses at the early epochs are significantly higher despite the demonstrated skill).

From the visualisation of the simulation and real digging trajectories with the best skill parameters in Figure~\ref{fig:task-tr}, it can be seen that the similarity between the simulated and real dynamics is significant. The solutions can dig holes highly similar to the target shapes that were manually created. \rp{To statistically evaluate the performance, we calculate the centres, depths and areas of the real-world dug holes using their height maps. Table~\ref{tab:holes} compares these statistics of the task targets and the holes dug by the best skill found by our method. It was found that, over the six tasks, the average difference in hole centres is no more than $6$ mm, the average difference in hole depth is $2.5$ mm, and the average area difference is $10.8$ cm$^2$.} This proves the feasibility of deploying the optimised skills directly on a real robot.

\section{RESULTS - BASELINE and ABLATION}\label{sec:ablation}
\rp{This section presents the baseline performances and ablation study on the soil digging tasks with demonstrations.}

\subsection{Baselines} 
\begin{figure}[h]
\centering
    \begin{subfigure}{\columnwidth}
        \includegraphics[width=\columnwidth]{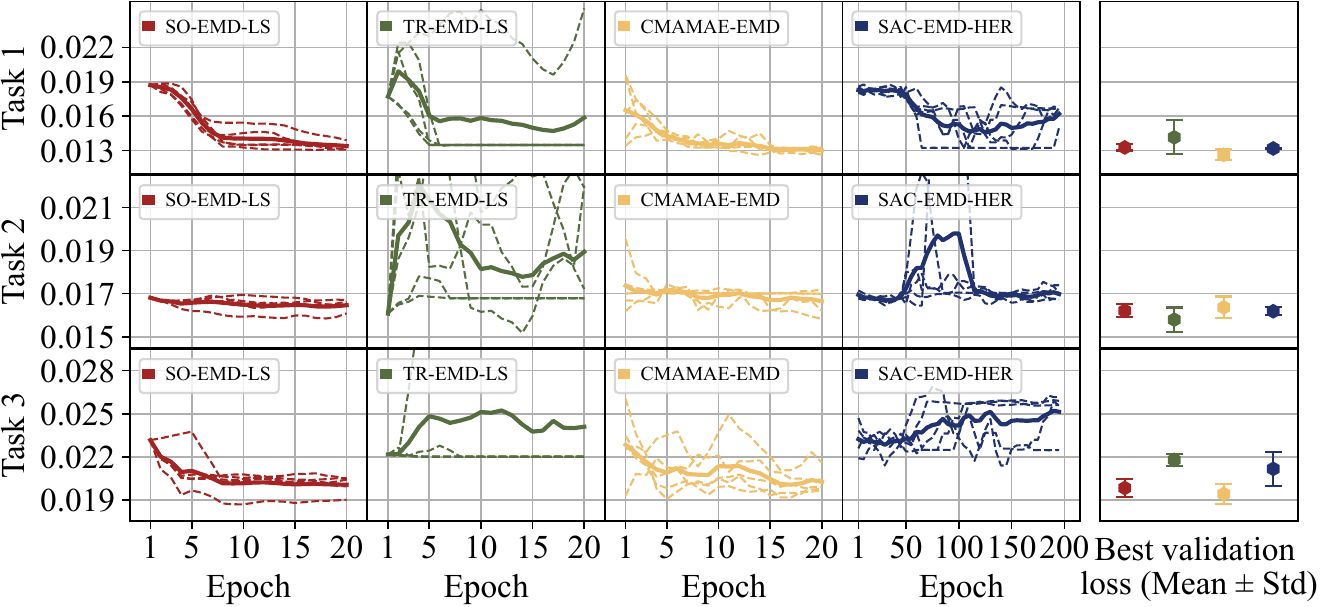}
    \caption{Performances achieved by different methods on soil digging tasks with demonstrations. SO: gradient-based skill optimisation. LS: line search. TR: gradient-based trajectory optimisation. CMAMAE: Covariance Matrix Adaptation MAP-Annealing. SAC: soft actor-critic. HER: hindsight experience replay.}\label{subfig:task-baseline}
    \end{subfigure}\\
    \begin{subfigure}{\columnwidth}
        \includegraphics[width=\columnwidth]{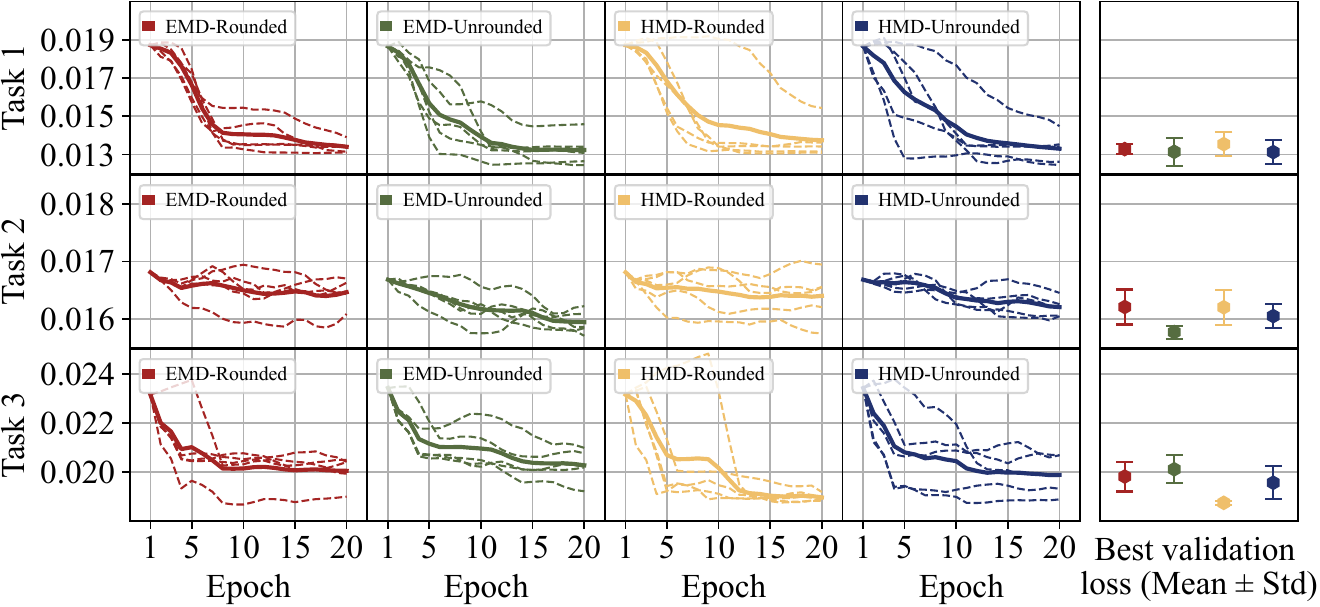}
    \caption{Gradient-based skill optimisation (with line search) ablation on using the unrounded number of timesteps in the skill-to-action mapping.}\label{subfig:task-rg}
    \end{subfigure}
\caption{Validation losses for baselines and ablation. Validation loss: the sum of the EMD and HMD loss averaged over the sampling resolution ($40\times40$). EMD: Earth mover's distance. HMD: height map distance.}\label{fig:task-bl-ab}
\end{figure}

From Figure~\ref{subfig:task-baseline}, it can be seen that across three \rp{soil digging} tasks \rp{given demonstrations}, gradient-based skill optimisation (red) \rp{and CMA-MAE (yellow)} consistently outperform direct trajectory optimisation (green) and SAC (blue). The only occasion that trajectory optimisation achieves lower validation loss than others is with a highly drastic and irregular trajectory, which happens to cause a lower loss value (see supplementary video). On the other hand, although the validation losses of the SAC agent show that it converges to a certain policy in all tasks, these policies are all stuck in local minima and found to produce very similar manipulation motions without accomplishing the tasks. The training statistics of SAC are given in Appendix E. 

In terms of computation cost, running gradient updates for either skills or trajectories directly for $20$ iterations takes about $18$ to $20$ minutes. In some cases, the algorithm converges to a good solution within $5$ iterations with a good initial guess, which takes only about $5$ minutes. Similarly, CMA-MAE also takes about $20$ minutes to perform $20$ iterations. The SAC agent also takes about $20$ minutes to finish training for $200$ episodes ($200$ datapoints and $19,000$ neural network updates), yet hardly escapes from local poor minima. 

\rp{The benchmark result provides a few insights to guide research and engineering practice. First, optimising in the skill parameter space is significantly more efficient than in the trajectory space, even with demonstrations. Secondly, when an efficient and physically realistic simulator is available, learning a neural network policy with reinforcement learning methods was found to perform worse compared to gradient descent or MPC algorithms. This is somewhat unsurprising because a DRL method is essentially optimising a significantly larger parameter space, in contrast to the skill parameters of size five. It suggests that physically realistic simulators, preferably differentiable, are of higher potential and importance to efficient and high-quality robotic manipulation. Lastly, our gradient descent method and the model predictive control (CMA-MAE) method perform comparably well in terms of both solution quality and computational cost in our task setting. This is aligned with comparisons in other domains, such as image generation or classic RL tasks~\cite{zhao2025cmamae-exp}. In practice, the hyperparameters of evolutionary algorithms, such as CMA-MAE, are generally more difficult to tune. The evolving direction of its stochastic search depends fully on the cost function, lacking consideration of the connection from the solution through the system dynamics to the post-manipulation states. This could lead to physically unreasonable solutions in complex tasks.}

\rp{\subsection{Unrounded number of timesteps} 
Figure~\ref{subfig:task-rg} compares the optimisation performances with and without using an unrounded number of timesteps in the phases 1 and 4 of the skill-to-action mapping (details in Section~\ref{subsec:skill}). The results show that the two cases perform similarly. In tasks 1 and 2, using an unrounded number of timesteps produces slightly better results, while in task 3, slightly worse. This result suggests that the use of an unrounded number of timesteps (a more complete gradient chain) could slightly improve the optimisation performances, but not always.}

\section{CONCLUSIONS}\label{sec:conclude}
\rp{Granular material manipulation is of high importance to many real-world scenarios, but the field significantly lacks research and engineering experiences, particularly in small-scale and high-precision manipulation tasks with unknown material properties.} This work thereby studied the challenging real-world granular material manipulation problem with unknown material properties in the context of \textit{small-scale and high-precision} digging. \rp{A differentiable digging robot (DDBot) system was developed for solving such manipulation tasks using a differentiable physics-based %(DP) 
simulator and a first-order gradient-based optimisation method (RMSProp). The differentiable simulator was tailored for granular material manipulation for fast data collection, trajectory evaluation and gradient computation, powered by GPU-accelerated parallel computing and automatic differentiation. A differentiable digging skill was designed to reduce the solution space and improve manipulation motion quality. The issues of exploding and fluctuating gradients were studied and resolved through gradient clipping and line search, enabling gradient-based optimisation. To speed up convergence, a digging target-oriented demonstration method is created to serve as an initial solution or form an explorative data distribution.}

\rp{The quantitative and visualisation results of extensive experiments confirmed that DDBot achieves high-quality system identification and skill optimisation performances with sand and soil materials. The optimised solutions were successfully deployed on a UR5e robotic arm in the real world without any adaptation (zero-shot sim-to-real transfer), achieving high digging precision (location and depth errors $<6$ mm, area error $\sim10.8$ cm$^2$). In addition, with GPU acceleration and the proposed gradient treatments, DDBot achieved nearly practically applicable optimisation efficiency (converges in $5$ minutes with a good initial solution).} 

\rp{The performance of DDBot was benchmarked against gradient-based trajectory optimisation, deep reinforcement learning (PointNet-backboned goal-conditioned soft actor-critic) and model predictive control (covariance matrix adaptation map-annealing). Results showed that trajectory optimisation and reinforcement learning (RL) failed in all tasks, while our method and model predictive control succeeded with comparable performances and efficiency. It was suggested that 1) optimisation in the trajectory space for long-horizon tasks should be avoided whenever possible, and 2) when an efficient and physically realistic simulator is available, DRL methods that learn a neural network are counterproductive, gradient-free methods like CMA-MAE lack considerations of the system dynamics, whereas first-order gradient descent methods are preferred for optimisation in the skill parameter space leveraging the gradients computed through the system dynamics.}

Currently, the proposed system is limited to a fixed task initiation (soil/sand always flat) and optimising only one skill. This is due to the difficulty of reconstructing 3D scenes involving granular materials with high precision. In the future, integrating an efficient 3D reconstruction module into the DDBot system will facilitate the optimisation of multiple skills. In addition, optimising the order of multiple skills is a discrete optimisation problem. \rp{The gradient of a discrete decision-making process requires careful treatment to enable differentiability (e.g., the Softmax operation). Computing the gradient chain will become combinatorially slow as it branches for multiple skills at every skill selection stage.} Studying the feasibility of differentiable simultaneous skill selection and optimisation \rp{is parallel to many other research topics such as task and motion planning and hierarchical reinforcement learning. A considerable amount of research and engineering effort is needed to develop new algorithms for such tasks in the future.}

\section*{ACKNOWLEDGMENT}
This work was supported by the UK Engineering and Physical Sciences Research Council (EPSRC) under grant No. EP/X018962/1. Minglun Wei was supported through an EPSRC Doctoral Training Partnership (No. EP/W524682/1).

\printbibliography

\begin{IEEEbiography}[{\includegraphics[width=1in,height=1.25in,clip,keepaspectratio]{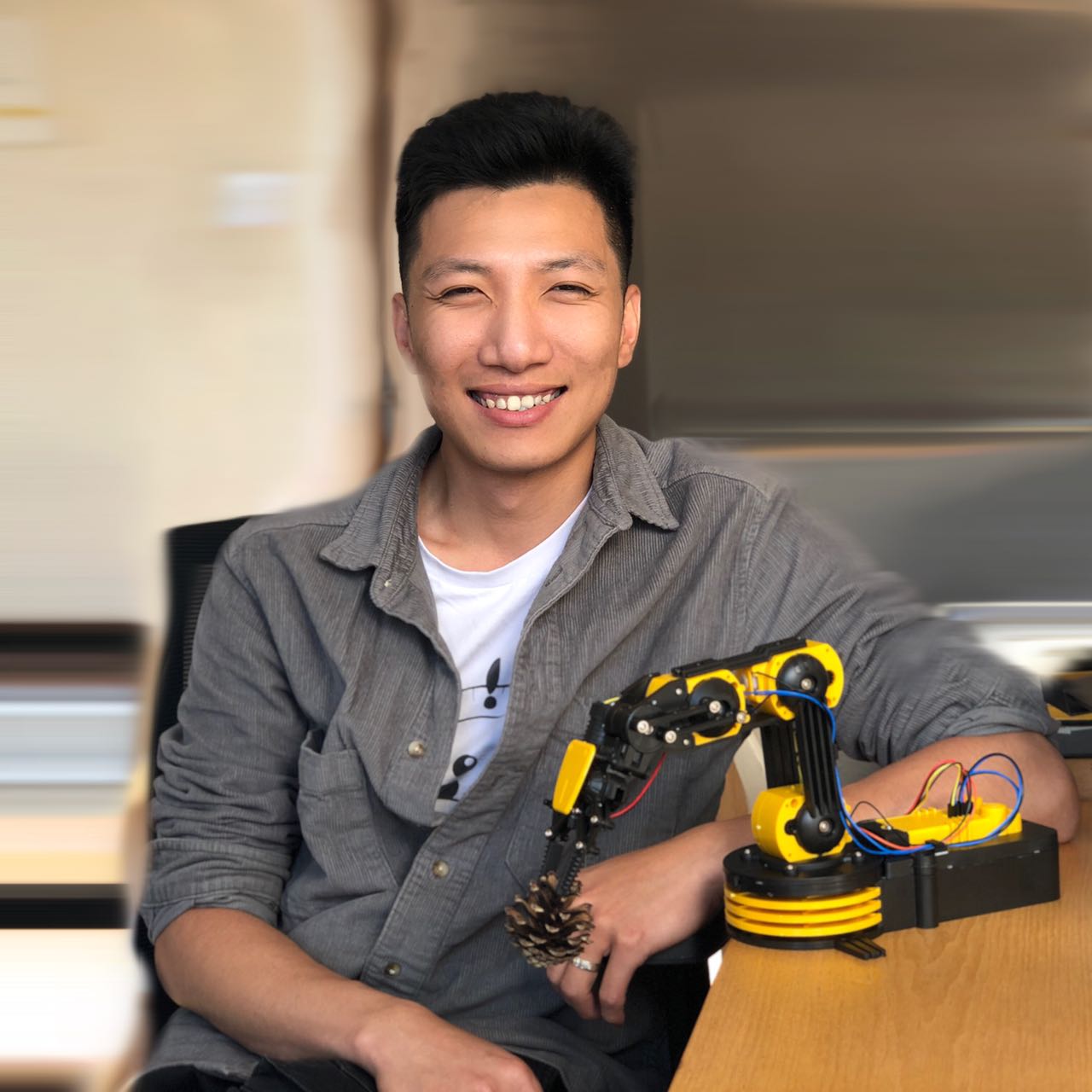}}]{Xintong Yang}
received his PhD from Cardiff University, Cardiff, U.K., in 2023, and his Bachelor’s and Master’s degrees in Mechanical and Industrial Engineering from Guangdong University of Technology, Guangzhou, China, in 2016 and 2019. He has been a research associate (postdoc) in the School of Engineering at Cardiff University since January 2023. He specialised in the robotic manipulation of real-world objects, rigid or deformable, through model-based and/or data-driven methods. Currently, he is primarily responsible for developing a robotic platform for automatically conducting biology/chemical experiments for AI-driven science discovery. He is also working on developing a real-world-applicable robotic manipulation system.
\end{IEEEbiography}

\begin{IEEEbiography}[{\includegraphics[width=1in,height=1.25in,clip,keepaspectratio]{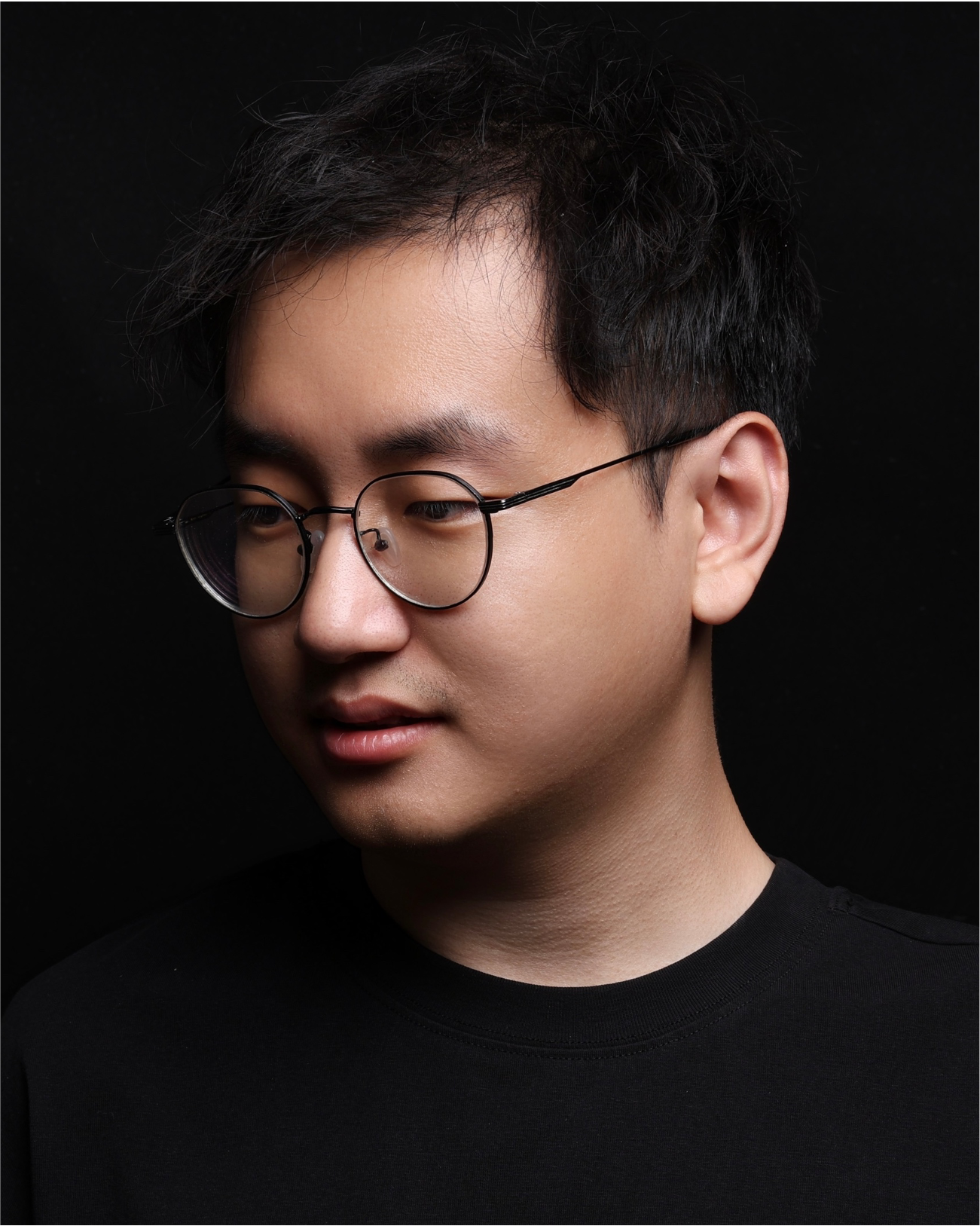}}]{Minglun Wei}
(Student Member, IEEE) received the B.Eng. degree in Microelectronics from Northwestern Polytechnical University, Xi’an, China, in 2020, and the M.Sc. degree in Signal Processing and Communications from The University of Edinburgh, U.K., in 2021. He is currently pursuing the Ph.D. degree with Cardiff University, Cardiff, U.K.Prior to starting his Ph.D. studies, he worked in industry on projects applying large language models to AI for Science. His research interests include robotic manipulation of deformable objects and data-driven modelling of dynamical systems.
\end{IEEEbiography}

\begin{IEEEbiography}[{\includegraphics[width=1in,height=1.25in,clip,keepaspectratio]{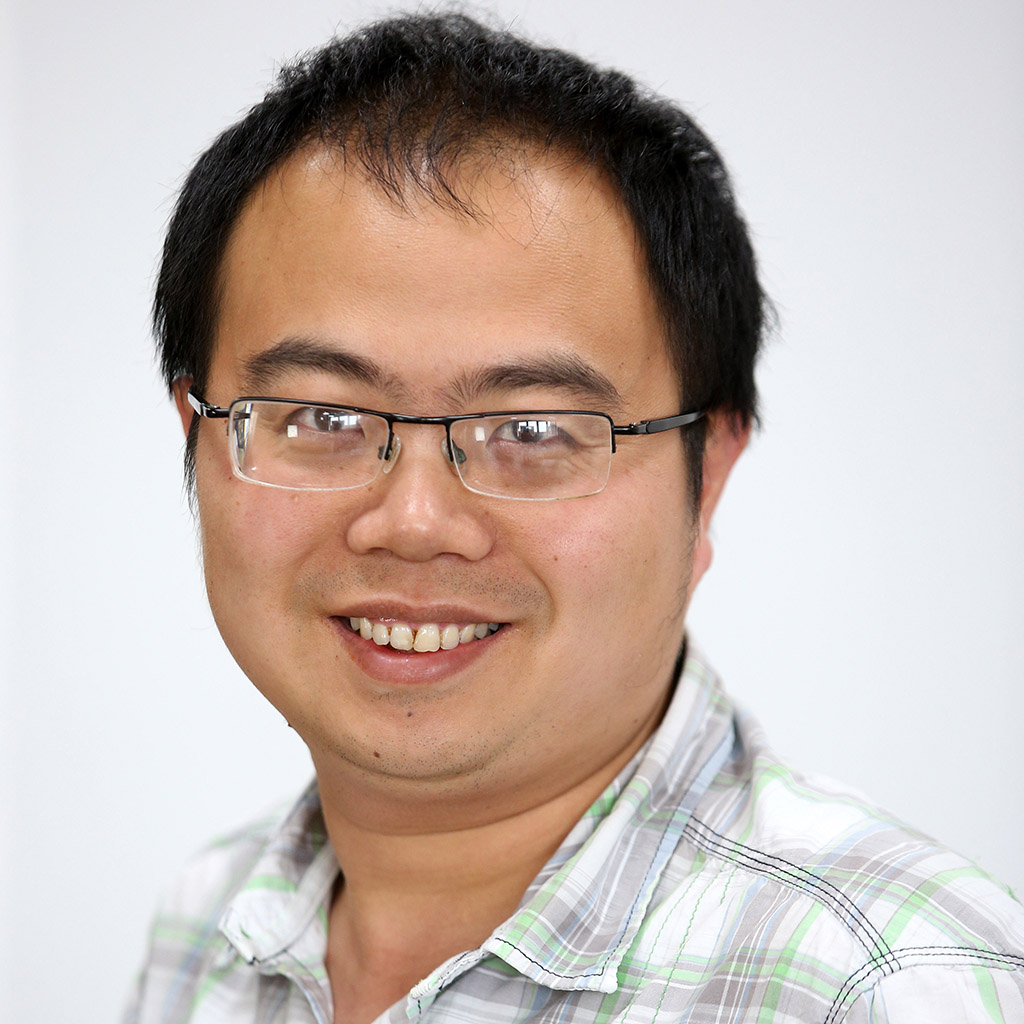}}]{Yu-Kun Lai} (Senior Member, IEEE) received his bachelor’s and PhD degrees in computer science from Tsinghua University, in 2003 and 2008, respectively. He is currently a professor in the School of Computer Science \& Informatics, Cardiff University, UK. His research interests include computer graphics, geometry processing, image processing, and computer vision. He is on the editorial boards of IEEE Transactions on Visualization and Computer Graphics, Computers \& Graphics, and The Visual Computer.
\end{IEEEbiography}

\begin{IEEEbiography}[{\includegraphics[width=1in,height=1.25in,clip,keepaspectratio]{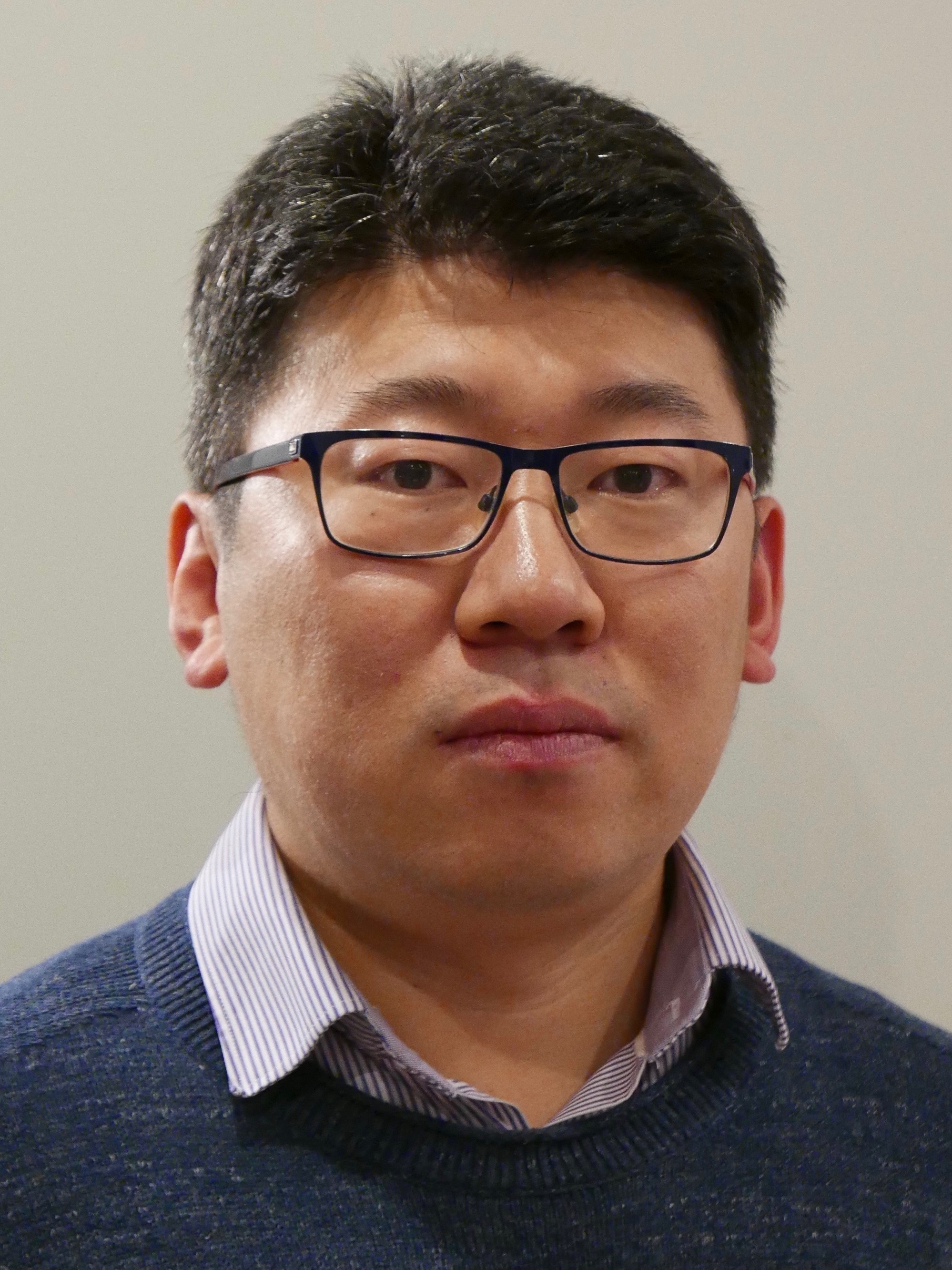}}]{Ze Ji}
(Member, IEEE) received the Ph.D. degree from Cardiff University, Cardiff, U.K., in 2007. He is a Reader with the School of Engineering, Cardiff University, UK. Prior to his current position, he was working in industry (Dyson, Lenovo, etc) on autonomous robotics. His research interests are broad, including robot manipulation, robot learning, autonomous robot navigation, physics-informed learning, computer vision, simultaneous localization and mapping (SLAM), acoustic localization, and tactile sensing. He is on the editorial boards of several journals, including IEEE/ASME Transactions on Mechatronics.
\end{IEEEbiography}

\clearpage
\section*{APPENDIX}
\subsection{Gradients of the skill parameters from actions through the skill-to-action mapping}
\rp{This subsection presents the gradient chain of the skill-to-action mapping function, which is defined mathematically by Eqs.~\ref{eq:d1} to~\ref{eq:delta_d4}. We are concerned of the gradients $\frac{\partial \bar{a}(\pmb{\theta})}{\partial \pmb{\theta}}$. The following derives the gradient of each parameter from phases 1 to 4, assuming the calculation of a rounded number of timesteps is adopted. For phases 1 and 4, the alternative approach of using a unrounded number of timesteps in per-action displacement calculation is also presented. Please also refer to Table~\ref{tab:skill-notation} for math notations.}

\rp{The first parameter, $\theta_{dispalce}$, is involved in computing only the first phase of the skill actions. Thus, with $T_1=0$ ($\theta_{dispalce}=\theta_{rotate}=0$), $\frac{\partial \bar{a}}{\partial \theta_{dispalce}}=\frac{\partial \bar{a}[0:T_1]}{\partial \theta_{dispalce}}=0$. Then, because $a_t$ is the same within each phase, for any timestep $t$ that satisfies $T_1>t>0$, we have:}

\begingroup\makeatletter\def\f@size{8}\check@mathfonts\rp{\vspace{-0.3cm}
\begin{flalign*}
&\frac{\partial \bar{a}[0:T_1]}{\partial \theta_{displace}}\Bigg|_{rounded\_T_1}=\sum_{t=0}^{T_1}\frac{\partial a_t}{\partial \theta_{displace}}=T_1\cdot\frac{\partial a_t}{\partial \Delta x_1}\cdot\frac{\partial\Delta x_1}{\partial d_1}\cdot\frac{\partial d_1}{\partial \theta_{displace}}&&
\end{flalign*}}
\endgroup\vspace{-0.3cm}

\noindent \rp{where,}

\begingroup\makeatletter\def\f@size{8}\check@mathfonts\rp{\vspace{-0.3cm}
\begin{flalign*}
&\frac{\partial a_t}{\partial \Delta x_1}=\frac{\text{d} a_t[0]}{\text{d} \Delta x_1}=1,\ \ \frac{\partial\Delta x_1}{\partial d_1}=\frac{\text{d}\Delta x_1}{\text{d} d_1}=\frac{1}{T_1}, && \\
&\frac{\partial d_1}{\partial \theta_{displace}}=\frac{\text{d}d_1}{\text{d}\theta_{displace}}=0.12&&
\end{flalign*}}
\endgroup\vspace{-0.3cm}

\rp{Thus, the gradient of $\theta_{dispalce}$ with the rounded number of timesteps being used in skill-to-action mapping is:}

\begingroup\makeatletter\def\f@size{8}\check@mathfonts\rp{\vspace{-0.3cm}
\begin{flalign*}
&\frac{\partial \bar{a}[0:T_1]}{\partial \theta_{displace}}\Bigg|_{rounded\_T_1}=0.12&&
\end{flalign*}}
\endgroup\vspace{-0.3cm}

\rp{If $T_1$ is replaced with the unrounded $T^{float}_1$ to compute the per-step action, for any timestep $t$ that satisfies $T_1>t>0$, we have:}

\begingroup\makeatletter\def\f@size{8}\check@mathfonts\rp{\vspace{-0.3cm}
\begin{flalign*}
&\hspace{-0.5cm}\frac{\partial \bar{a}[0:T_1]}{\partial \theta_{displace}}\Bigg|_{unrounded\_T_1}=\sum_{t=0}^{T_1}\frac{\partial a_t}{\partial \theta_{displace}}&&\\
=&T_1\cdot\left[\frac{\partial a_t}{\partial \Delta x_1}\cdot\frac{\partial\Delta x_1}{\partial \theta_{displace}}+\frac{\partial a_t}{\partial \Delta rx_1}\cdot\frac{\partial\Delta rx_1}{\partial \theta_{displace}}\right]&&\\
=&T_1\cdot\left[\frac{\partial a_t}{\partial \Delta x_1}\cdot(\frac{\partial\Delta x_1}{\partial d_1}\cdot\frac{\partial d_1}{\partial \theta_{displace}}+\frac{\partial\Delta x_1}{\partial T^{float}_1}\cdot\frac{\partial T^{float}_1}{\partial \theta_{displace}})\right.&&\\
&\left.+\frac{\partial a_t}{\partial \Delta rx_1}\cdot\frac{\partial\Delta rx_1}{\partial T^{float}_1}\cdot\frac{\partial T^{float}_1}{\partial \theta_{displace}}\right]&&
\end{flalign*}}
\endgroup\vspace{-0.3cm}

\noindent \rp{where,}

\begingroup\makeatletter\def\f@size{8}\check@mathfonts\rp{\vspace{-0.3cm}
\begin{flalign*}
&\frac{\partial\Delta x_1}{\partial d_1}=\frac{\text{d}\Delta x_1}{\text{d} d_1}=\frac{1}{T^{float}_1},&&\\
&\frac{\partial\Delta x_1}{\partial T^{float}_1}=\frac{\text{d}\Delta x_1}{\text{d} T^{float}_1}=\frac{-d_1}{(T^{float}_1)^2},&&\\
&\frac{a_t}{\Delta rx_1}=1,\ \ \frac{\partial\Delta rx_1}{\partial T^{float}_1}=\frac{\text{d}\Delta rx_1}{\text{d} T^{float}_1}=\frac{-\varphi_1}{(T^{float}_1)^2}&&\\
&\frac{\partial T^{float}_1}{\partial \theta_{displace}}=
    \begin{cases}
        \frac{\partial T^{float}_{d_1}}{\partial |d_1|}\cdot\frac{\partial |d_1|}{\partial \theta_{displace}}, &T^{float}_{d_1}\geq T^{float}_{\varphi_1}>0\\
        \frac{\partial T^{float}_{\varphi_1}}{\partial \theta_{displace}}=0, &0<T^{float}_{d_1}<T^{float}_{\varphi_1}
    \end{cases},&&\\
\end{flalign*}}
\endgroup\vspace{-0.3cm}

\noindent \rp{where,}

\begingroup\makeatletter\def\f@size{8}\check@mathfonts\rp{\vspace{-0.3cm}
\begin{flalign*}
&\frac{\partial T^{float}_{d_1}}{\partial |d_1|}=\frac{1}{v_l\cdot\Delta t},\ \ \frac{\partial |d_1|}{\partial \theta_{displace}}=\begin{cases}
    0.12, &\theta_{displace}>0\\
    -0.12, &\theta_{displace}<0\\
    0, &\theta_{displace}=0
    \end{cases}&&
\end{flalign*}}
\endgroup\vspace{-0.3cm}

\rp{Thus, we have}
\begingroup\makeatletter\def\f@size{8}\check@mathfonts\rp{\vspace{-0.3cm}
\begin{flalign*}
    &\frac{\partial \bar{a}[0:T_1]}{\partial \theta_{displace}}\Bigg|_{unrounded\_T_1}&&\\
    &=\begin{cases}
    T_1\cdot\left[1\cdot(\frac{1}{T^{float}_{d_1}}\cdot0.12 + \frac{-d_1}{(T^{float}_{d_1})^2}\cdot\frac{1}{v_l\cdot\Delta t}\cdot0.12)\right.\\
    \hspace{0.2cm}\left.+1\cdot\frac{-\varphi_1}{(T^{float}_{d_1})^2}\cdot\frac{1}{v_l\cdot\Delta t}\cdot0.12\right],\\
    \hspace{3cm} T^{float}_{d_1}\geq T^{float}_{\varphi_1}>0,\ \theta_{displace}>0\\
    T_1\cdot\left[1\cdot(\frac{1}{T^{float}_{d_1}}\cdot0.12 + \frac{-d_1}{(T^{float}_{d_1})^2}\cdot\frac{1}{v_l\cdot\Delta t}\cdot-0.12)\right.\\
    \hspace{0.2cm}\left.+1\cdot\frac{-\varphi_1}{(T^{float}_{d_1})^2}\cdot\frac{1}{v_l\cdot\Delta t}\cdot-0.12\right],\\ 
    \hspace{3cm} T^{float}_{d_1}\geq T^{float}_{\varphi_1}>0,\ \theta_{displace}<0\\
    T_1\cdot1\cdot\frac{1}{T^{float}_{\varphi_1}}\cdot0.12,\hspace{0.35cm}0<T^{float}_{d_1}<T^{float}_{\varphi_1}
    \end{cases}&&\\
    &=\begin{cases}
    T_1\cdot\frac{-\pi\cdot\theta_{rotate}\cdot v_l\cdot\Delta t}{0.36\cdot{\theta_{displace}}^2},\ \ T^{float}_{d_1}\geq T^{float}_{\varphi_1}>0,\ \theta_{displace}>0\\
    T_1\cdot\left(\frac{2\cdot v_l\cdot\Delta t}{\theta_{displace}}+\frac{\pi\cdot\theta_{rotate}\cdot v_l\cdot\Delta t}{0.36\cdot{\theta_{displace}}^2}\right),\\
    \hspace{3cm} T^{float}_{d_1}\geq T^{float}_{\varphi_1}>0,\ \theta_{displace}<0\\
    T_1\cdot\frac{0.12\cdot v_w\cdot\Delta t}{|\theta_{rotate}|\cdot\pi/3},\hspace{0.7cm} 0<T^{float}_{d_1}<T^{float}_{\varphi_1}
    \end{cases}&&
\end{flalign*}}
\endgroup\vspace{-0.3cm}

\rp{Secondly, as $\theta_{rotate}$ is involved in the computation of skill phases 1, 2 and 4, its gradient is $\frac{\partial \bar{a}}{\partial \theta_{rotate}}=\frac{\partial \bar{a}[0:T_1]}{\partial \theta_{rotate}}+\frac{\partial \bar{a}[T_1:T_1+T_2]}{\partial \theta_{rotate}}+\frac{\partial \bar{a}[T_1+T_2+T_3:T_1+T_2+T_3+T_4]}{\partial \theta_{rotate}}$. Again, with $T_1=0$ ($\theta_{dispalce}=\theta_{rotate}=0$), we have $\frac{\partial \bar{a}[0:T_1]}{\partial \theta_{rotate}}=0$. Then, for any timestep $t$ that satisfies $T_1>t>0$, we have:}

\begingroup\makeatletter\def\f@size{8}\check@mathfonts\rp{\vspace{-0.3cm}
\begin{flalign*}
    \frac{\partial \bar{a}[0:T_1]}{\partial \theta_{rotate}}\Bigg|_{rounded\_T_1}&=T_1\cdot\frac{\partial a_t}{\partial \Delta rx_1}\cdot\frac{\partial \Delta rx_1}{\partial \varphi_1}\cdot\frac{\partial \varphi_1}{\partial \theta_{rotate}}=\frac{\pi}{3}
\end{flalign*}}
\endgroup\vspace{-0.3cm}

\rp{Again, if $T_1$ is replaced with the unrounded value, we have}

\begingroup\makeatletter\def\f@size{8}\check@mathfonts\rp{\vspace{-0.3cm}
\begin{flalign*}
    &\frac{\partial \bar{a}[0:T_1]}{\partial \theta_{rotate}}\Bigg|_{unrounded\_T_1}&&\\
    &=T_1\cdot\left[\frac{\partial a_t}{\partial \Delta rx_1}\cdot(\frac{\partial \Delta rx_1}{\partial \varphi_1}\cdot\frac{\partial \varphi_1}{\partial \theta_{rotate}}+\frac{\partial \Delta rx_1}{\partial T^{float}_1}\cdot\frac{\partial T^{float}_1}{\partial \theta_{rotate}})\right.&&\\
    &\left.+\frac{\partial a_t}{\partial \Delta x_1}\cdot\frac{\partial\Delta x_1}{\partial T^{float}_1}\cdot\frac{\partial T^{float}_1}{\partial \theta_{rotate}}\right]&&
\end{flalign*}}
\endgroup\vspace{-0.3cm}

\noindent\rp{where,}

\begingroup\makeatletter\def\f@size{8}\check@mathfonts\rp{\vspace{-0.3cm}
\begin{flalign*}
&\frac{\partial\Delta rx_1}{\partial \varphi_1}=\frac{\text{d}\Delta rx_1}{\text{d} \varphi_1}=\frac{1}{T^{float}_1},&&\\
&\frac{\partial T^{float}_1}{\partial \theta_{rotate}}=
    \begin{cases}
        \frac{\partial T^{float}_{\varphi_1}}{\partial \theta_{rotate}}=0, &T^{float}_{d_1}\geq T^{float}_{\varphi_1}>0\\
        \frac{\partial T^{float}_{d_1}}{\partial |\varphi_1|}\cdot\frac{\partial |\varphi_1|}{\partial \theta_{rotate}}, &0<T^{float}_{d_1}<T^{float}_{\varphi_1}
    \end{cases},&&\\
&\frac{\partial T^{float}_{d_1}}{\partial |\varphi_1|}=\frac{1}{v_w\cdot\Delta t},\ \ \frac{\partial |\varphi_1|}{\partial \theta_{rotate}}=\begin{cases}
    \pi/3, &\theta_{rotate}>0\\
    -\pi/3, &\theta_{rotate}<0\\
    0, &\theta_{rotate}=0
    \end{cases}&&
\end{flalign*}}
\endgroup\vspace{-0.3cm}

\rp{Thus,}

\begingroup\makeatletter\def\f@size{8}\check@mathfonts\rp{\vspace{-0.3cm}
\begin{flalign*}
    &\frac{\partial \bar{a}[0:T_1]}{\partial \theta_{rotate}}\Bigg|_{unrounded\_T_1}&&\\
    &=\begin{cases}
    T_1\cdot\left[1\cdot(\frac{1}{T^{float}_{\varphi_1}}\cdot\frac{\pi}{3} + \frac{-\varphi_1}{(T^{float}_1)^2}\cdot\frac{1}{v_w\cdot\Delta t}\cdot\frac{\pi}{3})\right.\\
    \hspace{0.2cm} \left.+1\cdot \frac{-d_1}{(T^{float}_{\varphi_1})^2}\cdot\frac{\pi/3}{v_w\cdot\Delta t}\right],\\
    \hspace{2.9cm} T^{float}_{\varphi_1}>T^{float}_{d_1}>0,\ \theta_{rotate}>0\\
    T_1\cdot\left[1\cdot(\frac{1}{T^{float}_{\varphi_1}}\cdot\frac{\pi}{3} + \frac{-\varphi_1}{(T^{float}_1)^2}\cdot\frac{1}{v_w\cdot\Delta t}\cdot-\frac{\pi}{3}))\right.,\\ 
    \hspace{0.2cm} \left.+1\cdot \frac{-d_1}{(T^{float}_{\varphi_1})^2}\cdot\frac{-\pi/3}{v_w\cdot\Delta t}\right],\\
    \hspace{2.9cm} T^{float}_{\varphi_1}>T^{float}_{d_1}>0,\ \theta_{rotate}<0\\
    T_1\cdot1\cdot\frac{1}{T^{float}_{d_1}}\cdot\frac{\pi}{3},\hspace{0.5cm}T^{float}_{\varphi_1}\leq T^{float}_{d_1}
    \end{cases}&&\\
    &=\begin{cases}
    T_1\cdot\frac{-0.36\cdot\theta_{displace}\cdot v_w\cdot\Delta t}{{\theta_{rotate}}^2\cdot\pi},\\
    \hspace{2.9cm} T^{float}_{\varphi_1}>T^{float}_{d_1}>0,\ \theta_{rotate}>0\\
    T_1\cdot\left(\frac{2\cdot v_w\Delta t}{\theta_{rotate}}+\frac{0.36\cdot\theta_{displace}\cdot v_w\cdot\Delta t}{{\theta_{rotate}}^2\cdot\pi}\right),\\
    \hspace{2.9cm} T^{float}_{\varphi_1}>T^{float}_{d_1}>0,\ \theta_{rotate}<0\\
    T_1\cdot\frac{\frac{\pi}{3}\cdot v_l\cdot\Delta t}{|\theta_{displace}|\cdot0.12},\hspace{0.4cm} 0<T^{float}_{\varphi_1}\leq T^{float}_{d_1}
    \end{cases}&&
\end{flalign*}}
\endgroup\vspace{-0.3cm}

\rp{Note that we assume $T^{float}_{1}=T^{float}_{d_1}$ when $T^{float}_{d_1}=T^{float}_{\varphi_1}$ for the $max()$ operation (Eq.~\ref{eq:Ti1}).}

\rp{The next item of the gradient of $\theta_{rotate}$ is $\frac{\partial \bar{a}[T_1:T_1+T_2]}{\partial \theta_{rotate}}$, where $T_2$ is kept an integer. Again, it is zero with $T_2=0$, and for any timestep $t$ that satisfies $T_1+T_2>t>T_1$, we have:}

\begingroup\makeatletter\def\f@size{8}\check@mathfonts\rp{\vspace{-0.3cm}
\begin{flalign*}
    &\frac{\partial \bar{a}[T_1:T_1+T_2]}{\partial \theta_{rotate}}\Bigg|_{rounded\_T_2}=T_2\cdot\left[\frac{\partial a_t}{\partial \Delta x_2}\cdot\frac{\partial \Delta x_2}{\partial \varphi_1}\cdot\frac{\partial \varphi_1}{\partial \theta_{rotate}}\right.&&\\
    &\left.+\frac{\partial a_t}{\partial \Delta z_2}\cdot\frac{\partial \Delta z_2}{\partial \varphi_1}\cdot\frac{\partial \varphi_1}{\partial \theta_{rotate}}\right]&&\\
    &=T_2\cdot\frac{-d_2}{T_2}\cdot\frac{\pi}{3}\cdot\left(\sin(\varphi_1+\frac{\pi}{2})+\cos(\varphi_1+\frac{\pi}{2})\right)&&\\
    &=-d_2\cdot\frac{\pi}{3}\cdot\left(\sin(\varphi_1+\frac{\pi}{2})+\cos(\varphi_1+\frac{\pi}{2})\right)&&
\end{flalign*}}
\endgroup\vspace{-0.3cm}

\noindent \rp{where $d_2=(\theta_{insert\_dist}+1)\cdot 0.03$ and $\varphi_1=\theta_{rotate}\cdot\frac{\pi}{3}$. Note that phases 2 and 3 are not suitable for the unrounded number replacement of the number of timesteps due to the zero gradient issue.}

\rp{The third part of the gradient of $\theta_{rotate}$ comes from skill phase 4. Because $T_4>=T^{float}_{lift}>0$, for any timestep $t$ that satisfies $T_1+T_2+T_3+T_4>t>T_1+T_2+T_3$, we have}

\begingroup\makeatletter\def\f@size{8}\check@mathfonts\rp{\vspace{-0.3cm}
\begin{flalign*}
    &\frac{\partial \bar{a}[T_1+T_2+T_3:T_1+T_2+T_3+T_4]}{\partial \theta_{rotate}}\Bigg|_{rounded\_T_4}&&\\
    &=T_4\cdot\frac{\partial a_t}{\partial \Delta rx_4}\cdot\frac{\partial \Delta rx_4}{\partial \varphi_1}\cdot\frac{\partial \varphi_1}{\partial \theta_{rotate}}&&\\
    &=T_4\cdot1\cdot-\frac{1}{T_4}\cdot\frac{\pi}{3}=-\frac{\pi}{3}&&
\end{flalign*}}
\endgroup\vspace{-0.3cm}

\rp{If $T_4$ is replaced with the unrounded number, we have}
    
\begingroup\makeatletter\def\f@size{8}\check@mathfonts\rp{\vspace{-0.3cm}
\begin{flalign*}
    &\frac{\partial \bar{a}[T_1+T_2+T_3:T_1+T_2+T_3+T_4]}{\partial \theta_{rotate}}\Bigg|_{unrounded\_T_4}&&\\
    &=T_4\cdot\left[\frac{\partial a_t}{\partial \Delta rx_4}\cdot(\frac{\partial \Delta rx_4}{\partial \varphi_1}\cdot\frac{\partial \varphi_1}{\partial \theta_{rotate}}+\frac{\partial \Delta rx_4}{\partial T^{float}_4}\cdot\frac{\partial T^{float}_4}{\partial \theta_{rotate}})\right.&&\\
    & \left.+\frac{\partial a_t}{\partial \Delta z_4}\cdot\frac{\partial\Delta z_4}{\partial T^{float}_4}\cdot\frac{\partial T^{float}_4}{\partial \theta_{rotate}}\right]&&
\end{flalign*}}
\endgroup\vspace{-0.3cm}

\noindent\rp{where,}

\begingroup\makeatletter\def\f@size{8}\check@mathfonts\rp{\vspace{-0.3cm}
\begin{flalign*}
&\frac{\partial\Delta rx_4}{\partial \varphi_1}=\frac{\text{d}\Delta rx_4}{\text{d} \varphi_1}=-\frac{1}{T^{float}_4},&&\\
&\frac{\partial\Delta rx_4}{\partial T^{float}_4}=\frac{\text{d}\Delta rx_4}{\text{d} T^{float}_4}=\frac{-\varphi_1}{(T^{float}_4)^2}&&\\
&\frac{\partial T^{float}_4}{\partial \theta_{rotate}}=
    \begin{cases}
        \frac{\partial T^{float}_{d_4}}{\partial \theta_{rotate}}=0, &T^{float}_{lift}> T^{float}_{\varphi_1}>0\\
        \frac{\partial T^{float}_{\varphi_1}}{\partial |\varphi_1|}\cdot\frac{\partial |\varphi_1|}{\partial \theta_{rotate}}, &0<T^{float}_{lift}\leq T^{float}_{\varphi_1}
    \end{cases}&&
\end{flalign*}}
\endgroup\vspace{-0.3cm}

\rp{Thus, we have}

\begingroup\makeatletter\def\f@size{8}\check@mathfonts\rp{\vspace{-0.3cm}
\begin{flalign*}
    &\frac{\partial \bar{a}[T_1+T_2+T_3:T_1+T_2+T_3+T_4]}{\partial \theta_{rotate}}\Bigg|_{unrounded\_T_4}&&\\
    &=\begin{cases}
        T_4\cdot\frac{-3\cdot d_4\cdot v_w\cdot\Delta t}{{\theta_{rotate}}^2\cdot\pi}, \ \ T^{float}_{\varphi_1}\geq T^{float}_{lift}>0, \ \theta_{rotate}>0\\
        T_4\cdot\left(\frac{-2\cdot v_w\Delta t}{\theta_{rotate}}+\frac{3\cdot d_4\cdot v_w\cdot\Delta t}{{\theta_{rotate}}^2\cdot\pi}\right),\\
        \hspace{2.45cm} T^{float}_{\varphi_1}\geq T^{float}_{lift}>0, \ \theta_{rotate}<0\\
        T_4\cdot\frac{-\pi\cdot v_l\cdot\Delta t}{3\cdot d_4}, \hspace{0.6cm} 0<T^{float}_{\varphi_1}<T^{float}_{lift}
    \end{cases}
\end{flalign*}}
\endgroup\vspace{-0.3cm}

\rp{Note $d_4$ is a given constant.} 

\rp{Thirdly, $\frac{\partial \bar{a}}{\partial \theta_{insert\_dist}}=\frac{\partial \bar{a}[T_1:T_1+T_2]}{\partial \theta_{insert\_dist}}$. Again, with $T_2>0$ ($\theta_{insert\_dist}>-1$ and $T_2$ remains an integer), for any timestep $t$ that satisfies $T_1+T_2>t>T_1$, we have:}

\begingroup\makeatletter\def\f@size{8}\check@mathfonts\rp{\vspace{-0.3cm}
\begin{flalign*}
    &\frac{\partial \bar{a}[T_1:T_1+T_2]}{\partial \theta_{insert\_dist}}\Bigg|_{rounded\_T_2}&&\\
    &=T_2\cdot\left[\frac{\partial a_t}{\partial \Delta x_2}\cdot\frac{\partial \Delta x_2}{\partial d_2}\cdot\frac{\partial d_2}{\partial \theta_{insert\_dist}}+\frac{\partial a_t}{\partial \Delta z_2}\cdot\frac{\partial \Delta z_2}{\partial d_2}\cdot\frac{\partial d_2}{\partial \theta_{insert\_dist}}\right]&&\\
    &=T_2 \cdot\left[1\cdot\frac{\cos\varphi_2}{T_2}\cdot0.03 +1\cdot\frac{-\sin\varphi_2}{T_2}\cdot0.03\right]&&\\
    &=0.03\cdot(\cos\varphi_2-\sin\varphi_2)&&
\end{flalign*}}
\endgroup\vspace{-0.3cm}

\rp{Fourthly, we derive the gradient of $\theta_{push\_dist}$. Again, with $T_3=0$, we have $\frac{\partial \bar{a}}{\partial \theta_{push\_dist}}=\frac{\partial \bar{a}[T_1+T_2:T_1+T_2+T_3]}{\partial \theta_{push\_dist}}=0$, while when $T_3>0$ ($T_3$ remains an integer), for any timestep $t$ that satisfies $T_1+T_2+T_3>t>T_1+T_2$, we have:}

\begingroup\makeatletter\def\f@size{8}\check@mathfonts\rp{\vspace{-0.3cm}
\begin{flalign*}
    &\frac{\partial \bar{a}[T_1+T_2:T_1+T_2+T_3]}{\partial \theta_{push\_dist}}\Bigg|_{rounded\_T_3}&&\\
    &=T_3\cdot\left[\frac{\partial a_t}{\partial \Delta x_3}\cdot\frac{\partial \Delta x_3}{\partial d_3}\cdot\frac{\partial d_3}{\partial \theta_{push\_dist}}+\frac{\partial a_t}{\partial \Delta z_3}\cdot\frac{\partial \Delta z_3}{\partial d_3}\cdot\frac{\partial d_3}{\partial \theta_{push\_dist}}\right]&&\\
    &=T_3\cdot\left[1\cdot\frac{\cos(\varphi_3)}{T_3}\cdot0.1+1\cdot\frac{\sin(\varphi_3)}{T_3}\cdot0.1\right]&&\\
    &=0.1\cdot\left[\cos(\varphi_3)+\sin(\varphi_3)\right]&&
\end{flalign*}}
\endgroup\vspace{-0.3cm}

\rp{Lastly, we derive the gradient of $\theta_{push\_angle}$. With $T_3=0$, we have $\frac{\partial \bar{a}}{\partial \theta_{push\_angle}}=\frac{\partial \bar{a}[T_1+T_2:T_1+T_2+T_3]}{\partial \theta_{push\_angle}}=0$, while when $T_3>0$ ($T_3$ remains an integer), for any timestep $t$ that satisfies $T_1+T_2+T_3>t>T_1+T_2$, we have:}

\begingroup\makeatletter\def\f@size{8}\check@mathfonts\rp{\vspace{-0.3cm}
\begin{flalign*}
    &\frac{\partial \bar{a}[T_1+T_2:T_1+T_2+T_3]}{\partial \theta_{push\_angle}}\Bigg|_{rounded\_T_3}&&\\
    &=T_3\cdot\left[\frac{\partial a_t}{\partial \Delta x_3}\cdot\frac{\partial \Delta x_3}{\partial \varphi_3}\cdot\frac{\partial \varphi_3}{\partial \theta_{push\_angle}}+\frac{\partial a_t}{\partial \Delta z_3}\cdot\frac{\partial \Delta z_3}{\partial \varphi_3}\cdot\frac{\partial \varphi_3}{\partial \theta_{push\_angle}}\right]&&\\
    &=T_3\cdot\left[1\cdot\frac{d_3}{T_3}\cdot-\sin(\varphi_3)\cdot\frac{\pi}{3}+1\cdot\frac{d_3}{T_3}\cdot\cos(\varphi_3)\cdot\frac{\pi}{3}\right]&&\\
    &=\frac{\pi\cdot d_3}{3}[\cos(\varphi_3)-\sin(\varphi_3)]
\end{flalign*}}
\endgroup\vspace{-0.3cm}

\noindent\rp{where $d_3=(\theta_{push\_dist}+1)\cdot0.1+0.04$ and $\varphi_3=(\theta_{push\_angle}+3)\cdot\pi/3$.}

\subsection{Differentiability of the constitutive models}
We employ the Saint Venant–Kirchhoff (SVK) model as our elastic energy density function (constitutive model) and the Drucker-Prager (DP) following~\cite{klar2016drucker}. In our work, we require the gradients of key physical properties. To verify the differentiability of this process, we manually calculate the partial derivatives of the input variables (i.e., the Lam\'e constants of the soil) with respect to the output of the elastoplasticity model (Eqs.~\ref{eq:SVK} to~\ref{eq:DP}). In the DP plasticity model, for cases without projection and projection onto the cone tip, we have $\frac{\partial S'_p}{\partial \mu} = \frac{\partial S'_p}{\partial \lambda} = O_{dim}$, where $O_{dim}$ denotes the zero matrix of dimension $dim$. In contrast, for projection onto the cone surface:
\begin{equation}
\frac{\partial S'_p}{\partial \mu} = \frac{\partial S'_p}{\partial \delta_{\gamma_{p}}}\frac{\partial \delta_{\gamma_{p}}}{\partial \mu} \quad \frac{\partial S'_p}{\partial \lambda} = \frac{\partial S'_p}{\partial \delta_{\gamma_{p}}}\frac{\partial \delta_{\gamma_{p}}}{\partial \lambda}
\end{equation}
where the partial derivative of $S'_p$ with respect to $\delta_{\gamma_{p}}$ is:
\begin{equation}
\frac{\partial S'_p}{\partial \delta_{\gamma_{p}}} = -\exp(\pmb{\epsilon}_p-\delta \gamma_{p}\frac{\hat{\pmb{\epsilon}_p}}{\|\hat{\pmb{\epsilon}_p}\|}) \frac{\hat{\pmb{\epsilon}_p}}{\|\hat{\pmb{\epsilon}_p}\|}
\end{equation}
and the partial derivatives of $\delta_{\gamma_{p}}$ with respect to $\mu$ and $\lambda$ are:
\begin{equation}
\frac{\partial \delta_{\gamma_{p}}}{\partial \mu} = \frac{- \lambda dim \mathrm{tr}(\pmb{\epsilon}_p)\alpha_f}{2\mu^{2}} \quad \frac{\partial \delta_{\gamma_{p}}}{\partial \lambda} = \frac{dim \mathrm{tr}(\pmb{\epsilon}_p)\alpha_f}{2\mu}
\end{equation}
The partial derivatives of the updated deformation gradient $F'_p$ and its determinant $\det(F_p')$ with respect to $S'_p$ are given by:
\begin{equation}
\frac{\partial F'_p}{\partial S'_p} = U'_p V'^{T}_p \quad \frac{\partial \det(F_p')}{\partial S_p'} =
\det(F_p') F_p'^{-T} \frac{\partial F'_p}{\partial S'_p}
\end{equation}
And the partial derivatives of the Cauchy stress $\sigma_p$ with respect to $S'_p$ is:
\begin{align}
\frac{\partial \sigma_p}{\partial S_p'} &=
\frac{\partial}{\partial S_p'} \left( \frac{1}{\det(F_p')} \hat{\bm{P}_{p}} F_p'^{T} \right) \nonumber \\
&=
- \frac{1}{\det(F_p')^2} 
\frac{\partial \det(F_p')}{\partial S_p'} \hat{\bm{P}_{p}} F_p'^{T} \nonumber \\
&\quad + \frac{1}{\det(F_p')} \left(
\frac{\partial \hat{\bm{P}_{p}}}{\partial S_p'} F_p'^{T}
+ \hat{\bm{P}_{p}} \frac{\partial F_p'^{T}}{\partial S_p'}
\right)
\end{align}
where the partial derivative of $\hat{\bm{P}_{p}}$ with respect to $S'_p$ is:
\begin{equation}
\frac{\partial \hat{\bm{P}_{p}}}{\partial S'_p} = U'_p  \left(\frac{2 \mu (I_{dim} - \ln S'_p)}{(S'_p)^{2}} +  \frac{\lambda (I_{dim} - \mathrm{tr}(\ln S'_p))}{(S'_p)^{2}} \right) V'^{T}_p
\end{equation}
Therefore, we can obtain the partial derivative of 
the Cauchy stress $\sigma_p$ with respect to $\lambda$ and $\mu$:
\begin{equation}
\frac{\partial \sigma_p}{\partial \lambda} = \frac{\partial \sigma_p}{\partial S_p'} \frac{\partial S'_p}{\partial \lambda} \quad \frac{\partial \sigma_p}{\partial \mu} = \frac{\partial \sigma_p}{\partial S_p'} \frac{\partial S'_p}{\partial \mu}
\end{equation}

These derivatives confirm the differentiability of our constitutive model.

\subsection{Goal-conditioned soft actor-critic}
Goal-conditioned Reinforcement learning (GRL) algorithms optimise for a policy $\pi(\bar{a}|o)$ to take actions that maximise the future goal-conditioned return $R=\sum^T\mathbb{E}_{\bar{a}\sim\pi,o'\sim f(o,\bar{a})}[r(o',\hat{g})]$, where $T=1$ with a single temporal-abstracted action $\bar{a}=\pmb{\theta}$~\cite{sutton2018reinforcement,andrychowicz2017hindsight}. The soft actor-critic (SAC) method is an off-policy stochastic actor-critic RL method that maintains a so-called soft Q function, $q(o, \hat{g}, \bar{a})$, which estimates the entropy-augmented future expected return, and a stochastic policy that produces the mean and standard deviations of the action distribution (Gaussian) that maximising the following objective with data batches drawn from a replay buffer $\mathcal{D}$~\cite{haarnoja2018soft}: 

\begin{equation*}
\begin{split}
J(\pi) = \mathbb{E}_{o, \hat{g}\sim\mathcal{D}}\mathbb{E}_{\bar{a}\sim\pi}\Bigl[&min\{q_1^{\pi}(o, \hat{g},\bar{a}), q_2^{\pi}(o, \hat{g},\bar{a})\}\\
&-\alpha_{sac}\log\pi(\bar{a}|o, \hat{g})\Bigr]
\end{split}
\end{equation*}

\noindent where $\alpha_{sac}$ is a weight that controls the importance of maximising the policy entropy during policy optimisation, automatically updated during runtime following~\cite{haarnoja2018soft}. To reduce value overestimation, we maintain two q networks following~\cite{fujimoto2018addressing} and take the minimal value in policy optimisation. The objective of both critic networks in this one-step POGCMDP is to minimise the following mean-square error:

$$
J(q^{\pi}) = \mathbb{E}_{o, \hat{g},\bar{a},r\sim\mathcal{D}}(q^{\pi}(o, \hat{g},\bar{a}) - r)^2
$$

As indicated in~\cite{andrychowicz2017hindsight}, goal-relabelling helps to improve sample efficiency for goal-conditioned RL. Therefore, for every interaction experience ($o, \hat{g}, \bar{a}, r, o'$) collected during training, we duplicate the experience, replace the desired goal with the achieved goal (i.e., the resultant observation $o'\rightarrow\hat{g}$), recompute the goal-conditioned reward, and append the relabelled experience into the replay buffer. 

\textbf{Implementation details}: To process point cloud data, we employ the PointNet network architecture for a shared encoder for all q and policy networks~\cite{qi2017pointnet}. For the q networks, the encoded feature is passed to two MLP layers of size $1024$ and $512$ with batch normalisation and ReLU activation, with dropout ($0.4$) applied to the second layer, and then passed to the last MLP layer of size $256$, outputting the q value. For the policy network, the encoded feature is passed to three MLP layers of sizes $1024$, $256$ and $256$ with ReLU activation and then two separate MLP layers of size $256$ that output the mean (activated by the hyperbolic tangent function) and the logarithm of the standard deviation of a Gaussian policy distribution of the action's dimension. The parameters of the PointNet encoder are only updated through the critic loss function, while the policy objective is used to update the MLP layers only.

\subsection{Additional figures of gradients and loss landscapes}
\begin{figure}[h]
\begin{subfigure}{0.48\columnwidth}
\includegraphics[width=\columnwidth]{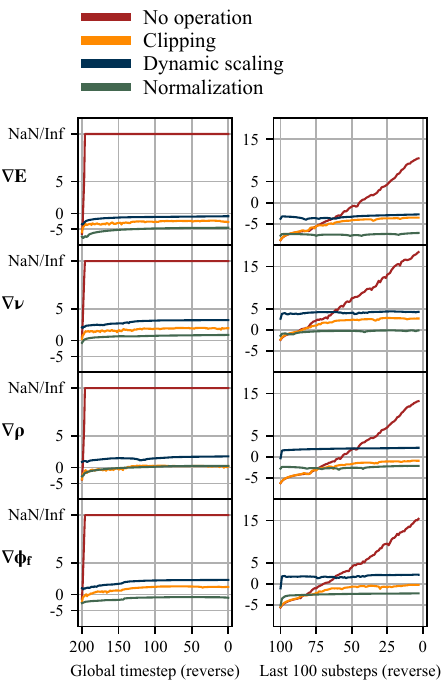}
\caption{Physics parameters (Young's modulus, Poisson's ratio, material density and sand friction angle)}\label{subfig:grad-params}
\end{subfigure}\hfil
\begin{subfigure}{0.48\columnwidth}
\includegraphics[width=\columnwidth]{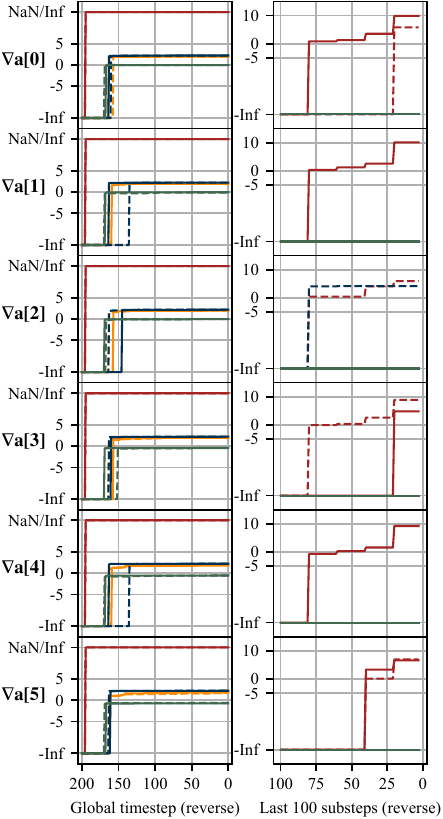}
\caption{Actions (6D)}\label{subfig:grad-action}
\end{subfigure}\\
\begin{subfigure}{0.48\columnwidth}
\includegraphics[width=\columnwidth]{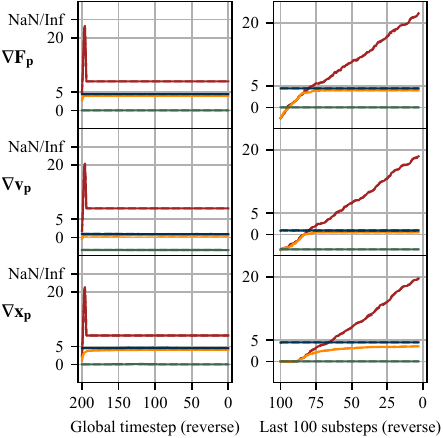}
\caption{Particle variables (deformation gradient, velocity and position)}\label{subfig:grad-p}
\end{subfigure}\hfil
\begin{subfigure}{0.48\columnwidth}
\includegraphics[width=\columnwidth]{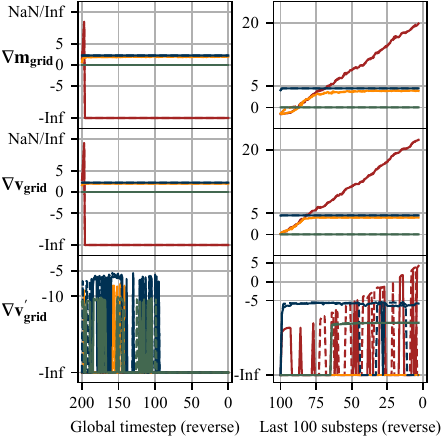}
\caption{Grid variables (mass, velocity before and after collision detection)}\label{subfig:grad-grid}
\end{subfigure}
\caption{\label{fig:ggrads}The log scale (base $10$) of the gradients of the key MLS-MPM variables, physics parameters, and actions from the last global timestep to the initial timestep (left side of each subfigure) as well as in the substep scale for $100$ substeps (right side of each subfigure). A global timestep $= 20$ substeps. Solid lines represent the maximal value and dash lines the minimal value of the gradient vector.}
\end{figure}

\rp{Figure~\ref{fig:ggrads} shows how the gradients of the key MLS-MPM variables, physics parameters, and actions evolve from the last global timestep to the initial timestep as well as how they evolve in the substep scale for $100$ substeps ($1$ global timestep $= 20$ substeps). These gradients are generated by simulating the trajectory produced by the skill prior and using the EMD loss to compute the loss with a soil-digging manipulation target.}

\rp{It is clear from the red lines (with no gradient operations) that the scales of the gradients grow sharply while back-propagating within $100$ substeps ($5$ global steps) and eventually result in $NaN$s or numerical infinities. On the other hand, with the three gradient operations, these gradients can be kept within a reasonable scale ($[-5, 5]$), as shown by the orange, blue, and green lines, corresponding to the clipping, dynamic scaling, and normalisation operations, respectively. This indicates the effectiveness of the three techniques in constraining the gradient scales of the key MLS-MPM variables, thus ensuring stable gradients for optimising the physics parameters and actions.}

\begin{figure*}
    \centering
    \begin{subfigure}{\linewidth}
    \centering
    \includegraphics[width=0.9\linewidth]{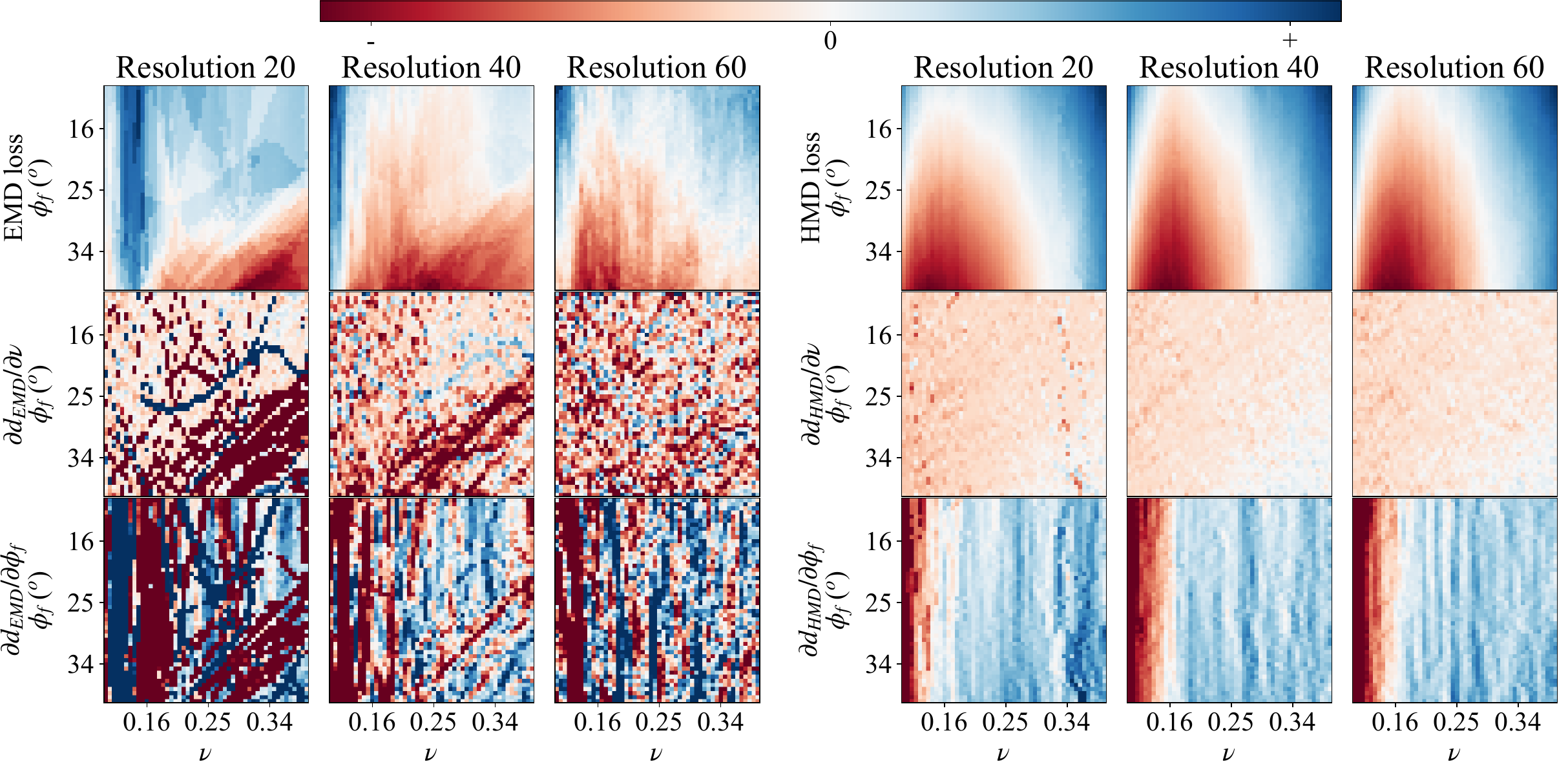}
    \caption{System identification task. On Poisson's ratio and sand friction angle.}\label{subfig:NS-loss}
    \end{subfigure}\\
    \begin{subfigure}{\linewidth}
    \centering
    \includegraphics[width=0.89\linewidth]{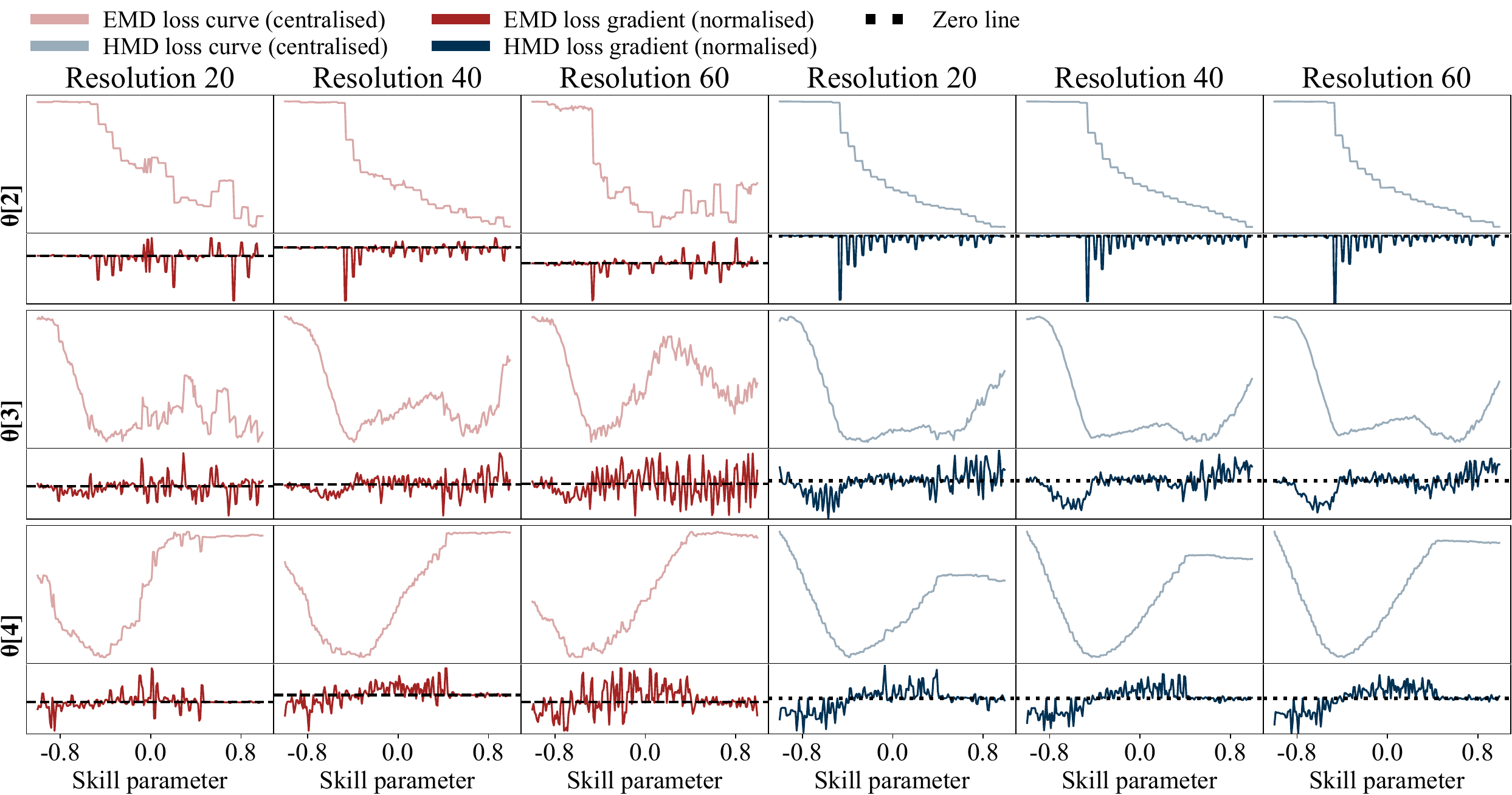}
    \caption{Skill optimisation task. On skill parameter $2$, $3$ and $4$.}\label{subfig:skill-loss-1}
    \end{subfigure}
    \caption{Loss landscapes and their gradient distributions.}
    \label{fig:loss-grad-app}
\end{figure*}

Figure~\ref{fig:loss-grad-app} shows the loss landscapes and the approximate gradients of the system identification task over Poisson's ratio and sand friction angle dimensions (top) as well as the second, third and fourth skill parameters for the first digging task (bottom). These results are also obtained with sampling resolutions $20$, $40$ and $60$. Similar to the analyses in subsection~\ref{subsec:grad-fluctuate}, it can be seen that HMD tends to produce smoother loss landscapes, indicated by the observation that the HMD gradients are less noisy.

\subsection{Training statistics of the reinforcement learning agent}
\begin{figure*}
    \centering
    \includegraphics[width=0.9\linewidth]{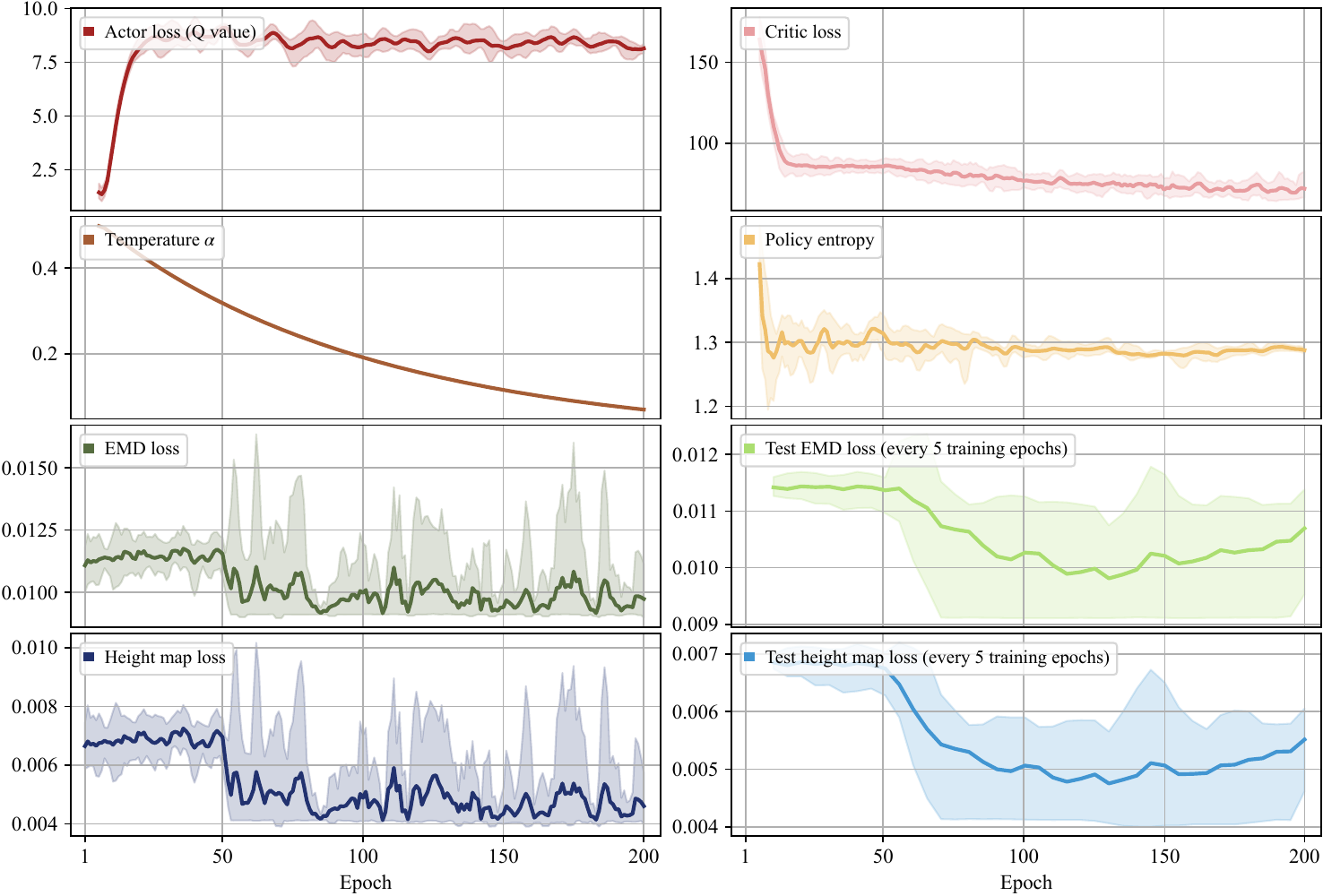}
    \caption{Goal-conditioned SAC training statistics on soil task 1.}
    \label{fig:stat-rl-0}
\end{figure*}

\begin{figure*}
    \centering
    \includegraphics[width=0.9\linewidth]{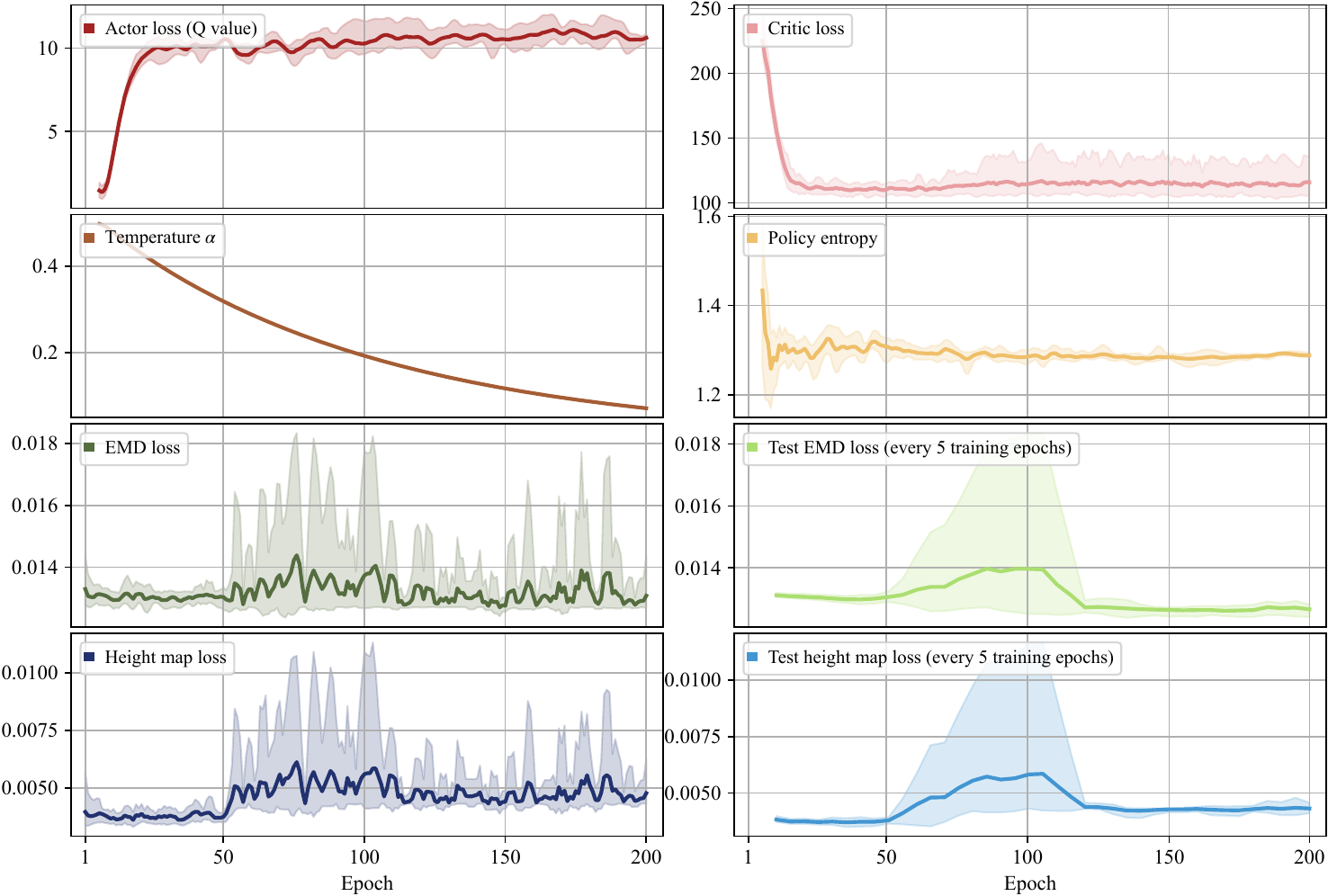}
    \caption{Goal-conditioned SAC training statistics on soil task 2.}
    \label{fig:stat-rl-1}
\end{figure*}

\begin{figure*}
    \centering
    \includegraphics[width=0.9\linewidth]{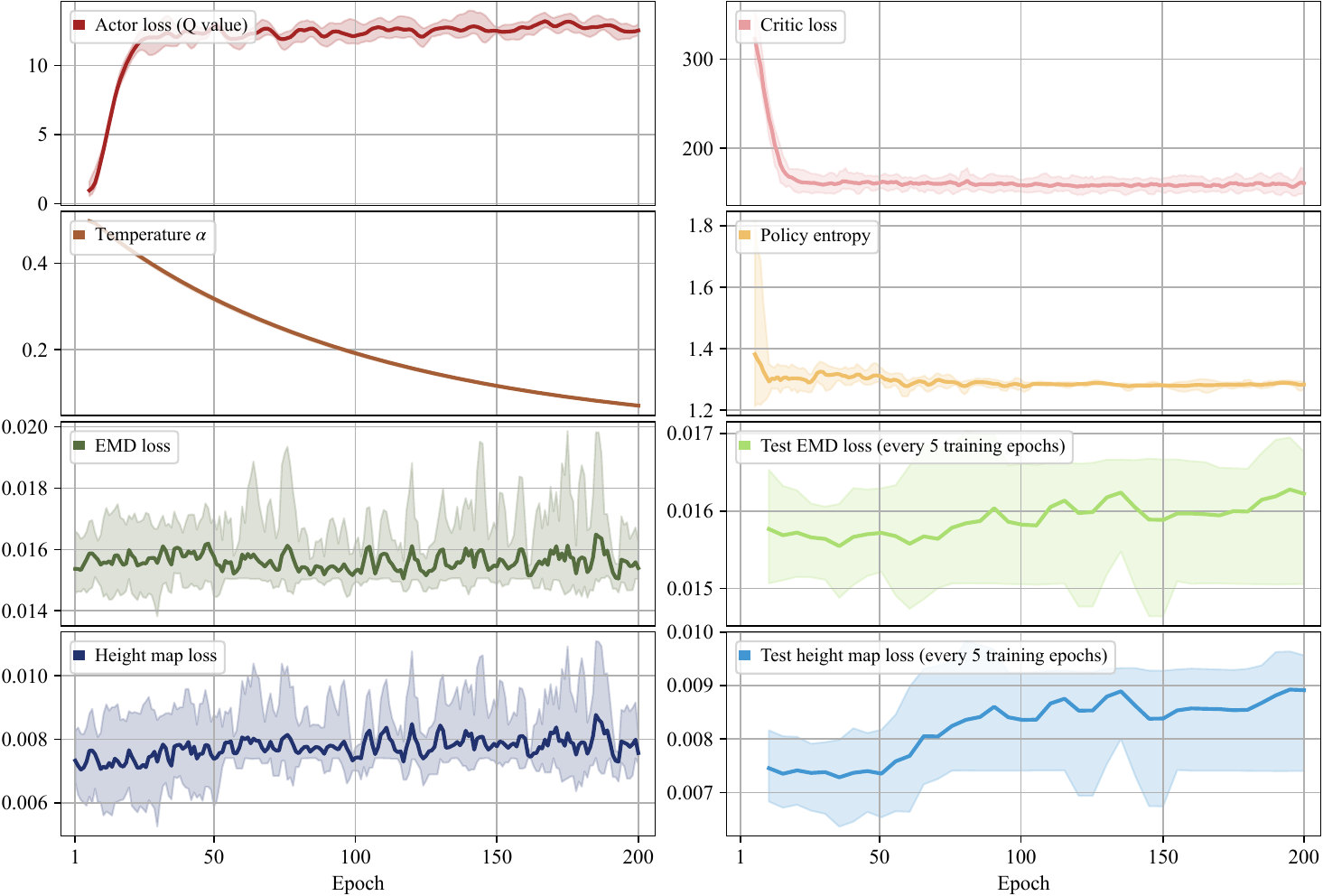}
    \caption{Goal-conditioned SAC training statistics on soil task 3.}
    \label{fig:stat-rl-2}
\end{figure*}

Figures~\ref{fig:stat-rl-0}, \ref{fig:stat-rl-1}, and \ref{fig:stat-rl-2} present the detailed statistics of the goal-conditioned SAC agent on the three soil digging tasks. In three tasks, the increased actor losses (estimated Q values), the decreased critic losses and the stabilised policy entropies indicate that the algorithms have converged. The EMD and height map losses, however, indicate that the algorithm could only find good solutions in task 1 while it failed at the other two tasks (this was also observed in Figure~\ref{subfig:task-baseline}).

\end{document}